\documentclass[hess, manuscript]{copernicus}
\nolinenumbers

\usepackage{amssymb}
\usepackage{ragged2e}
\usepackage[normalem]{ulem}
\usepackage{amsthm}
\usepackage{enumitem}
\usepackage[mathscr]{euscript}
\usepackage{array}
\usepackage{float}
\newcolumntype{P}[1]{>{\centering\arraybackslash}p{#1}}
\newcolumntype{M}[1]{>{\centering\arraybackslash}m{#1}}
\usepackage[utf8]{inputenc}
\usepackage[fleqn]{amsmath}
\usepackage{bm}
\usepackage{multirow}

\usepackage{mathtools}

\DeclareFontFamily{OT1}{pzc}{}
\DeclareFontShape{OT1}{pzc}{m}{it}{<-> s * [0.900] pzcmi7t}{}
\DeclareMathAlphabet{\mathpzc}{OT1}{pzc}{m}{it}
\usepackage{threeparttable}
\usepackage{graphicx}
\usepackage{picins}
\usepackage{fontawesome}
\usepackage{textgreek}
\usepackage{rotating}
\usepackage{makecell}
\usepackage{longtable}
\usepackage{pdflscape}

\usepackage{color}

\newcommand{\REDD}[1]{{\color{red}{#1}}}
\newcommand{\BLUEE}[1]{{\color{blue}{#1}}}

\usepackage{footmisc}
\usepackage{scrextend}
\usepackage{amsfonts}
\usepackage{blindtext}
\hypersetup{pdfauthor={Name}}
\usepackage{appendix}
\usepackage[authoryear]{natbib}

\usepackage{booktabs} 
\usepackage{stackengine} 
\newcommand\oast{\stackMath\mathbin{\stackinset{c}{0ex}{c}{0ex}{\ast}{\bigcirc}}}

\renewcommand{\cite}[1]{\citep{#1}}

\newcommand{\STAB}[1]{\begin{tabular}{@{}c@{}}#1\end{tabular}}

\newcommand{\midsepremove}{\aboverulesep = 0mm \belowrulesep = 0mm}
\newcommand{\midsepdefault}{\aboverulesep = 0.605mm \belowrulesep = 0.984mm}

\begin{document}

\title{Climate Adaptation-Aware Flood Prediction for Coastal Cities Using Deep Learning}

\Author[1][bilal.hassan@nyu.edu]{Bilal}{Hassan} %% correspondence author
\author[1]{Areg Karapetyan}
\author[1]{Aaron Chung Hin Chow}
\author[1]{Samer Madanat}

\affil[1]{Division of Engineering, New York University Abu Dhabi, Abu Dhabi, United Arab Emirates}

\runningtitle{TEXT}

\runningauthor{TEXT}

\received{}
\pubdiscuss{} 
\revised{}
\accepted{}
\published{}

\firstpage{1}

\maketitle

\begin{abstract}
Climate change and sea-level rise (SLR) pose escalating threats to coastal cities, intensifying the need for efficient and accurate methods to predict potential flood hazards. Traditional physics-based hydrodynamic simulators, although precise, are computationally expensive and impractical for city-scale coastal planning applications. Deep Learning (DL) techniques offer promising alternatives, however, they are often constrained by challenges such as data scarcity and high-dimensional output requirements. Leveraging a recently proposed vision-based, low-resource DL framework, we develop a novel, lightweight Convolutional Neural Network (CNN)-based model designed to predict coastal flooding under variable SLR projections and shoreline adaptation scenarios. Furthermore, we demonstrate the ability of the model to generalize across diverse geographical contexts by utilizing datasets from two distinct regions: Abu Dhabi and San Francisco. Our findings demonstrate that the proposed model significantly outperforms state-of-the-art methods, reducing the mean absolute error (MAE) in predicted flood depth maps on average by nearly 20\%. These results highlight the potential of our approach to serve as a scalable and practical tool for coastal flood management, empowering decision-makers to develop effective mitigation strategies in response to the growing impacts of climate change. \textbf{Project Page: https://caspiannet.github.io/}
\end{abstract}

%% main text
\section{Introduction}

Of the world's $34$ megacities (i.e., those with more than 10 million inhabitants), approximately $70$\%\footnote{Calculated from the data reported in \href{https://digitallibrary.un.org/record/4065171?v=pdf}{United Nations’ World Cities Report 2024}.} are situated on or near the coast. Coastal cities, including megacities, host nearly 10\% of the world's population, are 2.6 times denser populated as compared to inland areas, and are powerhouses of global trade and business activities (the legacy of maritime trade)~\cite{annurev112320-101903}. Yet, these cities, especially those in low-elevation coastal zones, are hotspots for climate-induced disasters. Notable examples range from Venice, Italy, and Miami, Florida, to Manila, Philippines~\cite{van2024sea,griggs2021coastal}. In fact, according to a 2021 article by UN-Habitat\footnote{\href{https://unhabitat.org/news/18-nov-2021/busan-un-habitat-and-oceanix-set-to-build-the-worlds-first-sustainable-floating}{UN-Habitat Press Release}}, 90\% of megacities are vulnerable to sea level rise (SLR) and, as analyzed in~\cite{Hallegatte2013}, the flood risk to coastal cities is expected to rise nine-fold by 2050. The problem is compounded by land subsidence, confronting coastal communities with the challenge of managing multiple, interacting sources of risk~\cite{CAO202187, ardha2024flood, Barnard2024}. 

To address the threat of coastal flooding, planners typically consider a portfolio of adaptation strategies, which are broadly classified as measures to protect the shoreline, accommodate rising waters, retreat from vulnerable areas, or avoid new development in hazard zones~\cite{intronew, IPCC_2022}. This study focuses on \textit{protect} strategies, which involve armoring shorelines with engineered structures such as seawalls, levees, and storm barriers ~\cite{beagle2019san, papacharalambous2013greaterNOLA}. Rather than evaluating the performance of specific engineering designs, our work addresses the more fundamental strategic question of the optimal spatial configuration of these defenses, that is, determining which shoreline segments are most critical to protect. A recent example of a large-scale protection effort involves New York City, where extensive flood risk modeling led to the fortification of a two-and-a-half mile stretch of Lower Manhattan’s shoreline, an intervention largely driven by the aftermath of Hurricane Sandy in 2012~\cite{andrew}. Construction of these coastal defense structures, however, significantly alters the shoreline geometry, subsequently influencing the local hydrodynamics and potentially creating regional impacts~\cite{hummel2021economic,wang2018influence,haigh2020tides}.

Therefore, for effective and responsible flood protection planning, it is essential to account for plausible hydrodynamic changes due to shoreline modifications. To this end, physics-based high-fidelity simulators, such as Delft3D~\cite{delft}, can be employed to resolve the detailed hydrodynamics. While these tools can simulate complex coastal processes with detailed accuracy, they are computationally expensive, often requiring days to simulate a single shoreline protection scenario~\cite{kyprioti2021storm,rohmer2023improved, aregkarapetyan2024deep}. This computational burden limits their practicality for planning routines where multiple simulations are needed~\cite{chen2024urban,du2024urban}. Furthermore, in complex urban terrain, flood dynamics are critically dependent on fine-scale features such as individual buildings, road elevations, and flood barriers. Accurately capturing these localized effects, which can determine whether critical infrastructure is inundated, necessitates high-resolution modeling~\cite{cint1,cint2,aregkarapetyan2024deep}.

In response to these limitations, data-driven methods, including machine learning (ML) and deep learning (DL) techniques, have emerged as promising alternatives for rapid flood prediction~\cite{hess3,mosavi2018flood,hess1,zuhairi2022review, hess2,hess4,hess5}. Numerous studies have employed traditional ML methods like random forests and support vector machines~\cite{mosavi2018flood,ali2022flood}, and more recently, hybrid models that combine ML with hydrodynamic simulations have gained favor~\cite{chen2024urban,du2024urban}. Compared to ML, DL algorithms have shown enhanced capabilities. While one-dimensional (1D) models like long short-term memory (LSTM) are effective for sequential forecasting, two-dimensional (2D) methods such as Convolutional Neural Networks (CNNs) are particularly well-suited for capturing the complex spatial patterns inherent in flood maps. These surrogate models aim to emulate high-fidelity simulators by learning complex input-output relationships without explicitly modeling the underlying physical processes~\cite{aregkarapetyan2024deep}.

However, despite these advancements, a significant challenge remains. Many existing flood prediction studies focus on singular triggers or short-term extreme events and, consequently, do not jointly consider the complex, long-term impacts of both SLR and dynamic shoreline adaptation strategies~\cite{jia2016surrogate,guo2021data}. Fulfilling the high-resolution modeling requirement with DL-based models also presents its own complications, chief among them being \textit{data scarcity} and the challenge of handling the \textit{high dimensionality of the output} (predicting an inundation value for every pixel in a large spatial grid)~\cite{kyprioti2021storm,rohmer2023improved, aregkarapetyan2024deep}. Previously, a 2D DL framework was introduced to address these challenges by recasting flood prediction as a computer vision task~\cite{aregkarapetyan2024deep}. The core of that approach was to transform discrete shoreline protection scenarios (a list of protected or unprotected segments) into 2D spatial input maps. By treating the problem as an image-to-image translation task, that framework allowed CNNs to inherently learn the geometric relationships between protected areas and the resulting flood patterns. Crucially, that image-based format also enabled the use of random cutouts data augmentation techniques to artificially expand the limited training dataset, a critical advantage in data-scarce domains. While this foundational work demonstrated the viability of the approach, their method was limited to a single location and a particular SLR scenario. Taking a step further, this work introduces a novel DL model designed to generalize across two distinct coastal regions and multiple SLR scenarios. More concretely, the key contributions of this study are as follows:

\begin{enumerate} 
    \item We propose a novel DL model (\textit{CASPIAN-v2}) designed to accurately predict high-resolution coastal flooding under various SLR scenarios and shoreline adaptation strategies. The architecture is developed as a lightweight CNN for fast and scalable prediction, aiming to significantly reduce computational time compared to  traditional high-fidelity hydrodynamic models  while maintaining high accuracy.

    \item We present two new, comprehensive datasets from vulnerable coastal cities, Abu Dhabi (AD) and San Francisco (SF). These datasets cover different sea-level rise scenarios and shoreline adaptations to facilitate future research in this domain.

    \item We conduct a rigorous evaluation of the proposed framework against state-of-the-art (SOTA) ML and DL models to benchmark its performance and test its generalization capabilities across diverse scenarios.

    \item We employ explainable artificial intelligence (AI) techniques to validate the outputs of the model, assess the physical plausibility of its predictions, and offer interpretability to support decision-making in flood risk assessment.
\end{enumerate}

Put together, these contributions can assist urban policymakers in designing more effective and reliable coastal protection programs. Additionally, we open-source the code and datasets in the hope of facilitating further research and attracting greater attention to this problem within the machine learning community.

\section{Study Area and Data Description}\label{Study}

In this research, we examine two vulnerable metropolitan coastal areas (Abu Dhabi and San Francisco Bay) to predict coastal flooding under various SLR and shoreline protection. Both locations feature low-lying topographies and significant urbanization, making them particularly susceptible not only to direct flooding, but also impacts on transportation links, and specifically whether important arterials such as shoreline highways or freeways will be flooded due to SLR. Our aim is to evaluate the effectiveness and applicability of the DL-based solution for forecasting inundation in these regions.

\subsection{Study Area Description}

\subsubsection{Abu Dhabi}
Abu Dhabi, located along the southern coast of the Arabian (Persian) Gulf, faces rising flood risks from climate change-induced SLR, tidal flooding, and storm surges driven by extreme winds such as Shamal \cite{Langodan_Shamal_2023116158}. Projections estimate that a 0.5~m SLR, expected by 2050–2100 based on IPCC AR6 \cite{IPCC_2021climate}, could inundate critical ecosystems like mangroves and artificial islands, potentially doubling flood zones when accounting for wind and wave action \cite{melville2021roadmap}. The shallow bathymetry of the region amplifies these risks, with even minor sea-level increases threatening key infrastructure and densely populated areas, where over 85\% of the population and 90\% of the infrastructure lie just meters above sea level \cite{al2019sea, melville2021roadmap}.

To assess flood risks, we divided the AD urban coastline into 17 operational landscape units (OLU), based on the Abu Dhabi Urban Structure Framework Plan 2030 \cite{ad-plan2030}, and adopted in previous studies e.g., \cite{aaronchow2022combining}. In the hydrodynamic model, the protection of a single OLU involves placing an impermeable seawall (that assumes no overtopping) along the coastal boundary of the OLU.  

\begin{figure}[htbp]  
\centering
\includegraphics[width=0.8\linewidth]{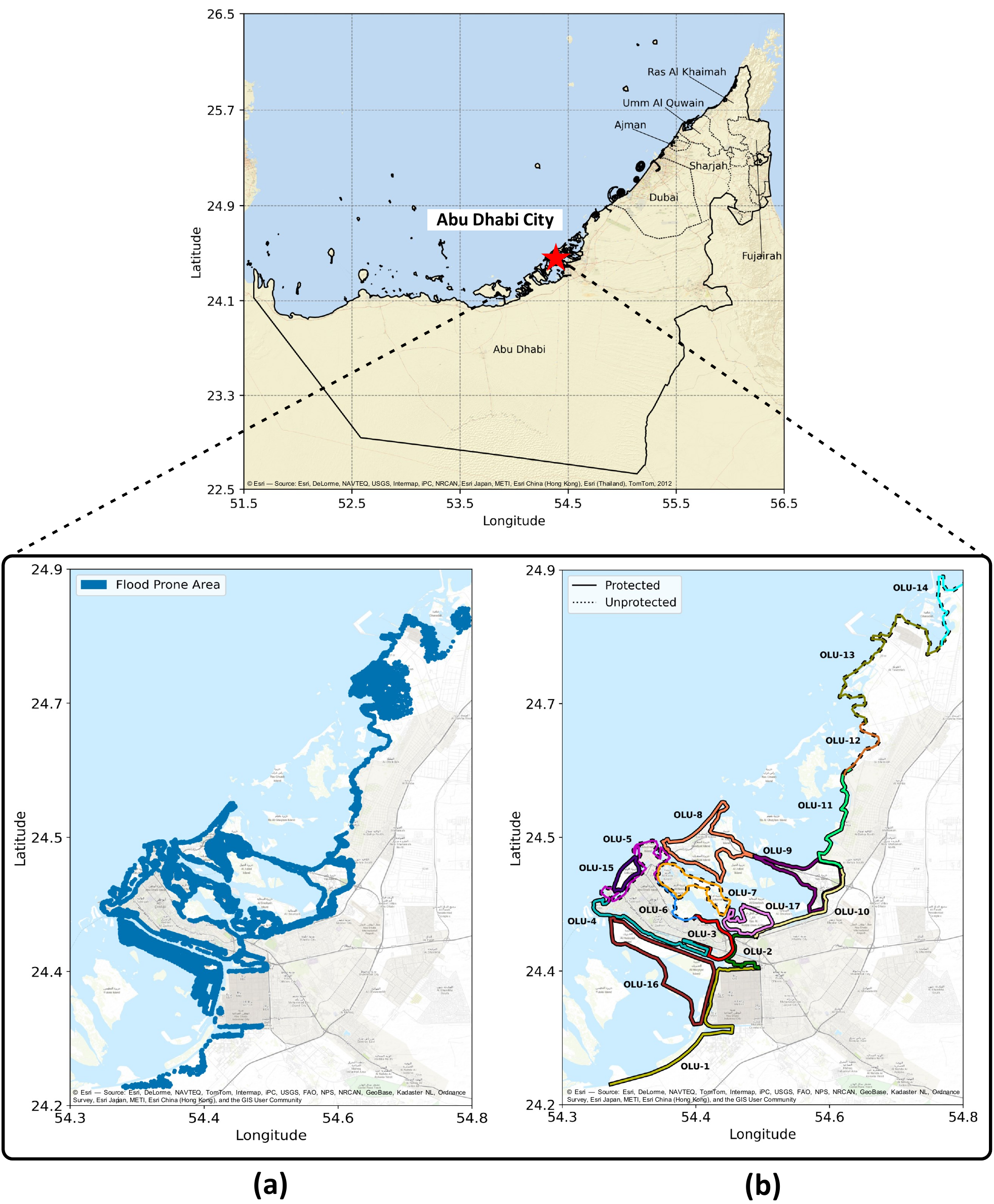}
\caption{AD study area shown on the map of the United Arab Emirates. (a) All areas susceptible to flooding under a 0.5~m SLR scenario without any shoreline protections (b) The 17 OLUs defined along the AD shoreline where protections are to be tested for their effectiveness.
}
\label{fig:AD_OLU}
\end{figure}

This framework captures unique features of both natural ecosystems and urban zones, enabling detailed flood vulnerability analyses under various shoreline adaptations. Figure~\ref{fig:AD_OLU} illustrates the AD coastline, the OLU divisions, and the inundation points for the 0.5~m SLR scenario.

%%%%%%%%%%%%%%%%%%%%%%%%%%%%%%%%
%%%%%%%%%%%%%%%%%%%%%%%%%%%%%%%%

%%%%%%%%%%%%%%%%%%%%%%%%
\begin{figure}[htbp]  
\centering
\includegraphics[width=0.8\linewidth]{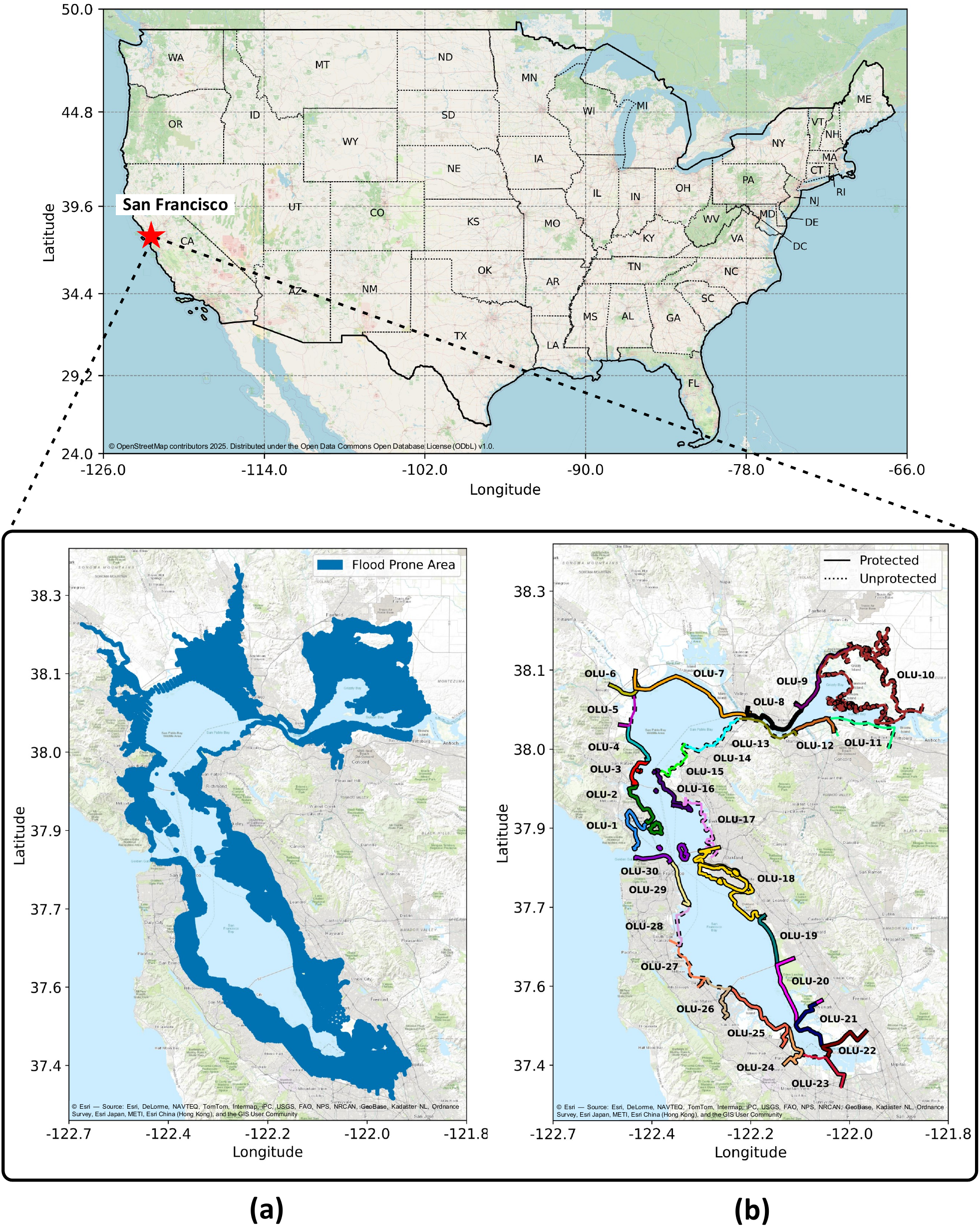}
\caption{SF Bay study area shown on the map of the United States (a) All areas susceptible to flooding under a 0.5~m SLR scenario without any shoreline protections (b) The 30 OLUs defined along the SF Bay shoreline where protections are tested for their effectiveness.
}
\label{fig:SF_OLU}
\end{figure}
%%%%%%%%%%%%%%%%%%%%%%%%%%%%%%%%

\subsubsection{San Francisco Bay Area}
Our second study area is the urban shoreline located along the banks of San Francisco Bay (Figure~\ref{fig:SF_OLU}). Owing to the location of San Francisco Bay as an inland bay, its shoreline communities are relatively sheltered from storm surges by the exterior Californian coastline, with mean significant wave heights within the Bay at about 0.07-0.2~m, in contrast to 2.0-3.0~m at Point Reyes located on the California coast outside the Bay \cite{usgs-sfbay}.

San Francisco Bay faces significant flood risks from SLR and tidal variability, which are exacerbated by climate change \cite{CEC_2018_SFBay,wang2018influence}, which in turn impacts low-lying urban zones, transportation networks, and hydrological systems (such as the Napa River Basin). However, our focus in this study is on tidal flooding within San Francisco Bay in order to highlight the unique tidal behavior within the Bay where the construction of sea walls along certain portions of the shoreline Bay may, in fact, exacerbate the sea level within the Bay to increase by up to 1~m  \cite{holleman2014coupling}. 

For San Francisco Bay, the discretization of coastline of the Bay Area into 30 OLUs was based on shoreline morphology, hydrology, and urban infrastructure, originally performed by  ~\cite{beagle2019san}, and used in previous studies  \cite{hummel2021economic, Jiayunsun2020multimodal}. Figure~\ref{fig:SF_OLU} illustrates the 30 OLUS for SF Bay Area, the OLU divisions, and the inundation points for the 0.5~m SLR scenario. In the hydrodynamic model, the protection of a single OLU involves placing an impermeable seawall (that assumes no overtopping) along the coastal boundary of the OLU. 

\subsection{Data Sources and Hydrodynamic Simulations} \label{sec:DataAndSims}
The ground truth flood data used for training and evaluating our surrogate model was generated through a series of physics-based hydrodynamic simulations using the Delft3D model. This model integrates key physical processes including SLR and tidal dynamics (see Supplementary Material, Section S1). High-resolution bathymetry and digital elevation models (DEM) (with data sources such as TanDEM-X, Landsat-8, and Nautical Charts) were used for both regions to ensure accurate modeling of coastal topography that transitions smoothly between sea and the land. While some authors \cite{dealmeida_2013applicability, neal2012subgrid, li_2019modeling, sanders2019primo_inundationmodel, nithila2024novel} address subgrid details by using separate subgrid nesting methods, we have retained the same governing equations but used a 30~m model grid in the areas of interest, and Delft3D is capable of automatically modeling wetting and drying of grid cells from one time step to the next.

The accuracy and reliability of these physics-based models were established through rigorous validation against real-world observations. For San Francisco Bay, the Delft3D model was adapted from the CoSMoS model originally developed by \cite{barnard2014development} and adapted to San Francisco Bay by \cite{wang2017interactions}, and validated in the past using tidal gages at 9 tidal gage locations in and around San Francisco Bay. Pearson correlation coefficients ranged from 0.9862 to 0.9996, while the root mean square (RMS) ratios (the ratio of modeled versus measured RMS amplitudes) ranged from 0.973 to 1.027 (please refer to \cite{wang2017interactions})

For Abu Dhabi, the Delft3D model was validated using water level data from 196 tidal gage locations throughout the Gulf (as the hydrodynamic model encompassed the entire Gulf in addition to the western portions of the Gulf of Oman). The water levels at these locations were compared with one month’s worth of hydrodynamic simulation, and the resulting absolute root mean square error (RMSE) values ranged from 0.0013 to 0.0043 m in the vicinity of Abu Dhabi. More validation details for Abu Dhabi can be found in \cite{aaronchow2022combining}. Given this strong validation, the outputs of the hydrodynamic simulations were considered a reliable proxy for ground truth for the purposes of training and evaluating our deep learning framework.

While the Gulf does not typically experience tropical cyclones, it is known for its northwesterly winds generally occurring with winds at about 20 m/s with sudden onset and sustained over a period of up to 3-5 days. These are called the Shamal winds (meaning “North” in Arabic) and occur at least 10 times annually, mainly during the winter months \cite{alsenafi2015shamals,  li2020physical-arabiangulf}. Accordingly, for Abu Dhabi, we applied a nested SWAN model to simulate wind and wave effects, particularly the impact of these Shamal winds, which can significantly intensify tidal flooding risks. Both the SWAN model and Delft3D models were forced using ERA5 meteorological data in the Gulf. 

In both geographic locations, our aim was to generate data that correspond to a hypothetical future extreme flooding scenario, where there was little to no flooding observed without SLR. For AD, simulations were based on a 0.5~m SLR scenario, consistent with regional projections for mid-century SLR (as described above) \cite{IPCC_2021climate}. The 0.5~m SLR scenario was then coupled with storm surges resulting from a sample 3-month long Shamal event. In contrast, flood simulations for the SF Bay Area were conducted under three SLR scenarios: 0.5~m, 1.0~m, and 1.5~m,  which reflects a possible future scenario for San Francisco Bay in the year (somewhere between 2050-2100 depending on the climate change scenario pathway (between SSP2-4.5 and SSP5-8.5) from IPCC AR6 report \cite{IPCC_2021climate}. Table~\ref{Table_1} provides a comprehensive overview of the datasets generated for this study, which are partitioned into three categories based on their purpose. The \textit{Main Set}, comprising the largest datasets from AD (0.5~m SLR) and SF (1.0~m SLR), was used for the primary training, validation, and testing of the CASPIAN-v2 model. The \textit{Holdout Set} consists of scenarios intentionally curated to be challenging (such as protecting one entire side of the SF Bay while leaving the other exposed) and was used for blind testing of the primarily trained model's performance on complex spatial schemes not seen during training (see Supplementary Material Section S4). Finally, the \textit{Generalizability Set} includes SF scenarios at different SLR levels (0.5~m and 1.5~m) and was used exclusively to evaluate the ability of the model to adapt to new environmental conditions via fine-tuning.

To balance the need to model a larger number of modeled tidal cycles per simulation, with the computational time and storage space used for the simulations, a 3-month simulation period was also applied for San Francisco Bay. Although our San Francisco model includes riverine input from the Sacramento and San Joaquin Rivers, the inflow rates into the Bay were baseline values rather than for extreme fluvial flood events. While we acknowledge that incorporating more hydrodynamic forcing conditions to include pluvial and riverine floods, as well as extreme storm events, can refine the hydrodynamic model to reflect more extreme flooding, our overall scope in this paper is in the use of machine learning to be able to act as a surrogate for a hydrodynamic model running under different SLR scenarios. The detailed protocols for how these datasets were split and used are described in Section~\ref{datasetsplit}.

\begin{table}[htb]
\renewcommand{\arraystretch}{1.15}
\centering
\footnotesize
\begin{threeparttable} % Wrap the entire table in threeparttable
\caption{\label{Table_1}Dataset details for AD and SF regions, including OLUs, SLR depths, and the number of unique shoreline protection scenarios. The Main Set was used for primary model training and testing. The Holdout Set was used for blind testing on challenging scenarios. The Generalizability Set was used to evaluate model adaptability to new SLR conditions via fine-tuning.}
\begin{tabular}{c| c c c}
\hline
\textbf{Region} & \textbf{OLUs} & \textbf{SLR} & \textbf{Protection Scenarios} \\
\hline
\multirow{4}{*}{\textbf{AD}} & \multirow{4}{*}{\parbox[l]{275pt}{\textbf{17 OLUs:}  1 \textit{(Mussafah)},  2 \textit{(Bain Al Jesrain)},  3 \textit{(Grand Mosque District)},  4 \textit{(AD Island West)},  5 \textit{(Marina, CBD, Al Mina)},  6 \textit{(AD Island East)},  7 \textit{(Al Reem Island)},  8 \textit{(Saadiyat Island)},  9 \textit{(Yas Island)},  10 \textit{(Al Raha Island)},  11 \textit{(Al Shahama)},  12 \textit{(Al Rahba)},  13 \textit{(New Port City)},  14 \textit{(Ghantoot)},  15 \textit{(Lulu Island)},  16 \textit{(Hudayriat Island)},  17 \textit{(Inner Islands)}}}& \multirow{4}{*}{0.5~m} & \\& & & 142 (Main Set)\\
& & & 32 (Holdout Set) \\
& & &  \\
\hline
\multirow{7}{*}{\textbf{SF}} 
& \multirow{7}{*}{\parbox[l]{275pt}{\textbf{30 OLUs:}  1 \textit{(Richardson)},  2 \textit{(Corte Madera)},  3 \textit{(San Rafael)},  4 \textit{(Gallinas)},  5 \textit{(Novato)},  6 \textit{(Petaluma)},  7 \textit{(Napa - Sonoma)},  8 \textit{(Carquinez North)},  9 \textit{(Suisun Slough)},  10 \textit{(Montezuma Slough)},  11 \textit{(Bay Point)},  12 \textit{(Walnut)},  13 \textit{(Carquinez South)},  14 \textit{(Pinole)},  15 \textit{(Wildcat)},  16 \textit{(Point Richmond)},  17 \textit{(East Bay Crescent)},  18 \textit{(San Leandro)},  19 \textit{(San Lorenzo)},  20 \textit{(Alameda Creek)},  21 \textit{(Mowry)},  22 \textit{(Santa Clara Valley)},  23 \textit{(Stevens)},  24 \textit{(San Francisquito)},  25 \textit{(Belmont - Redwood)},  26 \textit{(San Mateo)},  27 \textit{(Colma - San Bruno)},  28 \textit{(Yosemite - Visitacion)},  29 \textit{(Mission - Islais)},  30 \textit{(Golden Gate)}}} & & \\
& & \multirow{2}{*}{1.0~m} & 285 (Main Set) \\
& & & 46 (Holdout Set) \\
& & 0.5~m & 32 (Generalizability Set) \\
& & 1.5~m & 32 (Generalizability Set) \\
& & &  \\
& & &  \\
\hline
\end{tabular}
\end{threeparttable}
\end{table}

We ran individual Delft3D scenarios (each with a 3-month simulation time as described above) to collect hourly inland inundation data under different coastal protection scenarios to create a dataset for training and validating our DL model. Our findings highlight the importance of holistic regional flood control measures, especially given the intricate interplay between protected and unprotected zones. Further, the datasets from two regions allowed us to assess the applicability and reliability of the DL model in different vulnerable coastal settings.

The computational cost of generating a peak flood depth map using the coupled hydrodynamic model, which underscores the need for an efficient surrogate, varies significantly between the two study regions. For the coast of Abu Dhabi, the process to generate a map such as the one shown in Fig. \ref{fig:AD_OLU}(a) takes approximately 71 to 73 hours of elapsed runtime, equating to 1500 to 1660 CPU-hours, depending on the specific protection scenario. This comprehensive simulation includes Delft3D runs, which require 6 to 7 hours on 28 CPU cores (Intel Xeon E5-2680 @ 2.40 GHz; ≈ 168–196 CPU-hours), and SWAN simulations, which take about 10 to 11 hours on 128 CPU cores (AMD EPYC 7742 @ 2.25GHz; ≈ 1280–1408 CPU-hours). Subsequent post-processing and run-up calculations using Matlab scripts add approximately 55 hours on a single core. In contrast, generating a similar map for San Francisco Bay (see Fig. \ref{fig:SF_OLU}(a)) is computationally less demanding, requiring approximately 3.5 to 6.0 hours of elapsed time, or 84.5 to 141 CPU-hours. The Delft3D runs for this region take about 3 to 5 hours on 28 CPU cores, and the post-processing of these outputs takes between 0.5 and 1.0 hours on a single core. It is important to note that SWAN and run-up calculations were not performed for the San Francisco Bay shoreline, as its relatively sheltered inland location makes these components unnecessary, accounting for the substantial difference in computational cost.

\subsection{Data Preprocessing}
The raw, tabular data generated by the Delft3D simulator, which consists of inundation coordinates and corresponding peak water level (PWL) values, is not directly compatible with our 2D DL model. Therefore, a multi-step preprocessing pipeline was developed to transform this data into a structured grid format suitable for a computer vision task.

The first key step was to map the inundation coordinates onto a standardized $1024 \times 1024$ spatial grid. This was achieved by defining the grid boundaries based on the maximum spatial extent of all simulation data and then assigning each inundation point to its nearest grid cell. In cases where multiple inundation points mapped to the same cell due to the high density of the data, a conflict resolution strategy was employed that reassigned the conflicting points to the nearest available empty cell, ensuring a unique one-to-one mapping.

Subsequently, we incorporated the shoreline protection information. For each inundation point, we calculated its proximity to the nearest protected and unprotected OLUs and assigned it a class based on which was closer. This classification, along with the PWL values, was then used to construct the final input and output matrices for training. The shoreline protection scenarios were encoded  as binary strings, where '0' indicates unprotected OLUs and '1' denotes protected OLUs. This entire process ensures that the model receives spatially coherent input that encodes not just water levels, but also the crucial context of shoreline defense configurations. A full, detailed breakdown of each step, including the mathematical formulations for grid mapping and OLU classification, is provided in the Supplementary Material (Section S2).

%%%%%%%%%%%%%%%%%%%%%%
%%%%%%%%%%%%%%%%%%%%%%
\section{Method}

This section details the proposed deep learning framework for predicting coastal inundation under various SLR depths and shoreline protection scenarios. We first provide a high-level overview of the end-to-end workflow, from data generation to prediction, and then present the specific architecture of the CASPIAN-v2 model and the novel hybrid loss function used for its training.

\subsection{Proposed Framework}

\begin{figure*}[htbp]  
\centering
\includegraphics[width=1\linewidth]{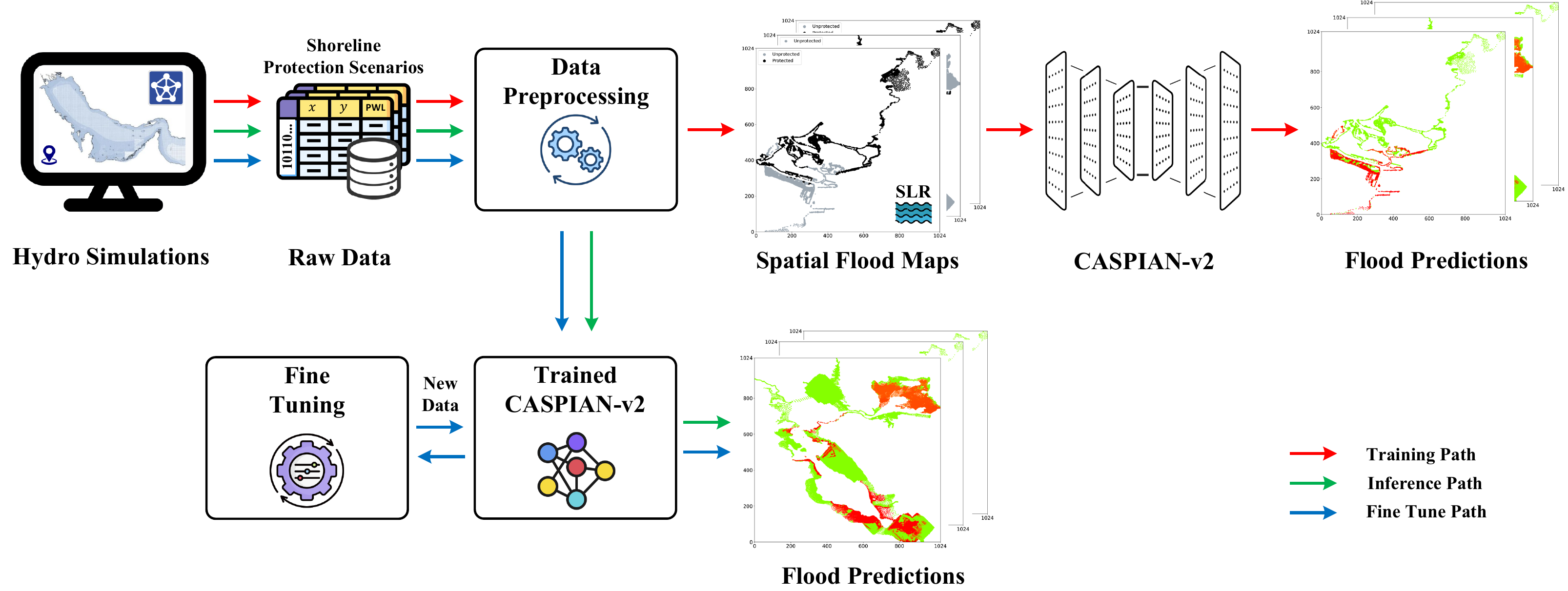}
\caption{An overview of the proposed framework for coastal flood prediction. It begins with hydrodynamic simulations based on SLR data and coastal protection scenarios to generate raw flood data, which is then processed into spatial flood maps. The CASPIAN-v2 model, trained on these maps, predicts inundation patterns and flood extent. The framework can be fine-tuned with new data for improved adaptability. The different colored paths represent training (red), inference (green), and fine-tuning (blue) stages.}
\label{fig:FW}
\end{figure*}

The proposed framework, illustrated in Figure~\ref{fig:FW}, provides an end-to-end pipeline for generating, processing, and predicting coastal flood data. The process is organized into several key stages, each represented by a colored path in the diagram.

\begin{itemize}
    \item \textbf{Data Generation and Preprocessing:} The process begins with running physics-based hydrodynamic simulations (e.g., Delft3D) using different shoreline protection scenarios and SLR levels as inputs. This generates raw, tabular flood data containing water levels at specific coordinates. This raw data is then put through a preprocessing pipeline, where it is transformed into 2D spatial flood maps suitable for a computer vision approach.

    \item \textbf{Training Path (Red):} The preprocessed spatial maps serve as the input-output pairs for training the CASPIAN-v2 model. The model learns the complex, non-linear relationships between the shoreline protection configurations (input) and the resulting flood inundation patterns (output).

    \item \textbf{Inference Path (Green):} Once trained, the model can be used for rapid inference. Given a new, unseen shoreline protection scenario, the model can predict the corresponding high-resolution flood map in a matter of seconds, bypassing the need for computationally expensive hydrodynamic simulations.

    \item \textbf{Fine-Tuning Path (Blue):} To enhance adaptability, the trained CASPIAN-v2 model can be fine-tuned on new data. This is particularly useful for adapting the model to different SLR scenarios or geographical regions for which only limited data might be available, allowing for efficient knowledge transfer without retraining from scratch.
\end{itemize}

This integrated framework provides a scalable and efficient solution for assessing the impact of diverse coastal adaptation strategies under the threat of climate change.

%%%%%%%%%%%%%%%%%%%%%%%%
\begin{figure*}[htbp]  
\centering
\includegraphics[width=1\linewidth]{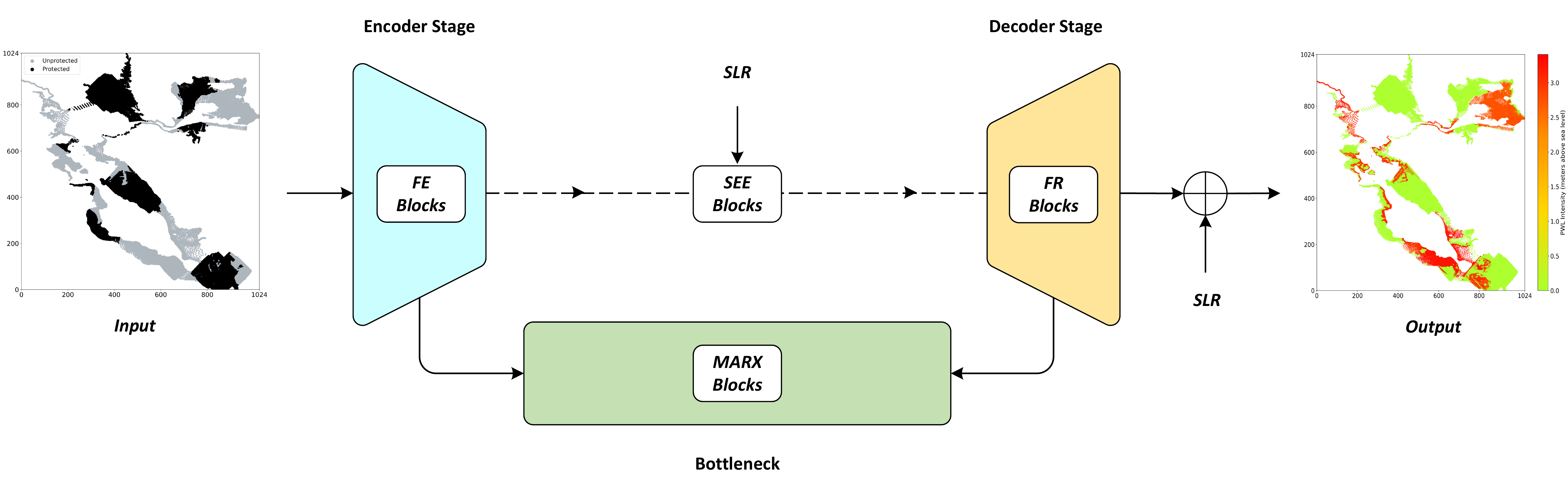}
\caption{A simplified schematic of the CASPIAN-v2 model architecture. The model consists of an encoder Stage that uses feature extraction (FE) blocks to create a compressed representation of the input map. At the Bottleneck, a series of multi-attention ResNeXt (MARX) blocks refine these features. The decoder stage then uses feature reconstruction (FR) blocks to generate the high-resolution output flood map. Crucially, SLR data is integrated into the decoder via the SLR-enhanced encoding (SEE) blocks and again before the final output, allowing the model to produce predictions conditioned on different climate scenarios.}
\label{fig:ModelArch}
\end{figure*}
%%%%%%%%%%%%%%%%%%%%%%%%%%%%%%%%

\subsection{CASPIAN-v2 Architecture}
The \textit{CASPIAN-v2} model improves and extends the capabilities of the previously developed CNN architecture to predict coastal flooding \cite{aregkarapetyan2024deep}. Unlike the previous version, CASPIAN-v2 integrates SLR data and has a more robust yet minimalistic architecture that generalizes across various geographical regions. Figure \ref{fig:ModelArch} illustrates the CASPIAN-v2 architecture, which consists of three main stages: encoder, bottleneck, and decoder. The following subsections provide a conceptual overview of the architecture, whereas a detailed exposition of all network layers and operations is presented in the Supplementary Material.

\subsubsection{Encoder Stage}

The encoder consists of a sequence of convolutional \textit{feature extraction} (FE) blocks that progressively reduce the spatial resolution of the input grid while increasing the depth of the feature maps. This hierarchical feature extraction allows the model to capture multi-scale patterns essential for accurate flood inundation prediction. Each FE block uses depthwise separable convolutions and pooling to condense the input feature maps, followed by pointwise convolutions that expand the feature depth. Moreover, residual skip connections are incorporated to preserve important spatial information and mitigate gradient vanishing, ensuring that critical low-level features are not lost. By the end of the encoder stage, the input grid is transformed into a concise feature representation, encapsulating both localized details, such as inundation patterns in specific regions, as well as a broader spatial context. It should be noted that the scalar SLR input is not passed through the encoder; instead, it is directly incorporated in the decoder stage to globally influence the reconstruction of flood patterns.

\subsubsection{Bottleneck Stage}

CASPIAN-v2 employs a novel multi-attention ResNeXt (MARX) block at the bottleneck (the deepest part of the network with the smallest spatial dimensions) to refine and enrich the encoded features. The MARX block incorporates ResNeXt blocks \cite{xie2017aggregatedRESNEXT}, an aggregated residual structure, alongside the convolutional block attention module (CBAM) \cite{woo2018cbam}, facilitating the model in concentrating on key features. Specifically, the encoded feature map is first processed by a residual block, then passed through a attention module which sequentially applies channel attention and spatial attention to reweight the feature map, and finally routed through a second residual block. This combination adaptively emphasizes critical features in both the channel and spatial dimensions, thereby enhancing the ability of the model to learn complex flood patterns under various scenarios. The output of the MARX block is a rich high-level representation of the input scenario that serves as input to the decoder.

\subsubsection{Decoder Stage}

The decoder stage progressively reconstructs the high-resolution flood inundation map through a sequence of \textit{Feature Reconstruction (FR)} blocks. Each block upsamples features using a transpose convolution before fusing them with the corresponding encoder output. This fusion via skip connections is crucial, as it serves to reintroduce fine-scale spatial details that were compressed during encoding. A key enhancement in CASPIAN-v2 is the incorporation of the SLR input into the decoder through a specialized \textit{SLR-Enhanced Encoding (SEE)} block. The SEE mechanism uses the scalar SLR value to modulate decoder features, effectively guiding the upsampling process with global sea-level context. In practice, the SEE block learns a set of weighting coefficients from the encoder’s pooled features and the SLR value, which are then applied to the decoder feature maps at each scale. Consequently, regions more susceptible to flooding under a given SLR scenario receive higher weights during reconstruction. After the final upsampling, a convolutional layer produces an initial output grid, which is further refined by adding back the SLR-weighted summed features from the last decoder layer before applying the final activation function. The resulting output is the predicted flood inundation map, where each cell reflects the likelihood or extent of flooding at that location given the input conditions.

\subsection{Loss Function}

Predicting PWL under different SLR scenarios is challenging due to outliers and the need to balance error sensitivity across multiple regions. To tackle these issues, we introduce a hybrid loss function that combines Huber~\cite{HUBER}, Log-Cosh~\cite{LogCosh}, and Quantile~\cite{Quantile} losses in a weighted setup. The Huber loss \( L_{h} \) aims to robustly minimize small prediction errors while limiting the impact of outliers, and it uses a threshold \(\delta\) to manage the sensitivity of the error. The \( L_{h} \) for each sample \(i\) is computed as expressed in Eq.~(\ref{Eq11}):

\begin{align}
L_{h, i} = 
\begin{cases} 
\frac{1}{2} (y_{\text{p}, i} - y_{\text{t}, i})^2 & \text{if } |y_{\text{p}, i} - y_{\text{t}, i}| \leq \delta, \\
\delta \cdot |y_{\text{p}, i} - y_{\text{t}, i}| - \frac{1}{2} \delta^2 & \text{otherwise}
\end{cases}
\label{Eq11}
\end{align}
where \( y_{\text{t}, i} \) and \( y_{\text{p}, i} \) represent the actual and estimated PWL values. We set \( \delta \) within the range of 0.3 and 0.7, which is dynamically determined to balance sensitivity and robustness. Moreover, we integrate Log-Cosh loss (\( L_{\text{cosh}} \)) to smooth gradients in regions with large variations, helping to maintain prediction stability in different areas affected by SLR. The \( L_{\text{cosh}} \) is expressed as in Eq. (\ref{Eq12}):

\begin{align}
L_{\text{cosh}, i} = \log\left(\cosh(y_{\text{p}, i} - y_{\text{t}, i})\right),
\label{Eq12}
\end{align}

In addition, the quantile loss \( L_{q} \) differentiates errors by assigning distinct penalties to underestimation and overestimation, dictated by a quantile parameter \( \tau = 0.75 \). This loss dynamically adjusts to minimize quantile-specific errors, calculated as in Eq. (\ref{Eq13}):

\begin{align}
L_{q, i} = 
\begin{cases} 
\tau \cdot (y_{\text{p}, i} - y_{\text{t}, i}) & \text{if } y_{\text{p}, i} \geq y_{\text{t}, i}, \\
(1 - \tau) \cdot (y_{\text{t}, i} - y_{\text{p}, i}) & \text{otherwise}
\end{cases}
\label{Eq13}
\end{align}

To achieve an optimal balance, we linearly combine the three loss components into a comprehensive hybrid loss function \( L_{\text{total}} \), weighted by empirically tuned coefficients \( \alpha_h, \alpha_c, \alpha_q\). The final loss is expressed as in Eq. (\ref{Eq14}):

\begin{align}
L_{\text{custom}} = \alpha_h \cdot L_{h} + \alpha_c \cdot L_{\text{cosh}} + \alpha_q \cdot L_{q},
\label{Eq14}
\end{align}
where \( \alpha_h, \alpha_c, \alpha_q \geq 0 \) and \( \alpha_h + \alpha_c + \alpha_q = 1 \). These weights are empirically determined to optimize predictive performance. By integrating these components, our custom hybrid loss function balances error sensitivity, maintains robustness to outliers, and addresses asymmetric error distributions, enhancing the model's predictive accuracy for PWL under varying SLR scenarios.

\section{Experimental Setup}
This section outlines the parameters employed to train, validate, and evaluate the proposed DL model. We detail the dataset splits, augmentation strategies, baseline models, and evaluation metrics to validate and compare the performance of the CASPIAN-v2 model.

\subsection{Dataset Splits} \label{datasetsplit}

Our research incorporates datasets from two regions (AD and SF), covering multiple SLR scenarios, as discussed in Section \ref{Study} and Table \ref{Table_1}. The data is divided into sets for primary model training and for subsequent fine-tuning to assess generalization. The composition of these datasets is detailed in Table~\ref{Table_2}.

%%%%%%%%%%%%%%%%%%%%%%%%%%%%
\begin{table}[htb]
\setlength{\tabcolsep}{8pt} 
\renewcommand{\arraystretch}{1.15}
\centering
\footnotesize
\begin{threeparttable} 
\caption{\label{Table_2}Dataset details for primary training and fine-tuning.}
\begin{tabular}{c c c c c c c}
\toprule
\textbf{Type} & \textbf{Region} & \textbf{SLR} & \textbf{Total} & \textbf{Train} & \textbf{Validation} & \textbf{Test} \\
\midrule
\multirow{2}{*}{Primary} & AD & 0.5 m & 142 & 96 & 10 & 36 \\
& SF & 1.0 m & 285 & 225 & 24 & 36 \\
\midrule
\multirow{2}{*}{Fine-tuning} &  SF & 0.5 m & 30 & 20 & 4 & 6 \\
& SF & 1.5 m & 30 & 20 & 4 & 6 \\
\bottomrule
\end{tabular}
\end{threeparttable}
\end{table}
%%%%%%%%%%%%%%%%%%%%%%%%%%%%

To enhance the model’s generalization ability and robustness for primary training, we employed a systematic data augmentation strategy on the AD (0.5 m) and SF (1.0 m) training and validation subsets. The augmentation process primarily involves a random remove function, which applies random spatial cutouts and scaling factors to the original samples. Specifically, this technique first identifies the spatial coordinates of the shoreline protection segments and then occludes small, square regions around a random subset of them in the input maps. This process simulates scenarios with imperfect or missing data, forcing the model to learn more robust contextual features rather than memorizing the impact of any single protection segment. 
We create distinct yet related variants of the original dataset by systematically applying these transformations multiple times (24$\times$ for AD and 10$\times$ for SF). Compared to the original sparse dataset, this strategy produces a richer dataset for primary training, comprising 2,304 training samples and 240 validation samples for AD, along with 2,250 training samples and 240 validation samples for SF. 

The fine-tuning datasets for SF (0.5 m and 1.5 m SLR) consist of 30 protection scenarios where one OLU was protected at a time (more details in Supplementary Material Section S5). For evaluation, 20\% of the data (6 samples) was reserved, while the remaining 80\% (24 samples) was used for fine-tuning and validation.

\subsection{Model Optimization and Training Protocol}
The CASPIAN-v2 model was implemented in Python 3.10 using TensorFlow 2.10.1 and was trained on a 64-bit Windows operating system. We utilized an Intel Core i9-14900K (3.20 GHz) machine with 64 GB of RAM and an NVIDIA GeForce RTX 4090 GPU. The final CASPIAN-v2 architecture was refined through extensive ablation studies that systematically evaluated the impact of each novel component. Key insights from these studies, which are detailed in Supplementary Material Section S3, are summarized in Table~\ref{tab:ablation_summary}. These experiments confirmed that the optimal design incorporates the custom \textit{Hybrid Loss} function and a bottleneck composed of four \textit{MARX blocks}. This bottleneck design (ResNeXt + CBAM) was empirically shown to be superior to simpler alternatives. Finally, the studies validated that our method of integrating SLR information via the \textit{SEE block} and just before the final output layer was the most effective approach.

%%%%%%%%%%%%%%%%%%%%%%%%%%%%
\begin{table}[htb]
\centering
\setlength{\tabcolsep}{18pt} 
\renewcommand{\arraystretch}{1.15}
\caption{Summary of ablation study results identifying the optimal configuration for each key model component. The final configuration of CASPIAN-v2 incorporates all these optimized choices.}
\label{tab:ablation_summary}
\begin{threeparttable}
\begin{tabular}{l l l}
\toprule
\textbf{Component} & \textbf{Optimal Configuration} & \textbf{Contribution} \\
\midrule
Bottleneck Architecture & MARX (ResNeXt+CBAM) & Superior spatial feature extraction \\
Loss Function & Hybrid & Balances error sensitivity and robustness \\
Number of MARX Blocks & 4 Blocks & Optimal balance of complexity and feature learning \\
Number of SEE Blocks & 1 Block\tnote{*} & Improves accuracy while maintaining efficiency \\
SLR Integration & SEE + Final Output Layer & Most effective placement for leveraging SLR data \\
\bottomrule
\end{tabular}
\begin{tablenotes}
    \item[*] Although 4 SEE blocks yielded the highest accuracy, 1 block was chosen for the final model to balance performance with computational efficiency, as detailed in the supplement.
\end{tablenotes}
\end{threeparttable}
\end{table}
%%%%%%%%%%%%%%%%%%%%%%%%%%%%

\subsubsection{Primary Training:} The model was first trained on the combined AD (0.5 m) and SF (1.0 m) datasets using the Adam optimizer and the proposed hybrid loss function. This phase lasted for 200 epochs with a batch size of 2, allowing the model to learn the core relationships between shoreline protection and flood dynamics. The remaining hyperparameters were fine-tuned using Bayesian Optimization and Random Search to ensure optimal performance.

\subsubsection{Fine-tuning for Generalization:} To assess adaptability to different SLR conditions, the pre-trained model was then fine-tuned on the new SF datasets (0.5 m and 1.5 m SLR). Fine-tuning spanned 100 epochs. To prevent catastrophic forgetting while adapting to the new data, we employed a curriculum-based strategy. This approach involved mixing the new SLR data with holdout data. The training began with batches containing 30\% new data and 70\% old data, with the proportion of new data gradually increasing to 70\% by the end of the fine-tuning process. The final performance on these new SLR levels was evaluated on the reserved test sets (6 samples each), which were not seen during either training or fine-tuning.

\subsection{Baseline Models}\label{baseline_models}
To ensure a fair and direct comparison, we selected and implemented a suite of SOTA models, as direct benchmarking against many methods in the literature is often not feasible due to a lack of publicly available code or differences in problem formulation. We assessed the performance of CASPIAN-v2 model for coastal flood prediction against several SOTA ML and DL techniques. We considered conventional ML methods, including the Naïve model, which utilizes a dummy regressor to forecast the mean value of the target variable to serve as a basic reference for assessing more advanced models. Additionally, we trained random forest, linear regression, extreme gradient boosting, support vector regression, lasso regression with polynomial features, and kriging with principal component analysis to establish an ML benchmark. The hyperparameters for training these models were optimized through a combination of Bayesian optimization and random search methods, allowing for efficient exploration of the parameter space while preventing overfitting on the validation set.

In addition to traditional ML baselines, we tested several DL models adapted to the flood prediction task. These include a simple feed-forward neural network architecture, specifically a multi-layer perceptron (MLP), and compact convolutional transformers (CCT) \cite{cct}, which serve as baseline 1D DL models. Furthermore, we evaluated several 2D DL models, including Attention-Unet \cite{attnunet}, and Swin-Unet \cite{cao2022swin}. To adapt these models for flood prediction, we replaced their segmentation heads with a $1\times1$ convolution layer followed by activation to output real-valued flood depth predictions. We evaluated two versions of Attention-Unet: one with randomly initialized weights and another (denoted as Atten-Unet*) with an encoder pre-trained on ImageNet \cite{deng2009imagenet}, leveraging transfer learning to improve performance in low-data scenarios. The final DL baseline was CASPIAN, which we previously proposed in \cite{aregkarapetyan2024deep}. All DL models were trained using the Adam optimizer and the proposed hybrid loss function ($L_{\text{custom}}$). Additionally, each model was trained for 200 epochs with a batch size of 2, and early stopping based on validation loss. The remaining training hyperparameters for each model were tuned using Bayesian Optimization and Random Search with the Keras Tuner to ensure a fair comparison.

\subsection{Evaluation Metrics}
To evaluate the performance of our model in predicting PWL values, we employ a comprehensive suite of metrics. Each metric is chosen to assess a different aspect of predictive accuracy, from point-wise water depth errors to the spatial correctness of the flood extent, ensuring a holistic evaluation relevant to practical flood risk management.

\begin{itemize}[left=0pt]
    \item \textbf{Average Relative Total Absolute Error (ARTAE):} In flood modeling, the significance of a prediction error is often relative to the local water depth. An error of 0.2m is critical in a shallow, 0.5m flood but less so in a deep, 4m flood. ARTAE addresses this by measuring error relative to the true value, providing a scale-invariant assessment of the model's accuracy. It quantifies the relative error between the predicted $y_{\text{p},i}$ and true values $y_{\text{t},i}$ using the normalized \( L_1 \) difference:
    \begin{align}
    \text{ARTAE} \triangleq \frac{1}{N} \sum_{i=1}^N \frac{\| y_{\text{t},i} - y_{\text{p},i} \|_1}{\| y_{\text{t},i} \|_1}
    \end{align}
    where \( N \) denotes the total data samples.

    \item \textbf{Average Root Mean Square Error (ARMSE):} For flood risk assessment, large prediction errors can have catastrophic consequences, such as failing to predict the inundation of a key evacuation route or a critical facility like a hospital. ARMSE is highly sensitive to these large deviations because it squares the errors before averaging. It is therefore used to penalize and highlight instances of significant prediction failures. It captures the root mean square error for each sample, as expressed:
    \begin{align}
    \text{ARMSE} \triangleq \frac{1}{N} \sum_{i=1}^N \sqrt{\frac{1}{d_y} \sum_{j=1}^{d_y} (y_{\text{t},i,j} - y_{\text{p},i,j})^2}
    \end{align}
    where \( d_y \) indicates the dimensionality of each sample.

    \item \textbf{Average Mean Absolute Error (AMAE):} In contrast to ARMSE, AMAE provides an intuitive measure of the average error magnitude across all spatial points, without being disproportionately skewed by a few extreme outliers. This offers a robust, general assessment of the model's expected performance on a per-pixel basis. The AMAE is calculated as:
    \begin{align}
    \text{AMAE} \triangleq \frac{1}{N} \sum_{i=1}^N \frac{1}{d_y} \sum_{j=1}^{d_y} |y_{\text{t},i,j} - y_{\text{p},i,j}|
    \end{align}

    \item \textbf{Coefficient of Determination (\( R^2 \)):} Beyond average error, it is important to know if the model correctly captures the spatial variability of a flood event. The \( R^2 \) metric assesses this by measuring the proportion of variance in the ground truth that is explained by the model. A high \(R^2\) value indicates the model is effective at predicting the location and severity of flood peaks and troughs. It is computed as:
    \begin{align}
    R^2 \triangleq \frac{1}{N} \sum_{i=1}^N \left( 1 - \frac{\sum_{j=1}^{d_y} (y_{\text{t},i,j} - y_{\text{p},i,j})^2}{\sum_{j=1}^{d_y} (y_{\text{t},i,j} - \bar{y}_{\text{t},i})^2} \right)
    \end{align}
    where \( \bar{y}_{\text{t}}^k \) is the mean of the true values for the \( k \)-th sample.

    \item \textbf{Threshold Exceedance Metric (\( \delta > \Delta \)):} This metric is directly tied to operational decision-making. In flood management, specific error thresholds (\(\Delta\)) often correspond to critical infrastructure limits, such as the floor height of a building or the elevation of a major roadway. This metric quantifies the frequency of 'critical failures' (cases where the prediction error exceeds this pre-defined safety margin). It is defined as:
    \begin{align}
    \delta > \Delta \triangleq \frac{1}{N} \sum_{i=1}^N \frac{\left| \{ j : |y_{\text{t},i,j} - y_{\text{p},i,j}| > \Delta \} \right|}{d_y}
    \end{align}

    \item \textbf{Non-inundated Prediction Accuracy (\(\text{Acc}[0]\)):} Given the high class imbalance in flood maps (most areas are dry), it is crucial to verify that the model is not prone to false alarms. This metric specifically measures the ability of the model to correctly identify non-inundated (safe) zones. High accuracy is essential for building trust in the model and ensuring the reliability of evacuation and land-use planning. It is computed as:
    \begin{align}
    \text{Acc}[0] \triangleq \frac{1}{N} \sum_{i=1}^N \frac{\left| \{ j : y_{\text{t},i,j} = 0 \} \right|}{d_y}
    \end{align}

    \item {\textbf{Dice Similarity Coefficient (DSC):}} To address spatial fitness, we introduce the DSC, a standard metric for evaluating the spatial overlap between predicted and true flood extents. Unlike the point-wise error metrics above, the DSC assesses the geometric accuracy of the inundation area. To compute the DSC, the continuous model outputs ($y_p$) and ground truth values ($y_t$) are first converted into binary inundation masks by applying a threshold (any pixel with a water depth > 0 is considered inundated). From these masks, we calculate the overlap:

    \begin{align}
    \text{DSC} \triangleq \frac{2 \times |\text{TP}|}{2 \times |\text{TP}| + |\text{FP}| + |\text{FN}|}
    \end{align}
    
    where true positives (\text{TP}) represents the area correctly predicted as flooded, false positives (\text{FP}) represents the overpredicted (wet where it should be dry) area, and false negatives (\text{FN}) represents the underpredicted (dry where it should be wet) area. This metric provides a direct measure of the model's ability to correctly delineate the flood boundaries.
\end{itemize}

\section{Results}
In this section, we evaluate the performance of CASPIAN-v2 model through quantitative and qualitative analyses. 

\subsection{Quantitative Results}
\subsubsection{Performance Metrics on Test Set}
We first report the performance of CASPIAN-v2 on the test set, as shown in Table \ref{Table_3}. For AD data, the model achieves an AMAE of 0.0586, ARMSE of 0.4079, and a high average R\textsuperscript{2} score of 0.9556, indicating excellent explanatory power. The ARTAE of 4.2793\% and low error percentages ($\delta>0.5\%$: 1.02\% and $\delta>0.1\%$: 4.37\%) highlight higher precision in accurately predicting flood inundation levels. Similarly for SF, the model achieves an AMAE of 0.0320, ARMSE of 0.2094, and an average R\textsuperscript{2} score of 0.9214. While the ARTAE is higher at 8.8129\%, the model maintains high accuracy metrics with an Acc[0] of 99.76\% compared to 99.04\% in AD.

On the combined dataset, CASPIAN-v2 performs consistently well with an AMAE of 0.0453, ARMSE of 0.3087, and an average R\textsuperscript{2} score of 0.9385. The combined ARTAE of 6.5461\% and low error percentages ($\delta>0.5$: 0.89\% and $\delta>0.1$: 3.55\%) demonstrate balanced performance across regions. The high Acc[0] of 99.39\% further underscores the reliability of the model in accurately predicting coastal inundation.

%%%%%%%%%%%%%%%%%%%%%%%%%
\begin{table}[htb]
\setlength{\tabcolsep}{7pt}
\renewcommand{\arraystretch}{1.25}
\centering
\footnotesize
\begin{threeparttable} 
\caption{\label{Table_3}Evaluation of CASPIAN-v2 on test set. $\downarrow$ indicates that lower values are better, and $\uparrow$ indicates that higher values are better.}
\begin{tabular}{c|ccccccc}
\hline
\textbf{Dataset}		&	\textbf{AMAE $\downarrow$}	&	\textbf{ARMSE $\downarrow$}	&	\textbf{ARTAE $\downarrow$}	&	\textbf{Avg. $\delta>0.5$ $\downarrow$} 	&	\textbf{Avg. $\delta>0.1$ $\downarrow$}	&	\textbf{Avg. R$^2$ Score $\uparrow$}	&	\textbf{Avg. Acc[0] $\uparrow$}		\\
\hline
AD	&	0.0586	&	0.4079	&	4.2793	&	1.02\%	&	4.37\%	&	0.9556	&	99.04\%	\\
SF	&	0.0320	&	0.2094	&	8.8129	&	0.75\%	&	2.72\%	&	0.9214	&	99.76\%	\\
Combined	&	0.0453	&	0.3087	&	6.5461	&	0.89\%	&	3.55\%	&	0.9385	&	99.39\%	\\
\hline
\end{tabular}
\end{threeparttable}
\end{table}
%%%%%%%%%%%%%%%%%%%%%%%%%%%%%%

\subsubsection{Performance Metrics on Holdout Set}
In this section, we present CASPIAN-v2 performance on the holdout set. The results are reported in Table \ref{Table_4}, where it can be observed that the model achieves an AMAE of 0.0792, an ARMSE of 0.4871, and an average R\textsuperscript{2} score of 0.9525 for AD. Furthermore, the small percentages of errors ($\delta>0.5$: 1.29\% and $\delta>0.1$: 5.48\%) underscore its accuracy in predicting flood inundation levels.

Similarly, for SF, CASPIAN-v2 achieves an AMAE of 0.0317, an ARMSE of 0.2259, and an average R\textsuperscript{2} score of 0.9694. Compared to AD, the ARTAE of 4.0009\% indicates slightly more predictions that have larger relative errors. However, with Acc[0] of 99.64\%, the model achieves better non-inundated prediction accuracy compared to 99.07\% in AD-Holdout.

Overall, CASPIAN-v2 achieves an AMAE of 0.0512, an ARMSE of 0.3331, and an average R\textsuperscript{2} score of 0.9625 on the aggregated holdout dataset. The ARTAE of 3.7167\% and small error percentages ($\delta>0.5$: 1.04\% and $\delta>0.1$: 4.17\%) signify consistent performance in both regions. The higher Acc[0] of 99.41\% further confirms its reliability in predicting flood inundation across diverse and challenging shoreline scenarios.

%%%%%%%%%%%%%%%%%%%%%%%%%%%
\begin{table}[htb]
\setlength{\tabcolsep}{7pt}
\renewcommand{\arraystretch}{1.25}
\centering
\footnotesize
\begin{threeparttable} 
\caption{\label{Table_4}Evaluation of CASPIAN-v2 on holdout set.}
\begin{tabular}{c|ccccccc}
\hline
\textbf{Dataset}		&	\textbf{AMAE $\downarrow$}	&	\textbf{ARMSE $\downarrow$}	&	\textbf{ARTAE $\downarrow$}	&	\textbf{Avg. $\delta>0.5$ $\downarrow$} 	&	\textbf{Avg. $\delta>0.1$ $\downarrow$}	&	\textbf{Avg. R$^2$ Score $\uparrow$}	&	\textbf{Avg. Acc[0] $\uparrow$}		\\
\hline
AD - Holdout	&	0.0792	&	0.4871	&	3.3081	&	1.29\%	&	5.48\%	&	0.9525	&	99.07\%	\\
SF - Holdout	&	0.0317	&	0.2259	&	4.0009	&	0.86\%	&	3.26\%	&	0.9694	&	99.64\%	\\
Combined	&	0.0512	&	0.3331	&	3.7167	&	1.04\%	&	4.17\%	&	0.9625	&	99.41\%	\\
\hline
\end{tabular}
\end{threeparttable}
\end{table}
%%%%%%%%%%%%%%%%%%%%%%%%%%%%

\subsubsection{Performance Benchmarking against SOTA Methods}
To comprehensively evaluate the performance of CASPIAN-v2, we benchmarked it against a suite of SOTA traditional ML and DL models. The selection and implementation details for these baseline models are described in Section~\ref{baseline_models}. This section presents a detailed comparison the prediction performance across all models, with the full results presented in Table~\ref{Table_5}. The analysis is broken down by model class, first comparing against traditional ML methods, and then against other DL architectures.

\textit{Comparison with Machine Learning Models:}

In this section, we compare the performance of CASPIAN-v2 against various traditional ML models for flood prediction, as shown in Table \ref{Table_5}. The Naïve model shows high errors with an AMAE of 1.5343, ARMSE of 3.5444, and an average R\textsuperscript{2} score of 0.5450. Among traditional approaches, linear regression reduces errors significantly, achieving an AMAE of 0.1272, ARMSE of 0.1946, and an average R\textsuperscript{2} score of 0.9464. The lasso with polynomial model further improves performance, giving an AMAE of 0.0937, ARMSE of 0.1202, and the highest average R\textsuperscript{2} score of 0.9618 among traditional ML models.

Compared to the best traditional model (lasso with polynomial), CASPIAN-v2 reduces the AMAE by 51.65\% (from 0.0937 to 0.0453). However, CASPIAN-v2 has a higher ARMSE of 0.3087 compared to 0.1202, indicating it minimizes mean errors effectively but may experience larger individual prediction errors. Despite this, CASPIAN-v2 outperforms traditional models across multiple metrics, leveraging DL and multi-dimensional data integration to achieve superior accuracy in flood prediction.

This trend is even more pronounced in the spatial accuracy results. While the lasso model achieved a DSC of 0.6438, CASPIAN-v2 scored 0.8437, representing a 31.05\% improvement. This significant gap underscores the inherent limitations of traditional ML models in capturing the complex geometric shape of flood events, a task for which our deep learning architecture is better suited.

%%%%%%%%%%%%%%%%%%%%%%%%%%%%%%%%%%%%
\begin{table*}[t]
\setlength{\tabcolsep}{2.5pt}
\renewcommand{\arraystretch}{1.25}
\centering
\footnotesize
\begin{threeparttable} 
\caption{\label{Table_5} A comprehensive performance comparison between our proposed CASPIAN-v2 and state-of-the-art models, grouped into a baseline physics-based simulator (Delft3D), traditional ML, and DL approaches. Prediction accuracy is evaluated using eight standard metrics, where arrows indicate the desired direction ($\uparrow$ for higher is better, $\downarrow$ for lower is better). Computational efficiency is assessed by three key indicators: the total number of trainable parameters (M = millions), the total training time (TT), and the average inference time (IT) per sample. In the physics-based simulations, PP denotes Post-Processing. The simulation results, which provide the ground truth data, are included for reference. The top-performing result for each metric is highlighted in \textcolor{red}{red}, and the second-best is highlighted in \textcolor{blue}{blue}.
}
\midsepremove
\begin{tabular}{c|c|cccccccc|ccc}
\toprule % Use \toprule for the top line
\multirow{2}{*}{\textbf{Type}} & \multirow{2}{*}{\textbf{Model}}	& \multicolumn{8}{c|}{\textbf{Prediction Accuracy}} & \multicolumn{3}{c}{\textbf{Computational Efficiency}} \\
\cmidrule(lr){3-10} \cmidrule(lr){11-13} % Use \cmidrule instead of \cline. (lr) trims the edges.
&	&	\textbf{AMAE $\downarrow$}	&	\textbf{ARMSE $\downarrow$}	&	\textbf{ARTAE $\downarrow$}	&	\textbf{$\delta>0.5$ $\downarrow$} 	&	\textbf{$\delta>0.1$ $\downarrow$}	&	\textbf{R$^2$ $\uparrow$}	&	\textbf{Acc[0] $\uparrow$} & \textbf{DSC $\uparrow$}	& \textbf{Param$\downarrow$} & \textbf{TT $\downarrow$} & \textbf{IT $\downarrow$}\\
\midrule % Use \midrule to separate header from data
\multirow{2}{*}{\textbf{Simulator}} & \textbf{AD:} Delft3D+SWAN+PP & \multicolumn{8}{c|}{\multirow{2}{*}{\textit{Served as the ground truth}}} & - & - & 71--73h \\
& \textbf{SF:} Delft3D+PP & \multicolumn{8}{c|}{} & - & - & 3.5--6.0h \\
\midrule % Use midrule to separate sections
\multirow{7}{*}{\STAB{\rotatebox[origin=c]{0}{\parbox[c]{20pt}{\centering\textbf{ML (1-D)}}}}} & Naïve	&	1.5343	&	3.5444	&	1746.0693	&	74.92\%	&	80.11\%	&	0.5450	&	31.01\%	& 0.3871 & - & \BLUEE{62s} & \BLUEE{0.15s}\\
& RF	&	0.5411	&	0.7310	&	264.9505	&	36.77\%	&	72.20\%	&	0.7962	&	34.19\%	& 0.4185 & - & 75s & 0.18s\\
& Linear	&	0.1272	&	\BLUEE{0.1946}	&	64.9859	&	7.87\%	&	14.03\%	&	0.9464	&	59.28\%	& 0.6279 & - & \REDD{65s} & \REDD{0.16s}\\
& XGBoost	&	0.2546	&	0.2446	&	164.1654	&	16.27\%	&	49.88\%	&	0.9347	&	44.10\%	& 0.4711 & - & 198s & 0.21s\\
& SVR	&	0.2069	&	0.2423	&	72.3122	&	9.24\%	&	41.17\%	&	0.9298	&	45.46	& 0.4889 & - & 79s & 0.19s\\
& Lasso with Poly	&	0.0937	&	\REDD{0.1202}	&	28.1565	&	4.47\%	&	15.04\%	&	\REDD{0.9618}	&	55.78\%	& 0.6438 & - & 72s & 0.17s\\
& Kriging	&	0.1098	&	0.2478	&	39.9073	&	5.22\%	&	11.59\%	&	0.9414	&	62.88\%	& 0.6359 & - & 76s & 0.18s\\
\hline
{\multirow{2}{*}{\STAB{\rotatebox[origin=c]{0}{\parbox{20pt}{\centering\textbf{DL (1-D)}}}}}} & MLP	&	0.6486	&	2.7247	&	524.1724	&	32.82\%	&	41.94\%	&	0.6572	&	36.91\%	& 0.4356 & \BLUEE{0.01M} & 14h & 5.03s\\
& CCT	&	0.9064	&	2.3292	&	843.5430	&	48.08\%	&	64.63\%	&	0.6649	&	34.01\%	 & 0.4228 & 11.05M & 18h & 0.26s
\\
\hline
\multirow{6}{*}{\STAB{\rotatebox[origin=c]{0}{\parbox{20pt}{\centering\textbf{DL (2-D)}}}}} & Atten-Unet	&	0.1061	&	0.3714	&	11.8245	&	3.14\%	&	16.70\%	&	0.9195	&	95.26\%	& 0.7390  & 12.07M & 46h & 0.24s\\
& Atten-Unet*	&	0.1032	&	0.3627	&	11.6585	&	3.31\%	&	15.62\%	&	0.9210	&	94.99\% & 0.7469 & 12.07M & 47h & 0.27s\\
& Swin-Unet	&	0.0629	&	0.2788	&	6.7244	&	1.47\%	&	12.94\%	&	\BLUEE{0.9514}	&	98.10\%	& 0.8014 & 8.29M & 26h & 0.24s\\
& CASPIAN	&	\BLUEE{0.0566}	&	0.3613	&	\REDD{5.8573}	&	\BLUEE{1.01\%}	&	\BLUEE{4.79\%}	&	0.9209	&	\BLUEE{98.84\%} & \BLUEE{0.8261}	& \REDD{0.36M} & 22h & 0.22s\\
%\specialrule{1.25pt}{0pt}{0pt} % Add extra spacing for the last row
\cmidrule[1.15pt]{2-13} % This line is fine, it separates the SOTA from "Ours"
% --- CORRECTED ROW BELOW ---
& \textbf{CASPIAN-v2 (Ours)} 
& \REDD{0.0453} & 0.3087 & \BLUEE{6.5461} & \REDD{0.89\%} & \REDD{3.55\%} & 0.9385 & \REDD{99.39\%} & \REDD{0.8437} & 0.38M & 22h & 0.22s\\
\hline

\end{tabular}
\midsepdefault
\begin{tablenotes}
    \item[$*$] with pre-trained encoder on ImageNet \cite{deng2009imagenet}.

\end{tablenotes}
\end{threeparttable}
\end{table*}
%%%%%%%%%%%%%%%%%%%%%%%%%%%

\textit{Comparison with Deep Learning Models:}

Existing 1D and 2D DL models show varied performance, as reported in Table \ref{Table_5}. The CCT model achieves an AMAE of 0.9064, an ARMSE of 2.3292, and an average R\textsuperscript{2} score of 0.6649, indicating moderate predictive capabilities. Atten-Unet and its variant Atten-Unet* improve performance with AMAE values of 0.1061 and 0.1032 and average R\textsuperscript{2} scores of 0.9195 and 0.9210, respectively. Swin-Unet achieves further improvements, reducing the AMAE to 0.0629 and attaining an average R\textsuperscript{2} score of 0.9514, reflecting its effectiveness in capturing spatial dependencies.

Compared to the second-best DL model, CASPIAN-v2 reduces the AMAE by 19.96\% (from 0.0566 to 0.0453) and achieves an exceptional average Acc[0] of 99.39\%, surpassing CASPIAN’s 98.84\%. These results highlight superior accuracy and robust generalization capabilities of CASPIAN-v2. 

In terms of spatial fitness, CASPIAN-v2 (with DSC of 0.8437) also demonstrates a clear advantage over the best-performing DL baseline, CASPIAN (0.8261), representing a 2.13\% improvement in spatial accuracy. Taken together, these results highlight the superior accuracy and robust generalization capabilities of CASPIAN-v2. The integration of advanced components such as the MARX and SEE blocks, combined with an optimized Hybrid loss function, enables the effective modeling of complex flood dynamics.

\subsubsection{Computational Efficiency Analysis}

A primary motivation for this research is to overcome the significant computational burden of physics-based hydrodynamic simulators. The final three columns of Table~\ref{Table_5} provide a comprehensive comparison of the computational efficiency of all evaluated models.

As expected, the traditional ML models are the fastest to train, typically requiring only a few minutes. However, this speed comes at the cost of significantly lower prediction accuracy. Among the more accurate DL models, CASPIAN-v2 demonstrates a highly favorable balance of performance and efficiency. With only 0.38 million parameters, it is one of the most lightweight 2D models, comparable in size to the original CASPIAN (0.36M) and substantially smaller than transformer-based models like Swin-Unet (8.29M) or other U-Net variants (12.07M). Its training time (22 hours) and inference time (0.22s per scenario) are also highly competitive within this high-performing group.

The most critical comparison, however, is against the physics-based simulator. Generating a single flood scenario is an exceptionally demanding task. For Abu Dhabi, a full simulation requires 71 to 73 hours of elapsed runtime on high-performance computing infrastructure due to the coupling of Delft3D and SWAN models and extensive post-processing. For San Francisco Bay, where the simulation was less complex, the process still required a substantial 3.5 to 6.0 hours (as detailed in Section~\ref{sec:DataAndSims}). Extrapolating these figures, simulating our full test set of 72 scenarios (36 for each region) would demand approximately 2,763 hours (nearly 115 days) of continuous computation. In stark contrast, CASPIAN-v2 can predict the outcomes for all 72 scenarios in just under 16 seconds. This represents a monumental reduction in computational time, transforming a months-long endeavor into a near-instantaneous task and positioning CASPIAN-v2 as a practical and scalable tool for real-world coastal planning.

\subsubsection{Numerical Assessment of Generalizability}

This section reports the generalization performance of CASPIAN-v2 on unseen data. The model was fine-tuned using new SF data corresponding to 0.5~m and 1.5~m SLR depths, encompassing 30 protection scenarios where one OLU was protected at a time (more details in Supplementary Material Section S5). For evaluation, 20\% of the data (6 samples) was reserved, while the remaining 80\% (24 samples) was used for fine-tuning and validation. Fine-tuning spanned 100 epochs with a progressive gradual recall approach, mixing the new data with the AD and SF holdout data in a 20:80 test/train ratio. The training set began with 70\% of the AD and SF holdout set combined with 30\% of the new data, gradually increasing to 70\% by the end of training.

The results in Table \ref{Table_6} demonstrate strong generalization by CASPIAN-v2 across SLR scenarios. For SF 0.5~m data, the model achieved an AMAE of 0.0626, ARMSE of 0.2996, and average R\textsuperscript{2} score of 0.9336. An ARTAE of 6.4240\% and low error percentages ($\delta>0.5$: 1.89\% and $\delta>0.1$: 7.79\%) highlight its precision. For SF 1.5~m data, the model showed slightly suboptimal performance with an AMAE of 0.1005, ARMSE of 0.4565, and average R\textsuperscript{2} score of 0.9196. The ARTAE of 4.3961\% indicates balanced performance, with an average Acc[0] of 98.23\% compared to 97.99\% for 0.5~m data.

When retaining existing knowledge, CASPIAN-v2 achieved an AMAE of 0.0567 and ARMSE of 0.2274 on the AD holdout set for 0.5~m SLR, with an average R\textsuperscript{2} score of 0.9901. The ARTAE of 2.5225\% and low error percentages ($\delta>0.5$: 0.53\% and $\delta>0.1$: 17.87\%) emphasize its precision. For the SF holdout set at 1.0~m SLR, the model achieved an AMAE of 0.0433, ARMSE of 0.2318, and average R\textsuperscript{2} score of 0.9685. The ARTAE of 4.6277\% and error percentages ($\delta>0.5$: 0.79\% and $\delta>0.1$: 9.61\%) reflect its ability to balance low absolute and relative errors, with an Acc[0] of 99.34\%.

Overall, the model achieved an AMAE of 0.0652, an ARMSE of 0.3040, and an average R\textsuperscript{2} score of 0.9520, revealing robust generalization abilities of the model across various SLR settings. Further, the model achieved an ARTAE of 4.5871\% and low error percentages ($\delta>0.5\%$: 1.31\% and $\delta>0.1\%$: 12.07\%), with a high Acc[0] of 98.69\%. These findings highlight the ability of the CASPIAN-v2 model to effectively generalize to new and previously unseen scenarios with minor fine-tuning, making it a reliable tool for real-world inundation prediction.

%%%%%%%%%%%%%%%%%%%%%%%%
\begin{figure}[htbp]  
\centering
\includegraphics[width=0.7\linewidth]{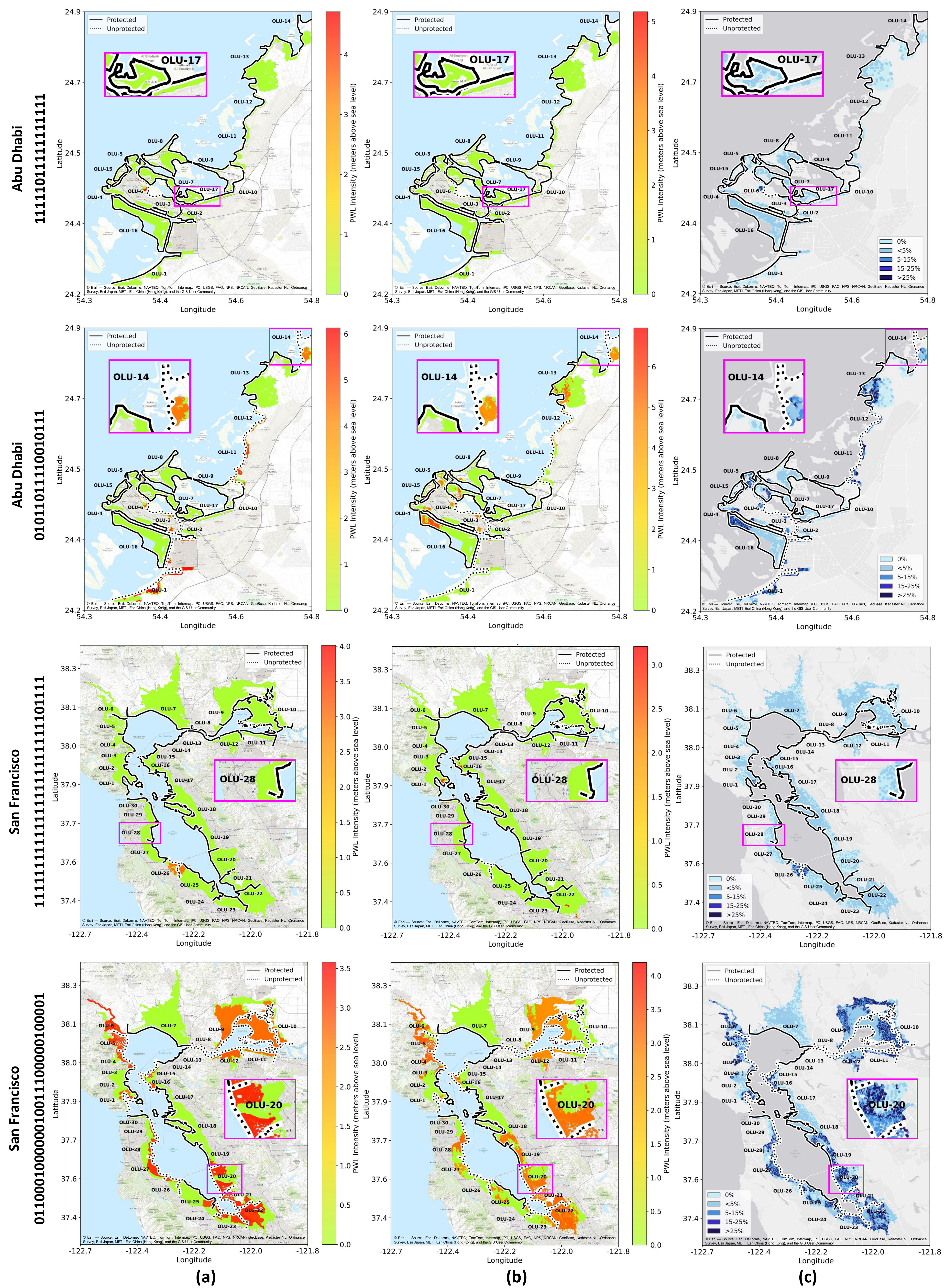}
\caption{Evaluation of CASPIAN-v2 on the test datasets. (a) Ground truth inundation maps for representative AD and SF scenarios. (b) Predicted inundation values. (c) Absolute error distributions of predicted inundation values. Darker shades of blue indicate higher absolute errors, ranging from near 0\% to greater than 25\%. The magenta insets provide zoomed-in views of specific OLUs to illustrate the effect of protection measures. For instance, the inundation is shown to be minimal inland of the protected OLU-17 in AD, whereas significant flooding occurs near the unprotected OLU-20, a dynamic that the model precisely captures.}
\label{fig:TestSet}
\end{figure}
%%%%%%%%%%%%%%%%%%%%%%%%%%%%%%%%

%%%%%%%%%%%%%%%%%%%%%%%%
\begin{table}[htb]
\setlength{\tabcolsep}{5pt}
\renewcommand{\arraystretch}{1.25}
\centering
\footnotesize
\begin{threeparttable} 
\caption{\label{Table_6}CASPIAN-v2 generalizability evaluation using different SLR data.}
\begin{tabular}{c|ccccccc}
\hline
\textbf{Dataset \textit{(SLR)}}		&	\textbf{AMAE $\downarrow$}	&	\textbf{ARMSE $\downarrow$}	&	\textbf{ARTAE $\downarrow$}	&	\textbf{Avg. $\delta>0.5$ $\downarrow$} 	&	\textbf{Avg. $\delta>0.1$ $\downarrow$}	&	\textbf{Avg. R$^2$ Score $\uparrow$}	&	\textbf{Avg. Acc[0] $\uparrow$}	\\
\hline
SF - Generalizability \textit{(0.5~m)}	&	0.0626	&	0.2996	&	6.4240	&	1.89\%	&	7.79\%	&	0.9336	&	97.99\%	\\
SF- Generalizability \textit{(1.5~m)}	&	0.1005	&	0.4565	&	4.3961	&	1.97\%	&	14.51\%	&	0.9196	&	98.23\%	\\
AD - Holdout \textit{(0.5~m)}	&	0.0567	&	0.2274	&	2.5225	&	0.53\%	&	17.87\%	&	0.9901	&	99.18\%	\\
SF - Holdout \textit{(1.0~m)}	&	0.0433	&	0.2318	&	4.6277	&	0.79\%	&	9.61\%	&	0.9685	&	99.34\%	\\
Overall	&	0.0652	&	0.3040	&	4.5871	&	1.31\%	&	12.07\%	&	0.9520	&	98.69\%	\\
\hline
\end{tabular}
\end{threeparttable}
\end{table}
%%%%%%%%%%%%%%%%%%%%%%%%%%%%

\subsection{Qualitative Results}

\subsubsection{Visual Performance on Test Set}
In this section, we provide a qualitative assessment of the performance of CASPIAN-v2 on the test set. Figure \ref{fig:TestSet} presents two randomly selected scenarios for the AD and SF regions, where it can be observed that the predicted inundation values of the proposed model closely align with the corresponding ground truth values. In single unprotected OLU scenarios (rows 1 and 3), the model accurately captures localized flooding effects, showing sensitivity to minor protection configuration changes. Similarly, CASPIAN-v2 effectively handles the increased complexity of mixed OLU protection statuses (rows 2 and 4). These results highlight the robustness of the model in generalizing across diverse regions and protection patterns. Figure \ref{fig:TestSet}(c) shows the absolute error maps, where it can be observed that the CASPIAN-v2 model produced minimal errors, with deviations occurring mainly in areas with sharp transitions in flood depths. However, these small variations minimally affect the overall prediction accuracy.

To illustrate the local impact of the protection measures on flood dynamics, zoomed-in insets are provided for specific OLUs. For instance, the first inset for AD highlights how inundation patterns are directly controlled by the protection status of the nearest OLU. When OLU-17 is protected, the area behind it remains largely dry, whereas significant flooding occurs inland of the unprotected OLU-14.

%%%%%%%%%%%%%%%%%%%%%%%%
\begin{figure}[htbp]  
\centering
\includegraphics[width=0.7\linewidth]{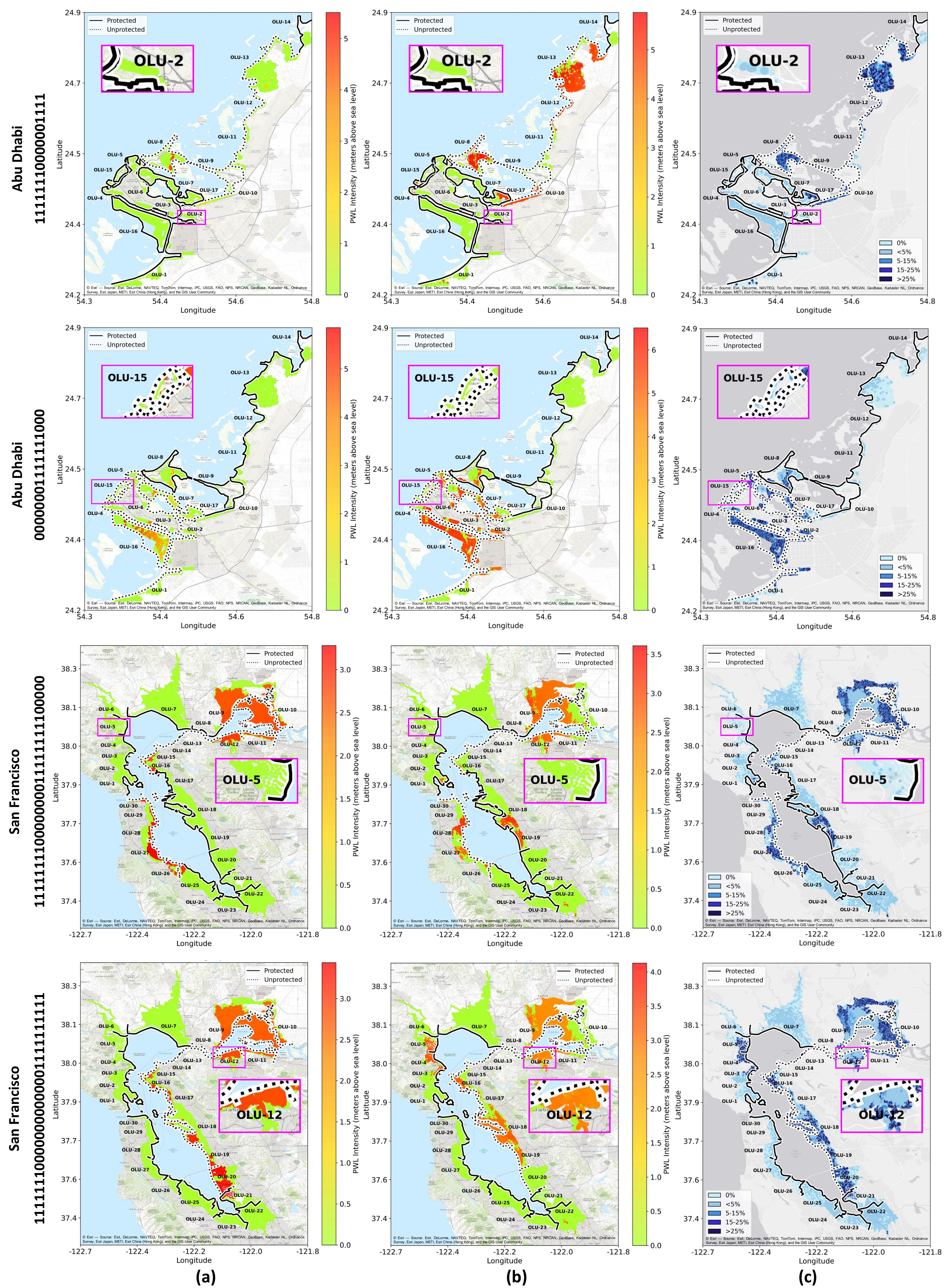}
\caption{Evaluation of CASPIAN-v2 on the holdout datasets. (a) Ground truth inundation maps for representative AD and SF scenarios. (b) Predicted inundation values. (c) Absolute error distributions of predicted inundation values. Darker shades of blue indicate higher absolute errors, ranging from near 0\% to greater than 25\%. The zoomed-in insets highlight fine-grained hydrodynamic effects. For instance, the successful prevention of inundation by a protected OLU-2 in AD, versus the widespread inland flooding resulting from an unprotected OLU-12 in SF.}
\label{fig:Holdout}
\end{figure}
%%%%%%%%%%%%%%%%%%%%%%%%%%%%%%%%

%%%%%%%%%%%%%%%%%%%%%%%%
\begin{figure}[htb]  
\centering
\includegraphics[width=1\linewidth]{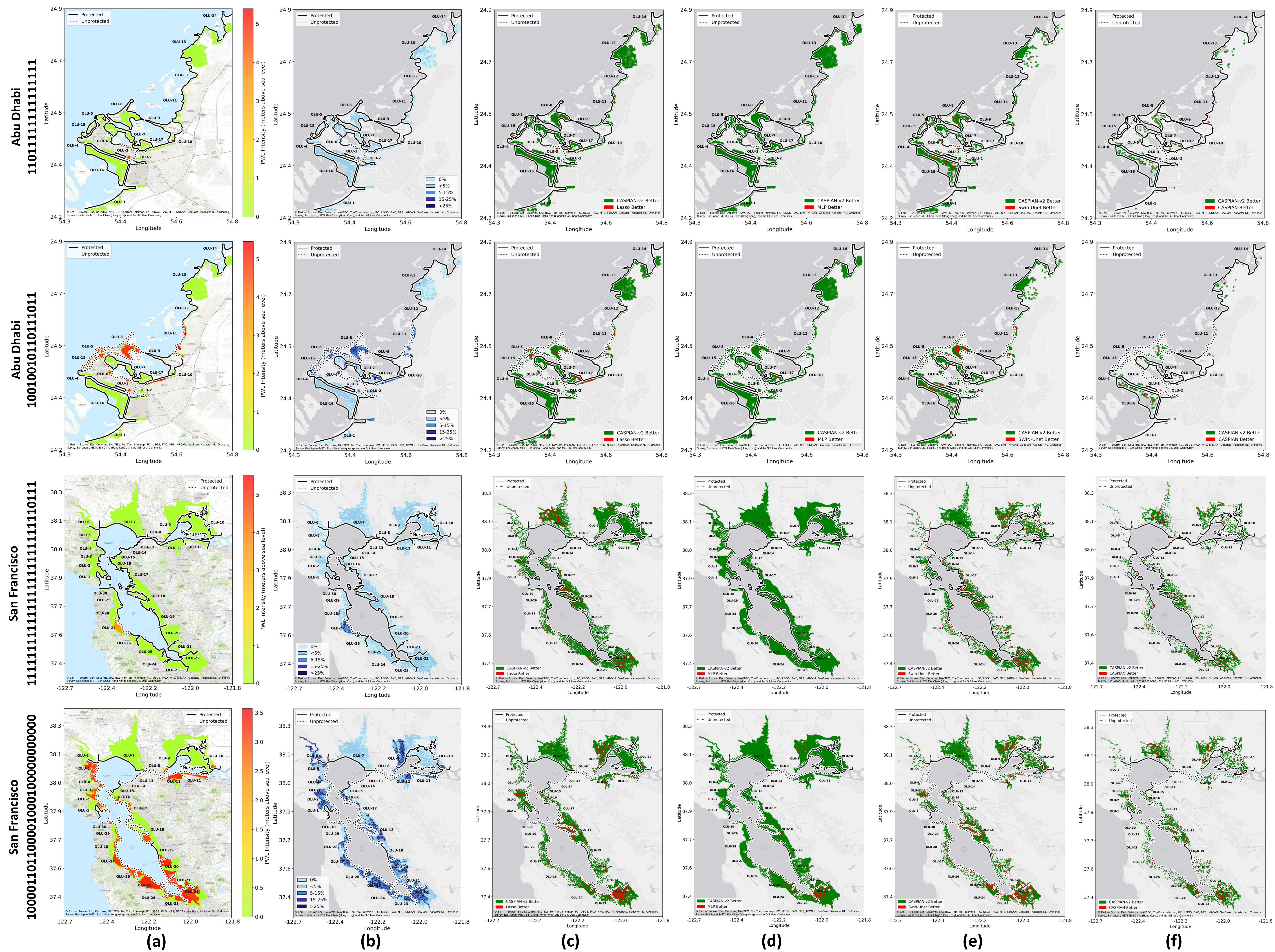}
\caption{Qualitative comparison of CASPIAN-v2 with SOTA approaches in predicting coastal flood inundation (a) Ground truth inundation maps for representative AD and SF scenarios. (b) Absolute error map for our proposed CASPIAN-v2 model, with darker blue indicating higher error. (c-f) Error difference maps comparing CASPIAN-v2 to key baselines. In these maps, green indicate regions where CASPIAN-v2 is more accurate than the baseline, red areas show where the baseline performed better, and transparent regions denote similar performance. The visualization clearly shows that CASPIAN-v2 provides a substantial improvement over the (c) Lasso, (d) MLP, (e) Swin-Unet, and (f) original CASPIAN models.}
\label{fig:SOTA}
\end{figure}
%%%%%%%%%%%%%%%%%%%%%%%%%%%%%%%%

\subsubsection{Visual Performance on Holdout Set}
In this section, we demonstrate the performance of CASPIAN-v2 using a holdout set composed of particularly challenging coastal protection scenarios. Figure ~\ref{fig:Holdout} showcases the performance of the model on two challenging configurations from the holdout set, which was specifically designed to test generalization across complex protection scenarios. These scenarios feature intricate mixes of protected and unprotected OLUs, creating sharp inundation boundaries where flooded and non-flooded regions meet. CASPIAN-v2 demonstrates high fidelity in these cases, accurately capturing these abrupt changes in local flood behavior. For instance, it correctly captures the inundation dynamics when one side of the SF bay is protected and the other is not (last row of Figure ~\ref{fig:Holdout}). 

The strong performance of the model here is particularly noteworthy given that it was trained on only a small subset of the thousands of possible protection combinations ($2^n$, where $n$ is the number of OLUs). This success on unseen, complex configurations indicates that CASPIAN-v2 is not merely memorizing training data but is learning the underlying spatial logic of how flood defenses influence inundation patterns. This affirms its robustness and reliability for real-world application.

\subsubsection{Visual Comparison with SOTA Methods}
We qualitatively evaluated the performance of the proposed CASPIAN-v2 by visually comparing its prediction errors with those of key SOTA baselines. Figure~\ref{fig:SOTA} presents this analysis for representative scenarios in both Abu Dhabi and San Francisco. Figure~\ref{fig:SOTA}(b) shows the absolute error map for our proposed CASPIAN-v2 model, demonstrating that errors are generally low and confined to complex hydraulic transition zones. The key insights, however, come from the error difference maps (Figure~\ref{fig:SOTA} (c-f)), which directly compare the spatial accuracy of CASPIAN-v2 to each baseline. In these maps, green areas highlight regions where CASPIAN-v2 is more accurate, while red indicates where the baseline had a lower error, and transparent areas denote regions where both models performed similarly.

Compared to the Lasso with polynomial features Figure~\ref{fig:SOTA}(c) and MLP Figure~\ref{fig:SOTA}(d) baselines, CASPIAN-v2 offers a dramatic improvement, with vast green areas indicating its superior ability to capture the fundamental flood patterns that these simpler models miss. The comparison with the more advanced Swin-Unet Figure~\ref{fig:SOTA}(e) and the original CASPIAN Figure~\ref{fig:SOTA}(f) models is also convincing. While these models are more competitive, the difference maps still show a clear and consistent advantage for CASPIAN-v2, which successfully reduces errors in many of the most deeply inundated and complex areas.

\begin{figure}[htbp]  
\centering
\includegraphics[width=1\linewidth]{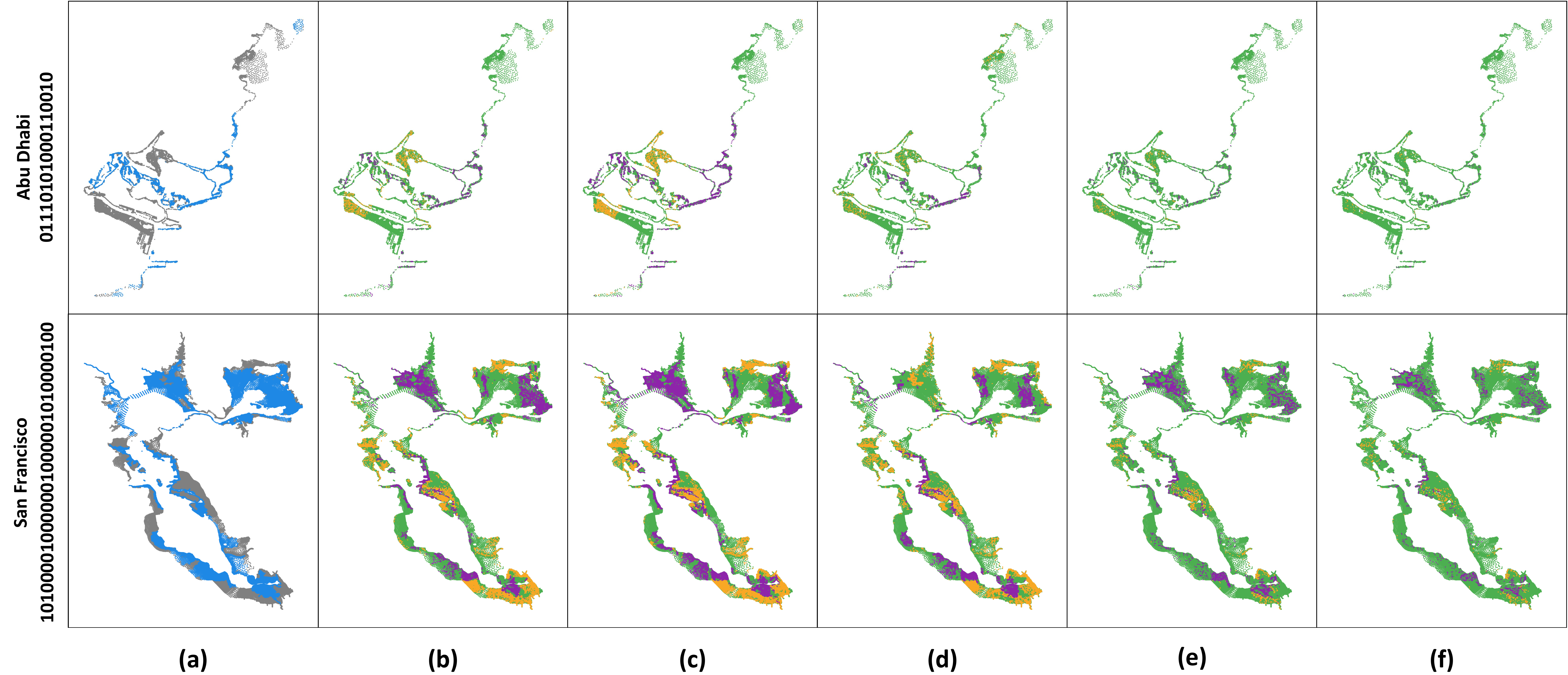}
\caption{Visual comparison of spatial prediction accuracy for CASPIAN-v2 versus the top-performing baseline model on a representative test case. Green indicates correctly predicted inundated areas (true positives), orange indicates over-prediction (false positives), and purple indicates under-prediction (false negatives). CASPIAN-v2 demonstrates a larger matched area and more coherent flood boundaries.}
\label{fig:Spatial_Fitness_Compare}
\end{figure}

Moreover, Figure~\ref{fig:Spatial_Fitness_Compare} visualizes the flood extents predicted by CASPIAN-v2 against the best-performing ML and DL baseline model. The map breaks down the predictions into correctly matched areas (green), over-predicted areas (orange), and under-predicted areas (purple). The visualization reveals that while the baseline models produce a more fragmented prediction with significant patches of both over- and under-prediction, the output of the proposed CASPIAN-v2 model aligns much more closely with the ground truth. Its predicted flood extent is more coherent and captures the true inundation boundaries with far fewer spatial errors. These qualitative comparisons align with the quantitative results in Table \ref{Table_5}, highlighting the ability of the proposed model to achieve higher accuracy and visually superior predictions.

%%%%%%%%%%%%%%%%%%%%%%%%
%%%%%%%%%%%%%%%%%%%%%%%%%%%%%%%%

\subsubsection{Visual Assessment of Generalizability}

\begin{figure}[htb]  
\centering
\includegraphics[width=0.95\linewidth]{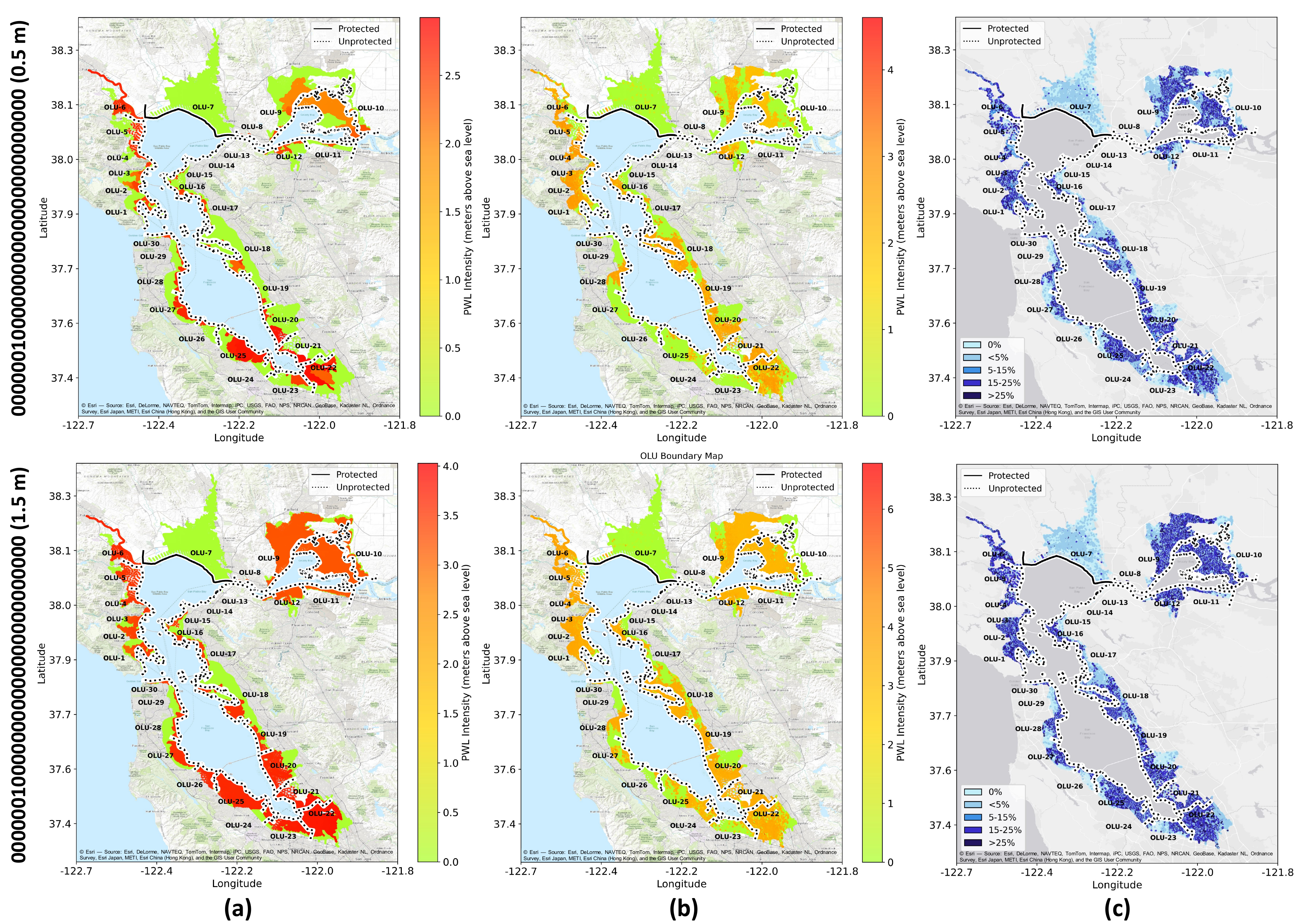}
\caption{Generalizability evaluation of CASPIAN-v2 fine-tuned for 0.5~m and 1.5~m SLR scenarios. (a) Ground truth inundation maps. (b) Predicted inundation values. (c) Absolute error distributions of predicted inundation values. Darker shades of blue indicate higher absolute errors, ranging from near 0\% to greater than 25\%.}
\label{fig:FT}
\end{figure}

We next evaluate the generalizability of CASPIAN-v2 under different environmental conditions by fine-tuning the model on two additional SLR data of 0.5~m and 1.5~m. Figure~\ref{fig:FT} shows the prediction results, illustrating that while the fine-tuned model exhibits some localized discrepancies (Figure ~\ref{fig:FT}(c)), these deviations remain modest given the minimal training data and limited fine-tuning epochs. In the 0.5~m SLR scenario, the model yields relatively lower absolute errors in predicting flood extents. By contrast, the 1.5~m scenario exhibits slightly higher errors, likely due to the increased variability in PWL values. Nonetheless, the predictions generally align well with the ground truth inundation patterns.

Overall, these findings underscore adaptability of the proposed model to evolving coastal conditions, suggesting that with sufficient training data and appropriately tuned hyperparameters, the model can maintain robust performance across a broad range of SLR scenarios.

%%%%%%%%%%%%%%%%%%%%%%%%
%%%%%%%%%%%%%%%%%%%%%%%%%%%%%%%%

\section{Discussion and Conclusion}

This research presents a novel DL model to predict coastal inundation across two geographical locations (AD and SF). The effectiveness of the proposed CASPIAN-v2 model is validated through extensive experiments, where it outperforms the existing SOTA methods, as shown in Table \ref{Table_5}. Although traditional ML approaches are relatively fast to train, these methods lack the ability to capture complex spatial patterns in the data, thus producing less accurate results. Similarly, we found that 1D DL approaches do not scale effectively to large, spatially focused grids. Furthermore, jointly training these methods on AD and SF datasets was less successful and yielded poor results due to inconsistent input features, particularly the different number of OLUs across regions and the need to address a broader array of shoreline adaptation scenarios. In comparison, the proposed 2D DL model can learn complex input patterns, enabling it to produce superior prediction results. Additionally, our data augmentation strategy, which involved creating new training samples by applying random spatial cutouts and scaling factors (as mentioned in Section \ref{datasetsplit}), exposes the model to a wider variety of conditions. This enhances its resilience to noise, missing data, and varying shoreline configurations.. Moreover, CASPIAN-v2 demonstrates strong generalizability across different levels of SLR, which underlines its utility for future resilience planning.

\begin{figure}[htb]  
\centering
\includegraphics[width=0.95\linewidth]{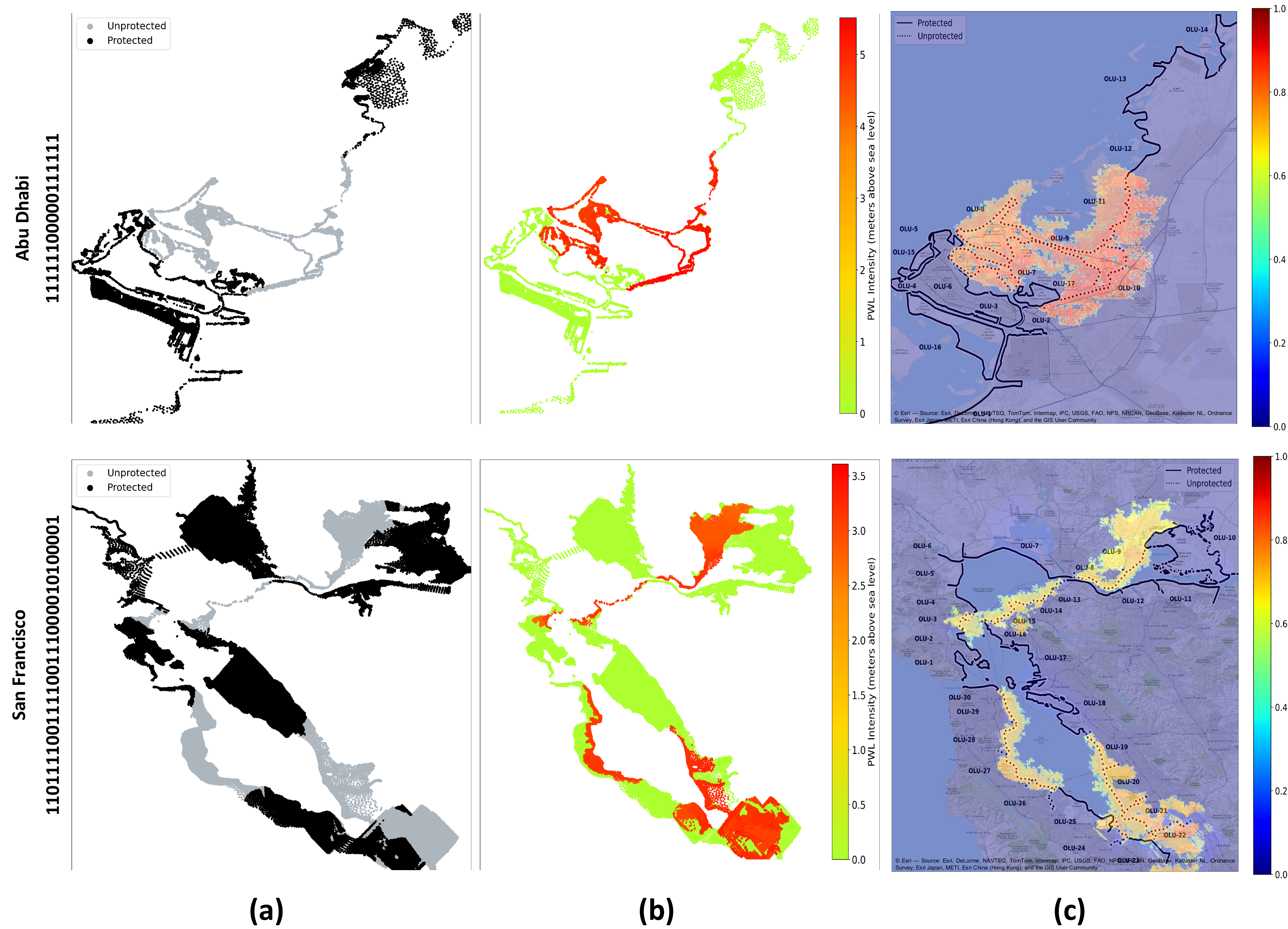}
\caption{CASPIAN-v2 inundation prediction for AD (top) and SF (bottom): (a) Input images representing protected and unprotected OLUs, (b) Predicted inundation with PWL intensity, (c) Grad-CAM visualizations highlighting model attention, where warmer colors indicate regions the model focused on most during prediction, aligning with unprotected and vulnerable areas.}
\label{fig:XAI}
\end{figure}

A critical aspect influencing model performance is the underlying data distribution. As is common in flood modeling, our dataset is highly imbalanced, with a vast majority of non-inundated (zero value) points compared to the relatively rare inundated points (see Supplementary Material, Figure S6). To address this significant challenge, our framework employs a multi-faceted strategy. First, our Hybrid Loss function is inherently designed to handle this skew. The Quantile loss component allows us to place more weight on correctly predicting the less frequent, but more critical, positive flood values, while the Huber loss prevents the numerous small errors in non-inundated areas from dominating the training process. Second, the attention mechanism within the MARX block is crucial. This theoretical benefit is substantiated by empirical evidence from our Grad-CAM analysis (Figure~\ref{fig:XAI}), which shows that the model focuses highly around the vulnerable, unprotected shoreline segments where inundation originates. This focus on salient regions prevents the model's learning from being diluted by the vast areas of non-inundated points. Finally, our choice of evaluation metrics, particularly the DSC and non-inundated accuracy, provides a more balanced assessment of performance. This combination of a tailored loss function, an attentive architecture with demonstrated focus, and robust evaluation allows CASPIAN-v2 to maintain high predictive fidelity despite the challenging data distribution.

%%%%%%%%%%%%%%%%%%%%%%%%
%%%%%%%%%%%%%%%%%%%%%%%%%%%%%%%%

Beyond its technical role in the model, this demonstrated interpretability provides critical insights for stakeholders. The clear spatial alignment between the focus of the model and known vulnerabilities (Figure~\ref{fig:XAI}c) serves to empirically validate its decision-making process. This level of transparency is instrumental for planners and policymakers, as it clarifies why specific areas are identified as high-risk, thereby fostering trust in DL-based solutions and aiding in the design of targeted resilience strategies.

\begin{figure}[htb]  
\centering
\includegraphics[width=\linewidth]{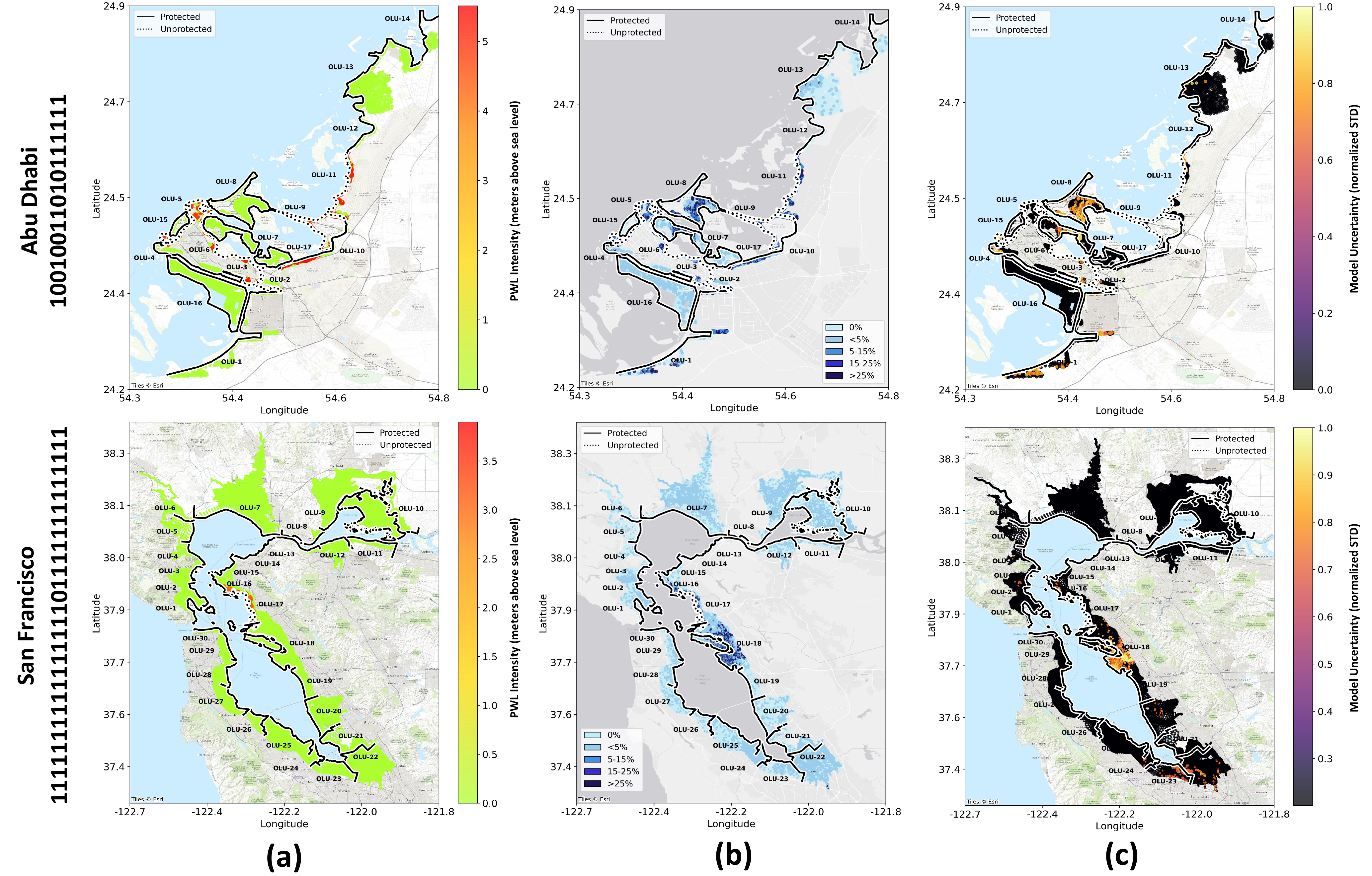}
\caption{Predictive uncertainty maps derived from the deep ensemble for representative AD and SF scenarios. (a) Ground truth inundation. (b) Absolute error of the ensemble mean prediction. (c) Pixel-wise predictive uncertainty, calculated as the normalized standard deviation of the five models output. lighter colors indicate higher uncertainty.}
\label{fig:uncertainty_maps}
\end{figure}

An essential element for any model designed for risk assessment, aside from interpretability, is the measurement of its uncertainty. To address this, we implemented a deep ensemble method to estimate the predictive uncertainty of CASPIAN-v2~\cite{lakshminarayanan2017simple}. We trained five independent models and used the pixel-wise standard deviation of their predictions as a direct proxy for model uncertainty (see Supplementary Material Section S8 for full quantitative results). The resulting maps, shown in Figure~\ref{fig:uncertainty_maps}, reveal a crucial characteristic of our model, which is a strong spatial correlation between predictive uncertainty and prediction error. The bright, high-uncertainty regions in panel Figure~\ref{fig:uncertainty_maps}(c) closely align with the areas of higher absolute error shown in panel Figure~\ref{fig:uncertainty_maps}(b), while the dark, low-uncertainty regions correspond to areas of high accuracy. This indicates that the model demonstrates a valuable form of self-awareness and it effectively learns to identify regions where its own predictions are less reliable. This is invaluable for coastal planners, as it allows them to trust the high-certainty predictions for general assessments while flagging the high-uncertainty zones as areas that require a higher margin of safety or further, more detailed hydrodynamic study. This ability to not only make accurate predictions but also to reliably signal its own confidence is instrumental for fostering trust and supporting real-world, risk-informed decision-making.

%%%%%%%%%%%%%%%%%%%%%%%%
% --- UNCERTAINTY FIGURE --- %
%%%%%%%%%%%%%%%%%%%%%%%%%%%%%%%%

To bridge the gap between our model and real-world coastal planning, we envision a conceptual workflow where CASPIAN-v2 serves as a rapid scenario-assessment tool. Planners and engineers could use the model to explore a vast design space of hundreds or thousands of potential shoreline protection configurations, a task that would be computationally infeasible with traditional hydrodynamic simulators alone. At key decision points, such as preliminary zoning or budget allocation, stakeholders could use the model to quickly identify a shortlist of the most effective protection strategies. These promising candidates could then be subjected to more detailed, rigorous analysis using the high-fidelity physics-based models. This two-tiered approach leverages the speed of our surrogate model for broad exploration and the precision of traditional simulators for final validation, creating a more efficient and comprehensive decision-making process for coastal resilience.

Although CASPIAN-v2 represents a significant advancement in surrogate modeling for flood prediction, it is essential to highlight certain limitations. The prediction accuracy of the model largely depends on the underlying hydrodynamic models, where any inaccuracies in land surface conditions, atmospheric influences, or bathymetric data can introduce biases that may compromise the model outputs. Future research could explore incorporating different geographical contexts, including detailed elevation data and hydro-connectivity information through additional input channels, thereby further enhancing inundation predictions and reliability. Moreover, domain adaptation techniques and incremental learning could accelerate the implementation of models in various domains. Additionally, improving overall operational utility would enhance scalability and interpretability through model compression, distributed training, or more sophisticated explainable artificial intelligence algorithms.

In conclusion, the CASPIAN-v2 model offers a robust, adaptable, and comprehensible approach to predicting coastal floods. The proposed model incorporates computer vision and a DL-inspired framework to address the complexities of diverse geographical regions, protection scenarios, and climate variability. The CASPIAN-v2 model effectively identifies critical inundation areas, handles uneven data distribution, and provides a clear rationale for its predictions. These strengths position CASPIAN-v2 as an essential tool for coastal resilience planning, helping decision makers, engineers, and legislators address current and future flood risks in the context of rapidly rising sea levels and changing coastal conditions.

\codedataavailability{The code and data supporting the findings of this study are available on the project page: https://caspiannet.github.io}

\authorcontribution{BH, AK, ACHC, and SM contributed to the conceptualization and validation of research. BH and AK performed the formal analysis, investigation, and visualization; BH also developed the software. ACHC managed data curation, contributed to the methodology, and provided essential resources. SM secured funding, oversaw project administration, and offered supervision. All authors participated in writing the original draft and in reviewing and editing the manuscript.} 

\competinginterests{The authors declare that they have no conflict of interest.}

\begin{acknowledgements}
This research was supported by New York University Abu Dhabi.
\end{acknowledgements}

%%%%%%%%%%%%%%%%%%%%%%%%%%%%%%%%%%%%%%
\bibliographystyle{Copernicus}
\bibliography{Refs.bib}

\pagebreak

\appendix

%%%%%%%%%%%%%%%%%%%%
\section{Hydrodynamic model data generation}\label{App:hydro}

This section provides additional detail regarding the hydrodynamic model simulations performed to generate training data for both the Abu Dhabi and San Francisco Bay shorelines.
For simulating the tidal dynamics of the considered region under different coastline protection scenarios, we utilized Delft3D (~\citep{delft}), a hydrodynamic model that dynamically solves the Reynolds-Averaged Navier-Stokes differential equations, i.e., it is a physics-based numerical model that considers the time-varying forces due to hydrostatic pressures (such as SLR), tidal forcing, meteorological stresses, bottom seabed friction, and river inflows over a finite-element computational grid (down to $30$~m in horizontal resolution) spread over variable bathymetry. For any point in the grid, Delft3D will provide time series outputs at $30$-minute intervals of water levels and local water circulation velocities throughout the specified simulation period. Importantly, the Delft3D model can handle computational grid cells that alternate between dry and wet states~\citep{barnard2014development}. 

The Delft3D model was validated (i.e., compared with observed tidal water levels) by running the simulator over a 3-month period between 1 January and 31 March 2017 (without wind forcing) and computing the root mean squared error between the model outputs of hourly water levels at 194 locations throughout the gulf and hourly tidal gauge water level data obtained from the TPXO8 Ocean Atlas for the same period (https://www.tpxo.net/global/tpxo8-atlas). The model was calibrated by adjusting the bottom Manning’s roughness coefficient to obtain the lowest error between the predicted water levels the tidal gauge data at located throughout the gulf. Further details concerning the validation of the  hydrodynamic model can be found in \citep{aaronchow2022combining}.

To account for storm surges and wave activity on the Abu Dhabi shoreline (which can be significant especially during prolonged Shamal events), the validated Delft3D model was rerun with wind forcing from the ERA5 database, and the results were fed to an additional spectral wave model, Simulating Waves Nearshore (SWAN) \citep{swan}, which allows capturing wind-wave generation, wave diffraction, amplification and refraction of water surface waves as they approach the shoreline. The SWAN model was applied at a scale of about $100$~km along the shoreline to about $50$~km offshore under the same forcing from the ERA5 database. Finally, where the waves are in contact with the coastline, the SWAN-computed significant wave heights and the local shoreline slope were used to compute the run-up elevations along the coastline where the waves hit the shore. 

\begin{figure}[htbp]
    \centering
    \includegraphics[width=0.95\textwidth]{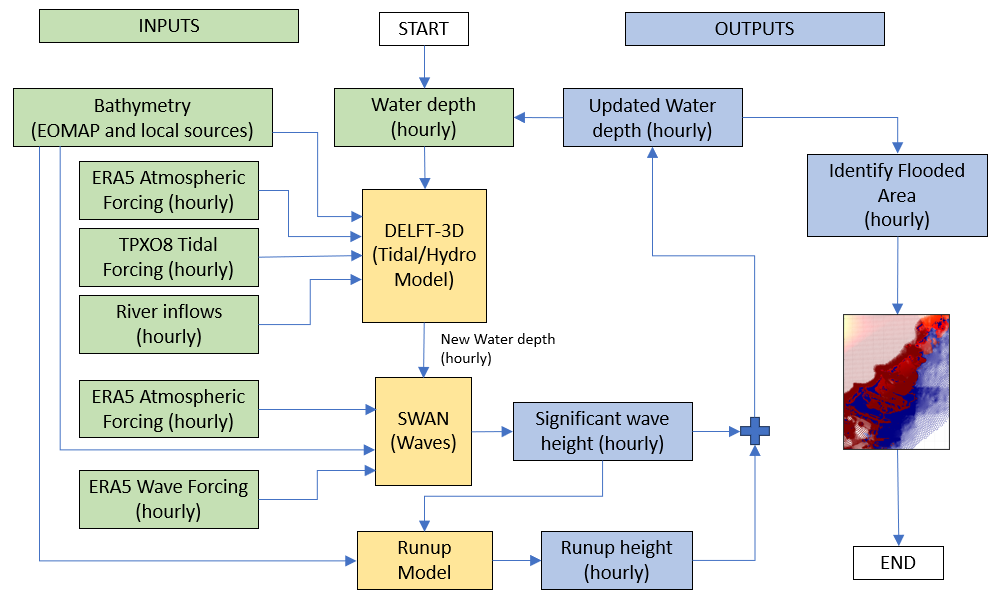}
    \caption{Schematic of hydrodynamic model elements used for the Abu Dhabi scenarios. Green elements denote input parameters; Yellow are the constituent sub-models; Blue are model outputs. Running one cycle of the above model will generate an hourly update of the water depth throughout the UAE coastline.}
    \label{fig:Hydro_Schematic_AD}
\end{figure}

\begin{figure}[htb]
    \centering
    \includegraphics[width=\textwidth]{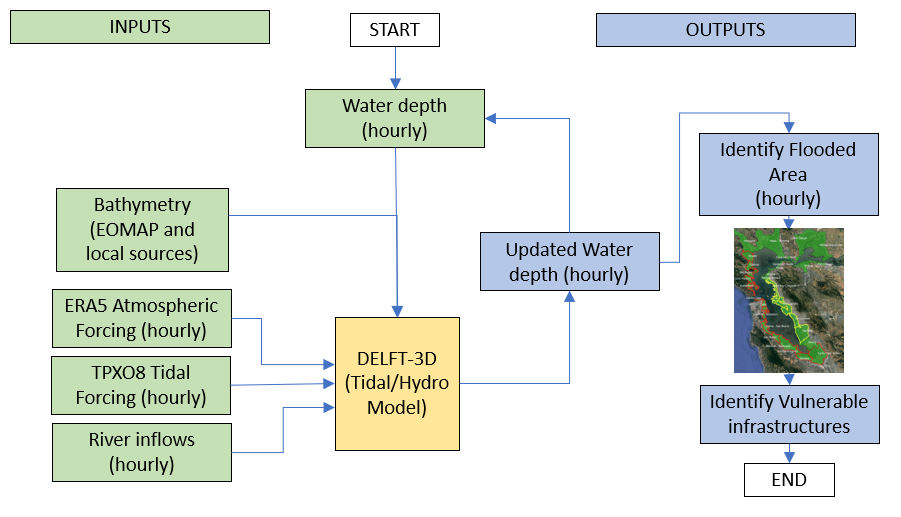}
    \caption{Schematic of hydrodynamic model elements used for the San Francisco scenarios. Green elements denote input parameters; Yellow are the constituent sub-models; Blue are model outputs. Running one cycle of the above model will generate an hourly update of the water depth throughout the shoreline of San Francisco Bay.}
    \label{fig:Hydro_Schematic_SF}
\end{figure}

Owing to the location of San Francisco Bay as an inland bay, the 30 individual OLU shorelines are sheltered from the storm surges by the exterior Californian coastline, with mean significant wave heights within the Bay at about 0.07-0.2~m (https://www.usgs.gov/data/modeled-surface-waves-winds-south-san-francisco-bay), in contrast to about 2.0-3.0~m at Point Reyes located on the California coast outside the Bay. As such, SWAN was not employed for San Francisco Bay, and Delft3D alone was used for the prediction of the coastal flooding events due to SLR, as shown in Fig.~\ref{fig:Hydro_Schematic_SF}. Further details about the San Francisco and Abu Dhabi hydrodynamic models are found in \citep{Jiayunsun2020multimodal} and \citep{aaronchow2022combining}, respectively.

\section{Data Preprocessing}\label{Preprocessing}
\begin{figure}[htb]  
\centering
\includegraphics[width=1\linewidth]{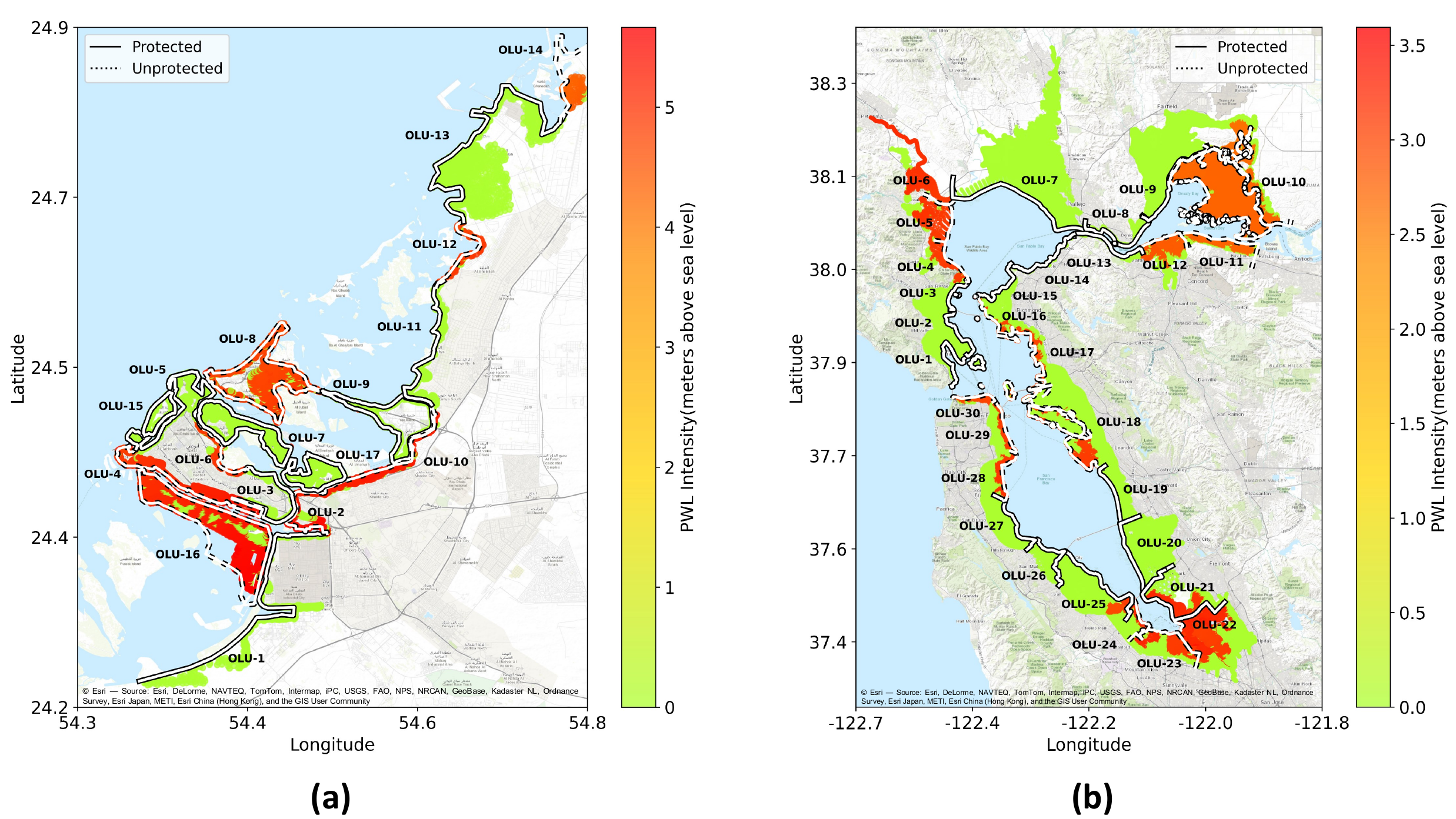}
\caption{Effects of adopted shoreline protection strategies on inundation for (a) Abu Dhabi and (b) San Francisco Bay Area. In AD, alternate OLUs are protected under a 0.5~m SLR scenario (10101010101010101\_0.5). For SF, sets of three consecutive OLUs alternate between protected and unprotected under a 1.0~m SLR scenario (111000111000111000111000111000\_1.0). Here, '1' represents protected OLUs, while '0' denotes unprotected OLUs.}
\label{fig:RandomProt}
\end{figure}

This section details the data preprocessing strategy employed to transform raw hydrological simulation data into tensors for training the proposed DL model. The raw data from the Delft3D simulator included inland inundation coordinates $x$ and $y$, along with the corresponding peak water level (PWL) values for each coordinate pair ($x,y$). The filenames for each case reflect the underlying OLU protection scenarios represented as binary digits, followed by an underscore and the SLR depth. These binary sequences encode the protection status of OLUs, where '1' indicates a protected OLU and '0' indicates an unprotected OLU. Figure \ref{fig:RandomProt} shows randomly selected scenarios for both AD and the SF Bay Area. In the AD scenario (Figure \ref{fig:RandomProt}(a)), the 17 OLUs are represented by \textit{“10101010101010101\_0.5”}, where every alternate OLU, starting from the first, is protected while the others remain unprotected. For the SF Bay Area (Figure \ref{fig:RandomProt}(b)), the 30 OLUs are represented by the binary sequence \textit{“111000111000111000111000111000\_1.0”}, where sets of three consecutive OLUs alternate between protected and unprotected status, with a 1.0~m SLR scenario applied. Additionally, the figures depict an inland flood map, allowing a detailed analysis of flood risks under these protection scenarios. The raw datasets were further processed using the following steps to generate a uniform grid representation suitable for training. 

\subsection{Coordinate Transformation}

The first step in preprocessing involves transforming the original coordinates to a common coordinate system, which is essential for accurate spatial analysis. Specifically, we transformed the coordinates $(x, y)$ of the inundation points from WGS 84 / UTM zone 10N to a standardized latitude and longitude system.
%%%%%%%%%%%%%%%%%%%%%%%%
%%%%%%%%%%%%%%%%%%%%%%%%%%%%%%%%

\subsection{Grid Mapping and Generation}

Next, we mapped the coordinates $(x,y)$ of each inundation point and their corresponding PWL values onto a regular grid to make the dataset compatible with the proposed DL framework. Since CNNs require input data in grid format for effective application of convolution operations, we first generated a grid and then transformed the remaining samples in the datasets accordingly. The grid is defined over the spatial extent of the dataset, with a resolution of $N \times N$. In this research, we set $N = 1024$ to provide an input of $1024 \times 1024$ to the proposed DL-based flood prediction model. The grid points $\xi$ and $\eta$ are computed, as expressed in Eq. (\ref{Eq1a}-\ref{Eq1b}):

\begin{align}
\xi &= \left\{ x_{\min}, x_{\min} + \frac{x_{\max} - x_{\min}}{N-1}, \dots, x_{\max} \right\} \label{Eq1a} \\
\eta &= \left\{ y_{\min}, y_{\min} + \frac{y_{\max} - y_{\min}}{N-1}, \dots, y_{\max} \right\}
\label{Eq1b}
\end{align}

where $x_{\min}, x_{\max}, y_{\min}, y_{\max}$ represent the minimum and maximum $x$ and $y$ values of the inundation points in the dataset. These values define the spatial boundaries for the grid, which serves as the foundational structure upon which the dataset of inundation points is projected. Next, we mapped each original inundation point $(x_i, y_i)$ to the nearest grid cell $(i', j')$, as expressed in Eq. (\ref{Eq2a}-\ref{Eq2b}):

\begin{align}
i' &= \max \left(0, \min \left(N-1, \left\lfloor \frac{x_i - x_{\min}}{x_{\max} - x_{\min}} \cdot (N-1) \right\rfloor \right) \right) \label{Eq2a} \\
j' &= \max \left(0, \min \left(N-1, \left\lfloor \frac{y_i - y_{\min}}{y_{\max} - y_{\min}} \cdot (N-1) \right\rfloor \right) \right)
\label{Eq2b}
\end{align}

This ensures that each inundation point is mapped to the closest grid cell, maintaining spatial consistency and ensuring that all points are correctly positioned within the defined grid space. Conflicts can occur during grid mapping when multiple inundation points are allocated to one grid cell, due to its finite resolution and data point density. To preserve spatial integrity and achieve a one-to-one correspondence between inundation points and grid cells, we adopted the following conflict resolution strategy. Initially, we identified all grid cells that had multiple inundation points mapped to them. Let $\mathcal{C}$ denote the set of conflicting grid cells, where each grid cell $(i', j') \in \mathcal{C}$ has multiple inundation points $\{(x_k, y_k)\}_{k=1}^{n}$ mapped to it. To resolve these conflicts, we reassigned the conflicting points to the nearest available grid cell. This was done by first identifying a subset of grid cells located within a specified distance from the conflicted point. From this subset, we then calculated the Manhattan distance ($d_{m} = |i_{new}' - i'| + |j_{new}' - j'|$) between each candidate cell and the conflicted point, and selected the grid cell with the smallest distance. By systematically resolving these conflicts, we ensured that each inundation point was uniquely assigned to a distinct grid cell, resulting in a bijective mapping $\mathcal{M}: (x_i, y_i) \rightarrow (i', j')$. The grid structure was then used to process the remaining samples and generate training and evaluation data for the proposed DL model.

\subsection{OLU-Based Classification}

In the next step, we classified each inundation point based on its proximity to protected and unprotected OLUs to incorporate the influence of shoreline protection measures into the model. As mentioned earlier in Sec~\ref{Preprocessing}, each protection scenario is represented by a binary string $s = s_1 s_2 \dots s_K$, where $K$ is the total number of OLUs, and each bit $s_k \in \{0, 1\}$ indicates the protection status of the $k$-th OLU ($0$ for unprotected, $1$ for protected). For each inundation point $(x_i, y_i)$, we determined its classification value $c_i$ based on the minimum great-circle distances to the protected and unprotected OLUs. The great-circle distance between an inundation point and a point on an OLU boundary (either protected or unprotected) on the Earth's surface was calculated using the Haversine formula. Specifically, $x_i$ and $y_i$ represent the latitude and longitude of the point of inundation, while $x_j$ and $y_j$ represent the latitude and longitude of a point on the boundary of the OLU. The Haversine formula calculates the distance $d_{ij}$, as expressed in Eq. (\ref{Eq3}):

\begin{align}
d_{ij} &= 2R \cdot \arcsin \left( \sqrt{ \sin^2 \left( \frac{\Delta x_{ij}}{2} \right) 
+ \cos(x_i) \cos(x_j) 
\sin^2 \left( \frac{\Delta y_{ij}}{2} \right) } \right),
\label{Eq3}
\end{align}

where $\Delta x_{ij} = x_j - x_i$, $\Delta y_{ij} = y_j - y_i$, and $R = 6,367$ km is the Earth's approximate radius. The minimum distances are determined by iterating over all points defining the OLU boundaries, allowing us to find the closest OLU boundary (either protected or unprotected) for each inundation point, as expressed in Eq. (\ref{Eq4a}-\ref{Eq4b}):

\begin{align} d_{\text{prot}_i} &= \min_{j \in \mathcal{P}} d_{ij}, \label{Eq4a}\\
d_{\text{unprot}_i} &= \min_{j \in \mathcal{U}} d_{ij}, 
\label{Eq4b}
\end{align}

where $\mathcal{P}$ and $\mathcal{U}$ are the sets of points on the boundaries of protected and unprotected OLUs, respectively. Lastly, the classification value $c_i$ is assigned to each inundation point based on the distances calculated, as expressed in Eq. (\ref{Eq5}):

\begin{align}
c_i = \begin{cases}
1, & \text{if } d_{\text{unprot}} \leq d_{\text{prot}}, \\
-1, & \text{if } d_{\text{prot}} < d_{\text{unprot}}.
\end{cases}
\label{Eq5}
\end{align}

A higher classification value (1) indicates that the inundation point is closer to an unprotected area. In scenarios where one of the distances is undefined (e.g., if there are no protected or unprotected areas in the scenario), the classification is determined based solely on the available distance. This allowed us to encode the binary input, providing the DL model with essential contextual information about shoreline protection, which can significantly influence inundation patterns and PWL values.

\subsection{Training Data Generation}

\begin{figure}[htb]  
\centering
\includegraphics[width=0.5\linewidth]{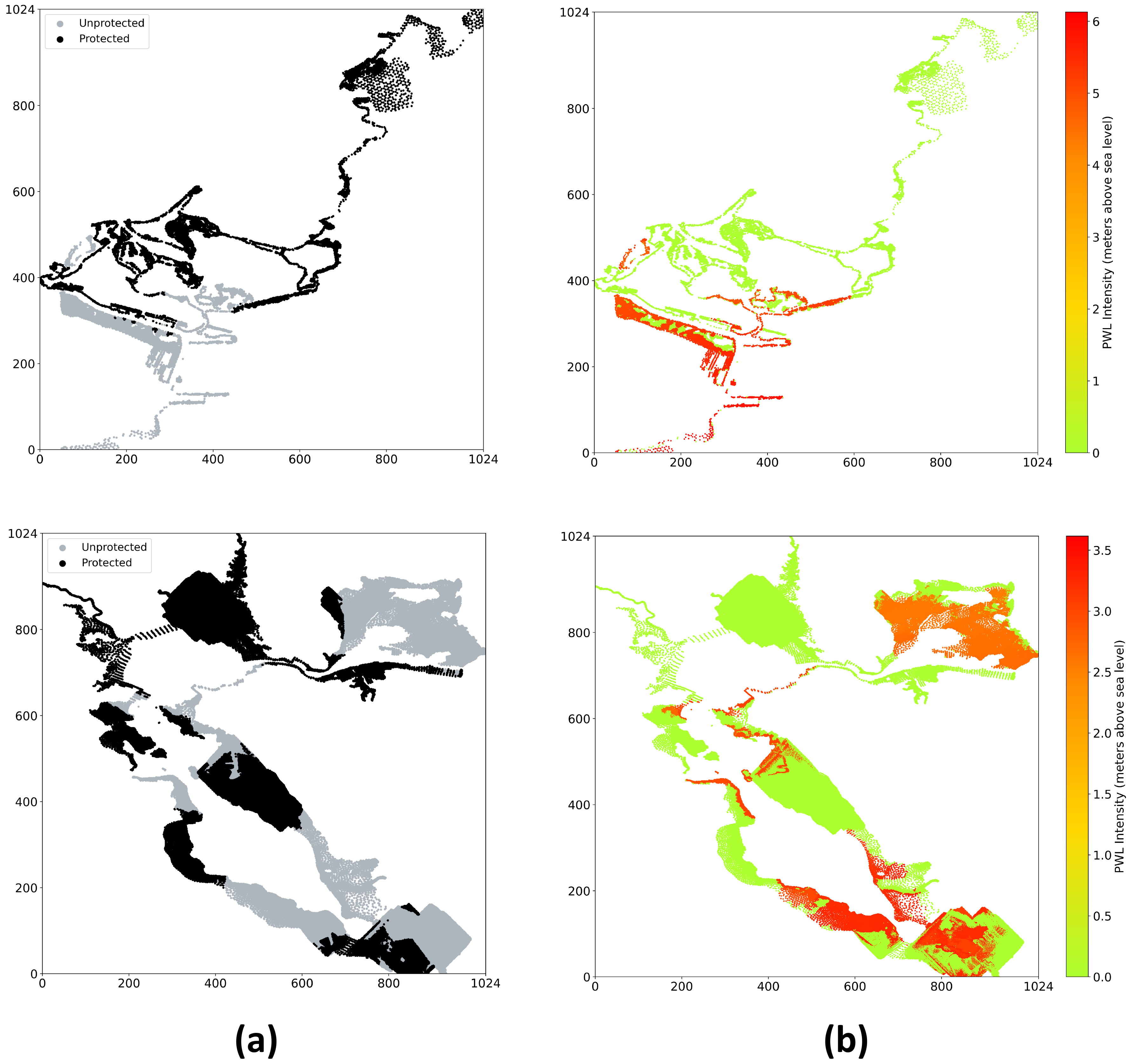}
\caption{Input and output representations for training the proposed deep learning model, with the top row corresponding to AD and the bottom row to the SF Bay Area. (a) Input images represent inundation points categorized as protected or unprotected based on on their proximity to the nearest OLU. (b) Output images depict inundation values at each point (ground-truth data for training).}
\label{fig:I/OImages}
\end{figure}

With the inundation points mapped to the grid and classified, the input and output matrices are generated to train and evaluate the DL model. The input matrix $\mathbf{I} \in \mathbb{R}^{N \times N}$ represents the spatial distribution of protected and unprotected areas, while the output matrix $\mathbf{O} \in \mathbb{R}^{N \times N}$ contains the PWL values associated with each grid cell. For each grid cell $(i', j')$, the input and output matrices are generated by assigning the values, as expressed in Eq. (\ref{Eq6}):

\begin{align}
(I_{i'j'}, O_{i'j'}) &= \begin{cases}
(c_i, z_i), & \text{if an inundation point is mapped to } (i', j'), \\
(0, 0), & \text{otherwise}.
\end{cases}
\label{Eq6}
\end{align}

here, $c_i$ is the classification value of the inundation point mapped to the grid cell $(i', j')$, and $z_i$ is the PWL value of the inundation point corresponding to grid cell $(i', j')$. In cases where no inundation point was mapped to a particular grid cell, both the input and output values were set to zero. Figure \ref{fig:I/OImages} shows randomly selected input and output matrices from both AD and SF datasets. The systematic data preprocessing pipeline ensured that the matrices accurately represented the spatial distribution of the data and were properly formatted, facilitating effective training and evaluation of the proposed DL-based flood prediction model.

%%%%%%%%%%%%%%%%%%%%%%%%
%%%%%%%%%%%%%%%%%%%%%%%%%%%%%%%%

%%%%%%%%%%%%%%%%%%%%%%%%
\begin{figure}[htbp]  
\centering
\includegraphics[width=1\linewidth]{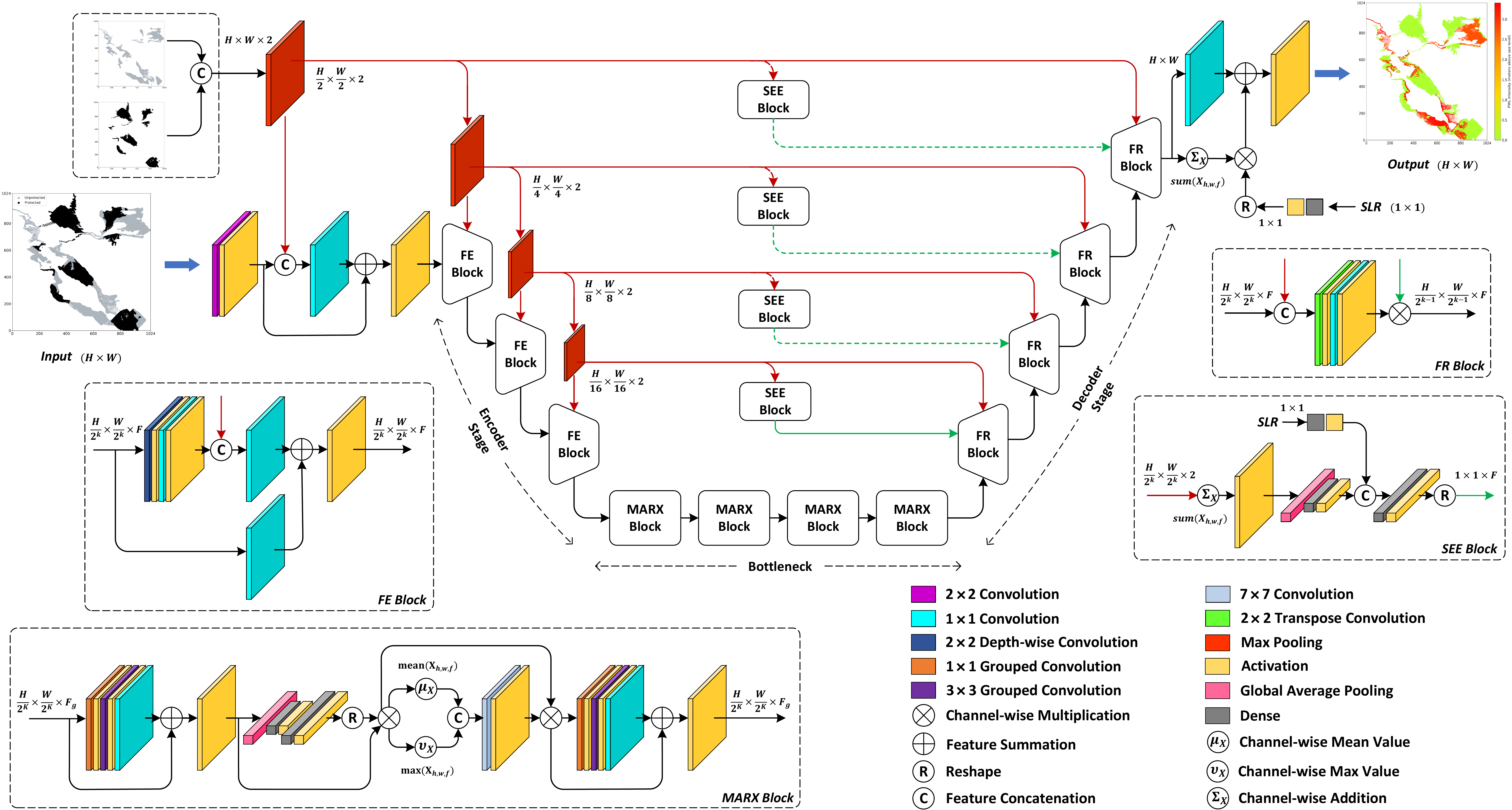}
\caption{The CASPIAN-v2 model architecture. The encoder extracts hierarchical features using FE blocks; the bottleneck employs MARX blocks to capture high-level representations; and the decoder reconstructs outputs using FR and SEE blocks. Different colors show separate layer operations.}
\label{fig:ModelArchS}
\end{figure}
%%%%%%%%%%%%%%%%%%%%%%%%%%%%%%%%

%%%%%%%%%xxxxxxxxxxxxxxxxxxxxxxxxxxxxxxxxxxxxxxxxxxxxxxxxxxxxxxxxxxxxxxxxxxxxxxxxxxxxxxxxxxxxxxxxxxxxxxxxxxxxxxxxxxxxxxxxxxxxxxxxxxxxxxxxxxxxxxxxxxxxxxxxxxxxxxxxxxxxxxxxxxxxxxxxxxxxxxxxxxxxxxxxxxxxxxxxxxxxxxxxxxxxxxxxxxxxxxxxxxxxxxxx
\section{CASPIAN-v2 Architecture}
This section provides a detailed technical implementation of the \textit{CASPIAN-v2} architecture (see Fig. \ref{fig:ModelArchS}), expanding on the conceptual overview presented in Section 4.1 of the main manuscript.

\subsection{Encoder Stage}

The encoder stage of the CASPIAN-v2 model is designed to extract hierarchical features by progressively reducing the spatial dimensions of the input grid while increasing the depth of the feature maps. This process enables the network to capture multi-scale patterns essential for accurate flood prediction.

The model accepts two inputs: an input grid $\mathbf{I} \in \mathbb{R}^{H \times W \times 1}$, where $H$ and $W$ are the spatial dimensions, and a scalar SLR value denoted as $\mathcal{S}$. Since the SLR input $\mathcal{S}$ contains global information affecting the entire spatial domain, it is integrated directly into the decoder part of the network. The encoder stage contains a series of \textit{feature extraction} (FE) blocks, allowing the model to capture both local features, such as specific inundation points and their immediate surroundings, and global patterns, including the overall spatial distribution of protected and unprotected areas.

First, the preprocessed input grid $\mathbf{I}$ is processed through a series of depthwise separable convolutional layers to reduce the spatial dimensions and extract complex features. At each depth level $k = 1$ to $K$, where $K$ is the total depth of the encoder, the feature map $\mathbf{X}_k$ undergoes several transformations. First, a $2 \times 2$ depthwise convolution with stride $2$ is applied to the input feature map $\mathbf{X}_k$, capturing spatial correlations within each channel independently while significantly reducing computational cost compared to standard convolutions. Following that, a single stride $1 \times 1$ pointwise convolution is used to combine the outputs across channels, allowing for feature interaction and increasing the depth of the feature map. In addition, $2 \times 2$ pooling operations with stride $2$ are applied at each depth level $k$. The pooled features from the previous layer are also concatenated with the output of the pointwise convolution, enhancing the feature representation by merging hierarchical features from different scales. This enables the network to capture intricate patterns by combining information in different resolutions, which is essential to interpret how local features contribute to the overall risk of flooding.

We used residual connections to maintain key spatial data and increase network depth. Incorporating a modified $\mathbf{X}_k$ into the concatenated output mitigates gradient vanishing and improves identity mapping learning. These connections preserve crucial features of the early layer and streamline network training.

This process is repeated for each FE block in the encoder, leading to a gradual reduction in the spatial dimensions of the feature maps. At each FE block \(k\), the spatial dimensions are reduced by factor 2, so the resulting feature maps have dimensions \(\tfrac{H}{2^k} \times \tfrac{W}{2^k} \times F\), where \(F\) is the number of channels after concatenation. Such a progressive decrease in the spatial dimensions allows the network to capture more significant receptive fields, collecting information from more extensive regions of the input grid, which is critical for the simulated spread of inundation under various SLR situations. The input grid $\mathbf{I}$, by the end of the encoder stage, transforms into dense feature maps \(\mathbf{X}_{\text{enc}} \in \mathbb{R}^{H' \times W' \times F}\) that capture both local and global data patterns, serving as input to the next stage.

%%%%%%%%%%%%%%%%%%%%%%%%
\subsection{Bottleneck Stage}
%%%%%%%%%%%%%%%%%%%%%%%%
The bottleneck stage of the CASPIAN-v2 comprises a novel block called the \textit{multi-attention ResNeXt} (MARX) block to enhance the ability of the model to focus on the most informative parts of the data. It integrates ResNeXt blocks \cite{xie2017aggregatedRESNEXT} with the \textit{convolutional block attention module} (CBAM) \cite{woo2018cbam}. The output feature maps from the encoder ($\mathbf{X}_{\text{enc}} \in \mathbb{R}^{H' \times W' \times F}$) serve as the input to the bottleneck stage. The MARX Blocks process feature maps through a sequence of operations, starting with a ResNeXt block, followed by a CBAM module, and concluding with an additional ResNeXt block. In the first ResNeXt block, the input feature map is divided into $G$ groups, and group-specific convolutions are applied using $1\times1$ and $3\times3$ kernels, as expressed in Eq. (\ref{Eq7}):
%%%%%%%%%
\begin{align}
    \mathbf{X}_{\text{R1}} = \sigma\left( \mathbf{X}_{\text{enc}} + \sigma\left( \sum_{g=1}^{G} \mathbf{W}_g \ast \mathbf{X}_{\text{enc}, g} + \mathbf{b}_g \right)\right),
\label{Eq7}
\end{align}
%%%%%%%%%
where $\mathbf{X}_{\text{enc}}$ is the input feature map, $\mathbf{W}_g$ and $\mathbf{b}_g$ are the weights and biases for the $g$-th group, $\oast$ denotes the grouped convolution operation, and $\sigma$ is the activation function. $\mathbf{X}_{\text{R1}} \in \tfrac{H}{2^{K}} \times \tfrac{W}{2^{K}} \times F_g$ is the output feature map of the first ResNeXt block, where $K$ is the total depth of the encoder stage, and $F_g$ is the number of channels determined by multiplying the cardinality $\mathcal{C}$ with the bottleneck width $\mathcal{B}$ ($F_g = \mathcal{C} \times \mathcal{B}$).

Next, to refine the feature maps and enable the model to focus on the most informative aspects of the data, we integrated the CBAM principle within the MARX block. The CBAM enhances the representational power of the model by sequentially applying attention mechanisms along both the channel and spatial dimensions. In the channel attention module, inter-channel relationships are captured by computing a channel attention map $\mathbf{M}_{\text{c}} \in \mathbb{R}^{1 \times 1 \times F}$, which reweights the channels of the feature map, as expressed in Eq. (\ref{Eq8}):
\begin{align}
    \mathbf{M}_{\text{c}} = \delta\left( \mathbf{W}_{\text{c2}} \cdot \sigma\left( \mathbf{W}_{\text{c1}} \cdot \mathbf{z} \right) \right),
\label{Eq8}
\end{align}
where $\mathbf{W}_{\text{c1}} \in \mathbb{R}^{F^r \times F}$ and $\mathbf{W}_{\text{c2}} \in \mathbb{R}^{F \times F^r}$ are the weights of the fully connected layers, $F^r$ is a reduction ratio parameter, $\sigma$ denotes the activation function, and $\delta$ is the sigmoid function. The aggregated channel descriptor $\mathbf{z} \in \mathbb{R}^{F}$ is obtained by applying global average pooling over the spatial dimensions.
\begin{align}
    \mathbf{z}_f  = \frac{1}{H' W'} \sum_{i=1}^{H'} \sum_{j=1}^{W'} \mathbf{X}_{\text{R1},f}(i,j),
\label{eq:S3}
\end{align}
where $\mathbf{X}_{\text{R1},f}$ is the $f$-th channel of the feature map $\mathbf{X}_{\text{R1}}$. The channel attention map $\mathbf{M}_{\text{c}}$ is then applied to the feature map via element-wise multiplication ( $\mathbf{X}_{\text{R1}}' = \mathbf{M}_{c} \odot \mathbf{X}_{\text{R1}}$). This operation emphasizes channels that are more informative for predicting inundation patterns influenced by SLR and protection measures. Following that, a spatial attention map $\mathbf{M}_{\text{s}} \in \mathbb{R}^{H' \times W' \times 1}$ is computed by initially aggregating the feature map across the channel dimension using the average pooling, as expressed in Eq. (\ref{Eqn}):
\begin{align}
    \mathbf{q}(i,j)  = \frac{1}{F} \sum_{f=1}^{F} \mathbf{X}_{\text{R1}}'(i,j,f),
\label{Eqn}
\end{align}
where $\mathbf{q} \in \mathbb{R}^{H' \times W'}$ are the aggregated feature maps. Afterward, a $7 \times 7$ convolution is performed to extract intricate inundation patterns. Next, the spatial attention map $\mathbf{M}_{\text{s}}$ is applied to the refined channel feature map ($\mathbf{X}_{\text{cbam}} = \mathbf{M}_{s} \odot \mathbf{X}_{\text{R1}}'$), allowing the model to concentrate on the spatial regions that are most pertinent for predicting flood inundation, such as areas with high vulnerability due to low elevation or insufficient protection. Finally, the output of the CBAM module, $\mathbf{X}_{\text{cbam}}$, is processed through a second ResNeXt block to better capture the representations of complex features.
\begin{align}
    \mathbf{X}_{\text{R2}} = \sigma\left(\mathbf{X}_{\text{cbam}} + \sigma\left( \sum_{g=1}^{G} \mathbf{W}_g' \ast \mathbf{X}_{\text{cbam}, g} + \mathbf{b}_g' \right)\right),
\label{eq:S5}
\end{align}
where $\mathbf{X}_{\text{R2}}$ is the output feature map of the MARX Block, and $\mathbf{W}_g'$, $\mathbf{b}_g'$ are the weights and biases for the $g$-th group in the second ResNeXt block. The MARX blocks allow the CASPIAN-v2 model to generalize across complex datasets by adaptively concentrating on the most informative features in both channel and spatial dimensions. By the end of the bottleneck stage, the feature maps ($\mathbf{X}_{\text{bn}} \in \mathbb{R}^{H' \times W' \times F}$) are transformed into rich, high-level representations that capture key information about the candidate input scenario. These refined features serve as a strong foundation for the decoder stage, where they are progressively upsampled and combined with the SLR scalar $\mathcal{S}$ to reconstruct the spatial resolution of the input grid.

%%%%%%%%%%
%%%%%%%%%%%%%
%%%%%%%%%%%%%%%%%%%
\subsection{Decoder Stage}

The decoder stage in the proposed CASPIAN-v2 model employs a series of \textit{feature reconstruction} (FR) blocks to progressively upsample the feature maps. After the bottleneck, the refined feature maps \(\mathbf{X}_{\text{bn}} \in \mathbb{R}^{H' \times W' \times F}\) serve as input to the decoder. The main goal is to restore the near-original spatial dimensions \(H \times W\). At each depth level \(k\), the decoder up-samples feature maps by a factor of 2 through transpose convolution, producing \(\mathbf{X}_{\text{dec},k} \in \mathbb{R}^{\tfrac{H'}{2^{k-1}} \times \tfrac{W'}{2^{k-1}} \times F}\). Upsampled maps are then concatenated to corresponding encoder outputs through skip connections, ensuring critical spatial details lost during downsampling are maintained.

To further strengthen the focus of the decoder on critical spatial regions and reflect SLR effects, we propose a novel \textit{SLR-enhanced encoding} (SEE) block. It learns dynamic weighting from the SLR input to adjust decoder features. In the SEE block, the pooled feature maps of each encoder level are aggregated and passed through dense layers to generate spatial weighting coefficients, as expressed in Eq.~(\ref{Eq9}). 
\begin{align}
\mathbf{w}_{\text{sp},k} &= \sigma\left( \mathbf{W}_{\text{sp},k} \,\mathcal{F}\left( \mathcal{AP}\left( \mathbf{X}_{\text{enc},k} \right)\right) + \mathbf{b}_{\text{sp},k} \right),
\label{Eq9}
\end{align}
where $\mathbf{X}_{\text{enc},k}$ is the $k$-th channel of the encoder feature map, $\mathcal{AP}$ denotes average pooling, $\mathcal{F}$ denotes the flattening operation, $\mathbf{W}_{\text{sp},k}$ is weight matrix, and $\mathbf{b}_{\text{sp},k}$ is bias term. Simultaneously, the SLR scalar input $\mathcal{S}$ is processed through a dense layer to produce $\mathbf{w}_{\text{SLR}}$. The spatial and SLR features are then concatenated to form $\mathbf{w}_{\text{comb}} = [ \mathbf{w}_{\text{sp},k} ; \mathbf{w}_{\text{slr}} ]$, which is passed through another dense layer with sigmoid activation to produce the final weighting coefficients $\mathbf{w}_{\text{see},k}$. These weighting coefficients are then reshaped and applied to the decoder feature map via element-wise multiplication $\mathbf{X}_{\text{dec}, k}' = \mathbf{X}_{\text{dec}, k} \odot \mathbf{w}_{\text{see},k}$, where $\mathbf{X}_{\text{dec}, k}$ is the decoder feature map at the corresponding depth level, and $\odot$ denotes element-wise multiplication. This configuration of the SEE Block allows the model to adaptively weigh the decoder features based on both spatial information from the encoder and the global influence of SLR, enhancing the model's ability to predict flood inundation patterns under varying SLR scenarios.

At the output of the final FR block, a convolutional operation is applied to produce a preliminary output grid $\mathbf{O}_{\text{conv}} \in \mathbb{R}^{H \times W \times 1}$. To further enhance this output, the model computes the sum of the features across the channels of the last decoder feature map $\mathbf{X}_{\text{dec}, K}' \in \mathbb{R}^{H \times W \times F}$. Moreover, the SLR input $\mathcal{S}$ is again incorporated at this stage by processing through a dense layer and then applying to the summed features via element-wise multiplication, as expressed in Eq. (\ref{Eq10}):
%%%%%%%
\begin{align}
\mathbf{X}_{\text{sum}} = \left(\sum_{f=1}^{F} \mathbf{X}_{\text{dec}, K}'^{(f)}\odot \mathbf{w}_{\text{slr}}\right),
\label{Eq10}
\end{align}
%%%%%%%%%%
where $\mathbf{X}_{\text{dec}, K}'^{(f)}$ is the $f$-th channel of the feature map. Finally, the enhanced summed features are added to the preliminary output grid before applying the activation function.
%%%%%%%%%%%%%%%
\begin{align}
\mathbf{O} = \sigma\left( \mathbf{O}_{\text{conv}} + \mathbf{X}_{\text{sum}} \right),
\end{align}
%%%%%%%%
where \(\mathbf{O} \in \mathbb{R}^{H \times W \times 1}\) is the final output grid representing the predicted flood inundation map, and \(\sigma\) is the activation function. This allows extra information from the decoder feature maps by summing across the channels, thereby enriching the final output with more comprehensive spatial features. The grid \(\mathbf{O}\) reflects the likelihood or extent of flooding at each spatial point, considering both the local features learned by the encoder and the broader SLR effects used in the decoder. This integrated design helps the CASPIAN-v2 model generate accurate and robust flood inundation maps, which are crucial for planning and mitigating coastal regions impacted by SLR.

\section{Ablation Study} 
In this section, we detail the ablation experiments carried out to optimize the CASPIAN-v2 architecture, including the evaluation of bottleneck components, loss functions, the MARX blocks, the SEE block and the integration of SLR. These experiments were designed to evaluate the impact of each component on model accuracy, generalization, and computational efficiency.

\subsection{Impact of Bottleneck Configurations}
To determine the optimal bottleneck architecture for the CASPIAN-v2 model, we conducted a series of experiments evaluating various configurations. We tested different residual backbones (ResNet, ResNeXt) and attention mechanisms (Squeeze-and-Excitation (SE), Convolutional Block Attention Module (CBAM)), both individually and in combination. Starting with a model that uses a simple convolutional bottleneck, we systematically replaced it with a single block of each new configuration to measure its contribution to performance. The results of this single-block comparison are presented in Table~\ref{tab:table_bottleneck}.

The findings in Table~\ref{tab:table_bottleneck} reveal a clear hierarchy of performance. Adding a standalone attention mechanism like SE or CBAM provides a notable improvement over the baseline. However, the most significant performance gain comes from introducing a residual backbone. Replacing the baseline with a ResNet block substantially reduces errors, and upgrading to a ResNeXt block improves them further, likely due to its ability to capture features from more diverse subspaces. The best results are achieved when these powerful backbones are paired with an attention mechanism. The combination of ResNeXt with CBAM attention, which constitutes a single instance of our proposed MARX block, yielded the optimal performance, achieving the lowest error (AMAE of 0.0605) and the highest R² score (0.8853). This systematic evaluation confirms that our proposed MARX block architecture is the most effective choice for capturing the complex spatial dependencies inherent in our flood prediction task.

\begin{table}[htb]
\setlength{\tabcolsep}{10pt}
\renewcommand{\arraystretch}{1.25}
\centering
\footnotesize
\begin{threeparttable} 
\caption{\label{tab:table_bottleneck}Ablation study of different single-block bottleneck configurations.}
\begin{tabular}{lcccc}
\toprule
\textbf{Bottleneck Configuration} & \textbf{ARMSE $\downarrow$} & \textbf{AMAE $\downarrow$} & \textbf{Avg. R2 Score} $\uparrow$ & \textbf{Model Size} $\downarrow$ \\
\midrule
\textit{Baseline} \\
\quad Convolutional Only & 0.4150 & 0.0791 & 0.8515 & 305150 \\
\midrule
\textit{Attention} \\
\quad SE & 0.4018 & 0.0754 & 0.8590 & 306430 \\
\quad CBAM & 0.3991 & 0.0735 & 0.8612 & 308210 \\
\midrule
\textit{Residual Backbones} \\
\quad ResNet & 0.3955 & 0.0711 & 0.8653 & 322624 \\
\quad ResNeXt & 0.3789 & 0.0685 & 0.8714 & 325480 \\
\midrule
\textit{Residual + Attention} \\
\quad ResNet + SE & 0.3892 & 0.0694 & 0.8688 & 323904 \\
\quad ResNet + CBAM & 0.3805 & 0.0667 & 0.8701 & 324764 \\
\quad ResNeXt + SE & 0.3711 & 0.0628 & 0.8785 & 326760 \\
\quad {ResNeXt + CBAM (MARX Block)} & {0.3627} & {0.0605} & {0.8853} & {327624} \\
\bottomrule
\end{tabular}
\end{threeparttable}
\end{table}

\subsection{Impact of Loss Functions on Model Performance} 
We next report the effect of various loss functions on the performance of the proposed flood prediction model. For the experiments, we considered the CASPIAN-v2$_{\text{BASE}}$ model consisting of an encoder and decoder depth of 4, integrated with 4 MARX blocks in the bottleneck and 4 SEE blocks between each encoder and decoder layer. The SLR scalar integer was incorporated at the output of the bottleneck before feeding features into the decoder. To ensure consistency across experiments, the initial weights were fixed and loaded for each run.

We evaluated the model using different loss functions, including the Huber \citep{HUBER} loss with 3 fixed delta values and 3 dynamic delta ranges, the Quantile \citep{Quantile} loss with $\tau$ values of 0.25, 0.50, and 0.75, the Log-Cosh \citep{LogCosh} loss, and a Hybrid loss function that is a weighted linear combination of 3 loss functions. The evaluation was conducted on the validation set using three metrics: ARMSE, AMAE, and avg. R2 Score. Table \ref{tab:tableA1} summarizes the performance of the CASPIAN-v2$_{\text{BASE}}$ model with different loss functions. As shown in Table \ref{tab:tableA1}, the choice of loss function significantly affects the model's performance. The CASPIAN-v2$_{\text{BASE}}$ model using the Hybrid loss function achieves the best overall performance, with the lowest ARMSE of 0.3136, the lowest AMAE of 0.0465, and the highest avg. R2 Score of 0.9356. This indicates that the Hybrid loss function effectively captures the nuances of the data, leading to more accurate predictions. 

Among the Huber loss configurations with fixed delta values, increasing the delta from 0.25 to 0.75 improves performance. Specifically, the model with 0.75 delta value achieves an ARMSE of 0.3482, an AMAE of 0.0508, and avg. R2 Score of 0.9236, outperforming the configurations with lower delta values by approximately 13.61\% in ARMSE. This suggests that a higher delta value in the Huber loss function allows the model to handle larger errors more effectively. For the Huber loss with dynamic delta ranges, the performance remains relatively consistent, with some variations. The model with a delta range of 0.3 to 0.7 achieves the best results with an ARMSE of 0.3282, an AMAE of 0.0494, and an avg. R2 Score of 0.9261, surpassing the fixed delta results. This  indicates that a dynamic delta may be more beneficial for this application.

The Quantile loss functions perform notably worse than the other loss functions, especially at $\tau = 0.25$, where the ARMSE is 0.8519, and the avg. R2 Score drops significantly to 0.4120. This suggests that the Quantile loss alone may not be suitable for this particular task or requires further tuning. At $\tau = 0.50$, the Quantile loss achieves better results but still lags behind other loss functions, with an ARMSE of 0.4285.

The Log-Cosh loss function results in an ARMSE of 0.3442 and an AMAE of 0.0525, which is competitive with the Huber loss configurations. This indicates that while Log-Cosh is effective, combining it with other loss functions in a hybrid approach yields better results. Overall, the Hybrid loss function, combining the Huber, Quantile, and Log-Cosh in a weighted approach, leads to superior performance, reducing the ARMSE by approximately 4.45\% compared to the next best configuration (Huber - 0.75). The higher avg. R2 Score indicates better explanatory power of the model when using the Hybrid loss function.

\begin{table}[htb]
\setlength{\tabcolsep}{15pt}
\renewcommand{\arraystretch}{1.25}
\centering
\footnotesize
\begin{threeparttable} 
\caption{\label{tab:tableA1}Impact of different loss functions on the CASPIAN-v2 performance.}
\begin{tabular}{ccccc}
\hline
\textbf{Config}	&	\textbf{ARMSE $\downarrow$}	&	\textbf{AMAE $\downarrow$}	&	\textbf{Avg. R2 Score} $\uparrow$	&	\textbf{Model Size} $\downarrow$	\\
\hline
CASPIAN-v2$_{\text{BASE}}$ (Huber* - 0.25)	&	0.3976	&	0.0628	&	0.9129	&	1731257	\\
CASPIAN-v2$_{\text{BASE}}$ (Huber* - 0.50)	&	0.3470	&	0.0534	&	0.9249	&	1731257	\\
CASPIAN-v2$_{\text{BASE}}$ (Huber* - 0.75)	&	0.3435 &	0.0508	&	0.9236		&	1731257	\\
CASPIAN-v2$_{\text{BASE}}$	(Huber\^ - 0.4 to 0.6) &	0.3482	&	0.0527	&	0.9242	&	1731257	\\
CASPIAN-v2$_{\text{BASE}}$	(Huber\^ - 0.3 to 0.7) &	0.3282	&	0.0494	&	0.9261	&	1731257	\\
CASPIAN-v2$_{\text{BASE}}$	(Huber\^ - 0.2 to 0.8) &	0.3450	&	0.0520	&	0.9114	&	1731257	\\
CASPIAN-v2$_{\text{BASE}}$ (Quantile* - 0.25)	&	0.8519	&	0.3213	&	0.4120	&	1731257	\\
CASPIAN-v2$_{\text{BASE}}$ (Quantile* - 0.50)	&	0.4285	&	0.0681	&	0.8865	&	1731257	\\
CASPIAN-v2$_{\text{BASE}}$ (Quantile* - 0.75)	&	0.5399	&	0.1034	&	0.8187	&	1731257	\\
CASPIAN-v2$_{\text{BASE}}$ (Log-Cosh)	&	0.3442	&	0.0525	&	0.9251	&	1731257	\\
CASPIAN-v2$_{\text{BASE}}$ (Hybrid$^\#$)	&	0.3136	&	0.0465	&	0.9356	&	1731257	\\
\hline
\end{tabular}
\begin{tablenotes}
\item $^{*}$ Fixed delta or $\tau$ value, $^{\wedge}$ Dynamic delta range, $^{\#}$ With 0.3 Huber, 0.5 Log-Cosh, and 0.2 Quantile loss weightages. The delta for Huber loss was set dynamically between 0.3 and 0.7, and $\tau$ for Quantile loss was fixed at 0.5.
\end{tablenotes}
\end{threeparttable}
\end{table}

\subsection{Impact of MARX Block on Model Performance}  
In this experiment, we examine the effect of varying the number of MARX blocks in the bottleneck of the CASPIAN-v2 model on flood inundation prediction accuracy. The MARX blocks are critical components designed to capture multi-attention features and enhance the model's learning capacity. To first establish the necessity of the block itself, we evaluated a baseline model with no MARX blocks (CASPIAN-v2$_{\text{NO MARX}}$), which showed significantly poorer performance. We then tested configurations with 2, 4, 6, and 8 MARX blocks, denoted as CASPIAN-v2$_{\text{MARX}\times n}$, where $n$ represents the number of MARX blocks.

The base model architecture remains consistent with the previous experiment, comprising an encoder and decoder depth of 4 layers and 4 SEE blocks between each encoder and decoder layer. The SLR scalar integer is incorporated at the output of the bottleneck before the features are fed into the decoder. To ensure a fair comparison, we employed the Hybrid loss function identified as the best-performing loss function from the previous experiment.

As shown in Table \ref{tab:tableA2}, the number of MARX blocks has a significant impact on the model's performance. The CASPIAN-v2$_{\text{MARX}\times4}$ configuration achieves the best overall performance, with the lowest ARMSE of 0.3136, the lowest AMAE of 0.0465, and the highest avg. R2 Score of 0.9356. This indicates that using 4 MARX blocks in the bottleneck effectively balances model complexity and learning capacity, leading to more accurate predictions.

When using only 2 MARX blocks (CASPIAN-v2$_{\text{MARX}\times2}$), the model shows a decrease in performance, with an ARMSE of 0.3249 and an avg. R2 Score of 0.9213. This suggests that with fewer MARX blocks, the model may not capture sufficient multi-scale features, limiting its ability to generalize well on the validation set.

Increasing the number of MARX blocks to 6 (CASPIAN-v2$_{\text{MARX}\times6}$) slightly improves the AMAE to 0.0469 compared to the 2-block configuration but does not surpass the performance of the 4-block model. The ARMSE remains higher at 0.3195, and the avg. R2 Score decreases to 0.9313. This indicates that adding more MARX blocks beyond a certain point does not necessarily lead to better performance and may introduce redundancy or overfitting.

Further increasing the number of MARX blocks to 8 (CASPIAN-v2$_{\text{MARX}\times8}$) results in a noticeable decline in performance. The ARMSE increases to 0.3346, the AMAE rises to 0.0526, and the avg. R2 Score drops to 0.9184. Additionally, the model size increases significantly to 1.93 million parameters, which may lead to increased computational cost and longer training times without corresponding performance gains.

The trend observed suggests that there is an optimal number of MARX blocks that maximizes the model's ability to learn complex patterns without overfitting. The CASPIAN-v2$_{\text{MARX}\times4}$ model strikes this balance effectively. By capturing sufficient multi-scale features through 4 MARX blocks, the model can generalize better and produce more accurate flood inundation predictions.

%These findings highlight the importance of architectural design choices in deep learning models for flood prediction. Selecting an appropriate number of MARX blocks is crucial for optimizing performance while maintaining computational efficiency. The results suggest that using 4 MARX blocks in the CASPIAN-v2 model provides the best trade-off between model complexity and predictive accuracy for the given task.

\begin{table}[htb]
\setlength{\tabcolsep}{15pt}
\renewcommand{\arraystretch}{1.25}
\centering
\footnotesize
\begin{threeparttable} 
\caption{\label{tab:tableA2}Impact of different number of MARX blocks on the CASPIAN-v2 performance.}
\begin{tabular}{ccccc}
\hline
\textbf{Config}	&	\textbf{ARMSE $\downarrow$}	&	\textbf{AMAE $\downarrow$}	&	\textbf{Avg. R2 Score} $\uparrow$	&	\textbf{Model Size} $\downarrow$	\\
\hline
CASPIAN-v2$_{\text{NO MARX}}$ & 0.4051 & 0.0782 & 0.8750 & 1529661 \\
CASPIAN-v2$_{\text{MARX$\times$2}}$	&	0.3249	&	0.0505	&	0.9213	&	1630459	\\
CASPIAN-v2$_{\text{MARX$\times$4}}$	&	0.3136	&	0.0465	&	0.9356	&	1731257	\\
CASPIAN-v2$_{\text{MARX$\times$6}}$	&	0.3195	&	0.0469	&	0.9313	&	1832055	\\
CASPIAN-v2$_{\text{MARX$\times$8}}$	&	0.3346	&	0.0526	&	0.9184	&	1932853	\\
\hline
\end{tabular}

\end{threeparttable}
\end{table}

\subsection{Impact of SEE Block on Model Performance} 
In this experiment, we assess the effect of varying the number of SEE blocks on the performance of the CASPIAN-v2 model for flood inundation prediction. The SEE blocks are designed to integrate the SLR features between the encoder and decoder layers, contributing to the model's ability to capture SLR-induced details in flood maps. A baseline model without any SEE blocks (CASPIAN-v2$_{\text{NO SEE}}$) was first evaluated, confirming that its inclusion provides a distinct performance benefit. We then tested configurations with 1, 2, 3, and 4 SEE blocks, denoted as CASPIAN-v2$_{\text{SEE}\times n}$, where $n$ indicates the number of SEE blocks used.

\begin{table}[htb]
\setlength{\tabcolsep}{15pt}
\renewcommand{\arraystretch}{1.25}
\centering
\footnotesize
\begin{threeparttable} 
\caption{\label{tab:tableA3}Impact of different number of SEE blocks on the CASPIAN-v2 performance.}
\begin{tabular}{ccccc}
\hline
\textbf{Config}	&	\textbf{ARMSE $\downarrow$}	&	\textbf{AMAE $\downarrow$}	&	\textbf{Avg. R2 Score} $\uparrow$	&	\textbf{Model Size} $\downarrow$	\\
\hline
CASPIAN-v2$_{\text{NO SEE}}$ & 0.3310 & 0.0512 & 0.9245 & 362950 \\
CASPIAN-v2$_{\text{SEE$\times$1}}$	&	0.3212	&	0.0485	&	0.9281	&	405938	\\
CASPIAN-v2$_{\text{SEE$\times$2}}$	&	0.3196	&	0.0479	&	0.9311	&	472927	\\
CASPIAN-v2$_{\text{SEE$\times$3}}$	&	0.3172	&	0.0475	&	0.9315	&	727308	\\
CASPIAN-v2$_{\text{SEE$\times$4}}$	&	0.3136	&	0.0465	&	0.9356	&	1731257	\\
\hline
\end{tabular}
\end{threeparttable}
\end{table}

The base model architecture is consistent with the previous experiments, featuring an encoder and decoder depth of 4 layers and 4 MARX blocks in the bottleneck, as determined optimal from prior analysis. The SLR scalar integer is integrated at the output of the bottleneck before feeding the features into the decoder. We utilized the Hybrid loss function, identified as the best-performing loss function in earlier experiments.

As depicted in Table \ref{tab:tableA3}, increasing the number of SEE blocks positively impacts the model's performance up to a certain point. The CASPIAN-v2$_{\text{SEE}\times4}$ configuration demonstrates the best overall performance, achieving the lowest ARMSE of 0.3136, the lowest AMAE of 0.0465, and the highest avg. R2 Score of 0.9356. This suggests that utilizing four SEE blocks enhances the model's ability to capture spatial edge features effectively, leading to more precise flood inundation predictions.

Starting with a single SEE block (CASPIAN-v2$_{\text{SEE}\times1}$), the model achieves an ARMSE of 0.3212 and an avg. R2 Score of 0.9281. While this configuration performs reasonably well, it lacks the detailed spatial feature enhancement provided by additional SEE blocks. Adding more blocks incrementally improves performance; with two SEE blocks (CASPIAN-v2$_{\text{SEE}\times2}$), the ARMSE reduces to 0.3196 and the avg. R2 Score increases to 0.9311. Similarly, three SEE blocks (CASPIAN-v2$_{\text{SEE}\times3}$) result in an ARMSE of 0.3172 and an avg. R2 Score of 0.9315. The most significant improvement is observed with four SEE blocks, where the avg. R2 Score rises to 0.9356, reflecting a stronger correlation between predicted and actual flood inundation levels.

However, while performance improves with additional SEE blocks, the gain diminishes beyond two blocks. Moreover, the number of parameters grows significantly, from 0.41 million parameters for CASPIAN-v2$_{\text{SEE}\times1}$ to 1.73 million parameters for CASPIAN-v2$_{\text{SEE}\times4}$. This increase leads to higher computational costs and longer training times, making the model less suitable for real-time applications.

Given the relatively modest performance gains compared to the steep increase in model size, we propose using a single SEE block in the final model to strike a balance between performance and efficiency. This minimalistic design ensures the model remains lightweight and suitable for real-time flood prediction tasks, aligning with the research's goal of developing a computationally efficient solution. Future work may explore alternative strategies, such as model pruning or efficient network designs, to further optimize performance without significantly increasing complexity.

\subsection{SLR Integration at Different Network Stages}

In this experiment, we investigate the effect of integrating the SLR scalar integer at various stages within the CASPIAN-v2 model architecture on flood inundation prediction accuracy. The SLR scalar is a critical input representing different sea-level rise scenarios, and understanding the optimal point of integration within the network can enhance the model's predictive capabilities. We explored integrating the SLR scalar at three key stages of the network: at the end of the bottleneck (BN), at the end of the last FR block in decoder stage (FR), and before the final output layer (End). Additionally, we examined combinations of these integration points with the SEE blocks to assess their combined effect on performance.

\begin{table}[htb]
\setlength{\tabcolsep}{15pt}
\renewcommand{\arraystretch}{1.25}
\centering
\footnotesize
\begin{threeparttable} 
\caption{\label{tab:tableA4}Performance evaluation of CASPIAN-v2 with integrating SLR at different stages.}
\begin{tabular}{ccccc}
\hline
\textbf{Config}	&	\textbf{ARMSE $\downarrow$}	&	\textbf{AMAE $\downarrow$}	&	\textbf{Avg. R2 Score} $\uparrow$	&	\textbf{Model Size} $\downarrow$	\\
\hline
CASPIAN-v2$_{\text{SLR$\rightarrow$BN}}$ 	&	0.3212	&	0.0485	&	0.9281	&	405938 \\
CASPIAN-v2$_{\text{SLR$\rightarrow$BN+SEE}}$ &	0.3196	&	0.0465	&	0.9313	&	405942 \\
CASPIAN-v2$_{\text{SLR$\rightarrow$FR}}$ 	&	0.3475	&	0.0597	&	0.9169	&	382950 \\
CASPIAN-v2$_{\text{SLR$\rightarrow$SEE+FR}}$ &	0.3259	&	0.0511	&	0.9170	&	383022 \\
CASPIAN-v2$_{\text{SLR$\rightarrow$End}}$ 	&	0.3199	&	0.0515	&	0.9306	&	383020 \\
CASPIAN-v2$_{\text{SLR$\rightarrow$SEE+End}}$ &	0.3126	&	0.0459	&	0.9320	&	383022 \\
\hline
\end{tabular}
\end{threeparttable}
\end{table}

For consistency, all configurations employed 4 MARX blocks and utilized the Hybrid loss function identified as optimal in previous experiments. We used only 1 SEE block in these experiments, as prior analysis indicated that increasing the number of SEE blocks did not significantly enhance performance but did increase the number of parameters.

The results in Table \ref{tab:tableA4} indicate that the point at which the SLR scalar is integrated into the network significantly impacts the model's performance. Integrating SLR at the end of the bottleneck layer (CASPIAN-v2$_{\text{SLR}\rightarrow\text{BN}}$) achieves an ARMSE of 0.3212 and an AMAE of 0.0485. Incorporating SEE blocks in this configuration (CASPIAN-v2$_{\text{SLR}\rightarrow\text{BN+SEE}}$) slightly improves performance, reducing ARMSE to 0.3196 and AMAE to 0.0465, and increasing the avg. R2 Score to 0.9313.

When integrating SLR at the end of the last FR block (CASPIAN-v2$_{\text{SLR}\rightarrow\text{FR}}$), the model shows the poorest performance with an ARMSE of 0.3475 and an AMAE of 0.0597. Adding SEE blocks to this configuration (CASPIAN-v2$_{\text{SLR}\rightarrow\text{SEE+US}}$) improves the ARMSE to 0.3259 and AMAE to 0.0511, but the performance remains inferior compared to other configurations.

Integrating SLR before the final output layer (CASPIAN-v2$_{\text{SLR}\rightarrow\text{End}}$) achieves an ARMSE of 0.3199 and an AMAE of 0.0515. The best performance is observed when combining SEE blocks with this integration point (CASPIAN-v2$_{\text{SLR}\rightarrow\text{SEE+End}}$), achieving the lowest ARMSE of 0.3126, the lowest AMAE of 0.0459, and the highest avg. R2 Score of 0.9320.

These findings suggest that integrating the SLR scalar before the final output layer, especially in combination with SEE blocks, allows the model to more effectively leverage the SLR information for accurate flood inundation predictions.  The model sizes remain relatively consistent across configurations, with minor differences due to the number of parameters introduced by the SLR integration and SEE blocks. The best-performing model (CASPIAN-v2$_{\text{SLR}\rightarrow\text{SEE+End}}$) has a model size of 0.38 million parameters, which is efficient given its superior performance.

\section{Holdout Dataset} 
To thoroughly evaluate the ability of our model to handle challenging conditions, we curated a specialized holdout set for both the AD and SF regions. The scenarios below were chosen based on the spatial configuration and proximity of the OLUs to ensure diverse yet demanding test cases for the model. Tables~\ref{tab:holdout_AD} and \ref{tab:holdout_SF} list all holdout scenarios for the AD and SF regions, respectively.

% --------------------- AD SCENARIOS --------------------- %
\begin{table}[htb]
\centering
\footnotesize
\caption{Holdout set scenarios for the AD region. \texttt{1} indicates a protected OLU, while \texttt{0} denotes an unprotected OLU.}
\label{tab:holdout_AD}
\begin{tabular}{llll}
\hline
\multicolumn{4}{c}{\textbf{AD Scenarios}}\\
\midrule
00000001111110000 & 00000001111111000 & 00000011111000000 & 00000011111100000 \\
00000111100000111 & 00000111110000011 & 00000111111100000 & 00001110000111000 \\
00001111000011110 & 00001111111110000 & 00011000110001100 & 00011100011100011 \\
00011111111111000 & 00110011001100110 & 00111111111111100 & 01010101010101010 \\
10101010101010101 & 11000000000000011 & 11001100110011001 & 11100000000000111 \\
11100011100011100 & 11100111001110011 & 11110000000001111 & 11110000111100001 \\
11110001111000111 & 11111000000011111 & 11111000001111100 & 11111000011111000 \\
11111100000011111 & 11111100000111111 & 11111110000000111 & 11111110000001111 \\
\hline
\end{tabular}
\end{table}

% --------------------- SF SCENARIOS --------------------- %
\begin{table}[htb]
\centering
\footnotesize
\caption{Holdout set scenarios for the SF region. \texttt{1} indicates a protected OLU, while \texttt{0} denotes an unprotected OLU.}
\label{tab:holdout_SF}
\begin{tabular}{lll}
\hline
\multicolumn{3}{c}{\textbf{SF Scenarios}}\\
\midrule
000001111111111111100000000000 & 000011111111100001111111010101 & 000111000111000111000111000111 \\
001000000110101010111001011111 & 001001100011010100010000110101 & 001100110011001100110011001100 \\
001100111100101000111010000010 & 001101101111101000010100001001 & 001111111010111000001001001100 \\
010011000111110100101010000000 & 010100001011101110100101100001 & 010101010101010101010101010101 \\
010101100100010101111100010000 & 011000001010000011110001111000 & 011000111000100011000001110010 \\
011000111100001111001101001110 & 011010010000101000111110110100 & 011011000000011111011000100101 \\
011100010110000001100011001011 & 011101000010011011111110001010 & 011101011000011111110101011001 \\
011110110100101000001001101110 & 011111000111000101011010001001 & 100000000010100101001101111111 \\
100010101001111100110000100100 & 100100000111111000001001001110 & 100101011111010111011101001100 \\
101001101011010011000100100110 & 101010000010001101100100001010 & 101010101010101010101010101010 \\
101011000111111100110100001100 & 101100111110011101010100111101 & 101110001110001001111001001001 \\
110000101001111101011001111101 & 110001101101101111101101000110 & 110011001100110011001100110011 \\
110111100111100111000010100001 & 110111110011100101000010001100 & 111000111000111000111000111000 \\
111001111010010011111010110010 & 111100100010111111101100110100 & 111101111001000101111101100011 \\
111110001000010010110111011100 & 111111100000000000001111111111 & 111111100011111000011110000000 \\
111111110000000011111111100000 & {} & {} \\  % Last row: single scenario plus empty columns
\hline
\end{tabular}
\end{table}

\section{Generalizability Dataset} 
We further evaluated generalizability of the CASPIAN-v2 model for unseen SLR conditions of 0.5\,m and 1.5\,m. This dataset contains 32 protection scenarios for the SF region: 30 scenarios each with exactly one protected OLU (and the remaining unprotected), a completely unprotected scenario, and a fully protected scenario. Table~\ref{tab:slr_scenarios} lists these configurations in binary form, where \texttt{0} and \texttt{1} denote unprotected and protected OLUs, respectively.

\begin{table}[htbp]
\centering
\footnotesize
\caption{Generalizability set scenarios for the SF region under 0.5\,m and 1.5\,m SLR. Each row contains binary strings of length 30, with \texttt{1} indicating a protected OLU and \texttt{0} indicating an unprotected OLU.}
\label{tab:slr_scenarios}
\begin{tabular}{lll}
\hline
\multicolumn{3}{c}{\textbf{SLR Generalizability Scenarios (SF)}} \\
\midrule
000000000000000000000000000000 & 000000000000000000000000000001 & 000000000000000000000000000010 \\
000000000000000000000000000100 & 000000000000000000000000001000 & 000000000000000000000000010000 \\
000000000000000000000000100000 & 000000000000000000000001000000 & 000000000000000000000010000000 \\
000000000000000000000100000000 & 000000000000000000001000000000 & 000000000000000000010000000000 \\
000000000000000000100000000000 & 000000000000000001000000000000 & 000000000000000010000000000000 \\
000000000000000100000000000000 & 000000000000001000000000000000 & 000000000000010000000000000000 \\
000000000000100000000000000000 & 000000000001000000000000000000 & 000000000010000000000000000000 \\
000000000100000000000000000000 & 000000001000000000000000000000 & 000000010000000000000000000000 \\
000000100000000000000000000000 & 000001000000000000000000000000 & 000010000000000000000000000000 \\
000100000000000000000000000000 & 001000000000000000000000000000 & 010000000000000000000000000000 \\
100000000000000000000000000000 & 111111111111111111111111111111 & \\
\hline
\end{tabular}
\end{table}

%%%%%%%%%%%%%%%%%%%%%%%%
\begin{figure}[htb]  
\centering
\includegraphics[width=0.65\linewidth]{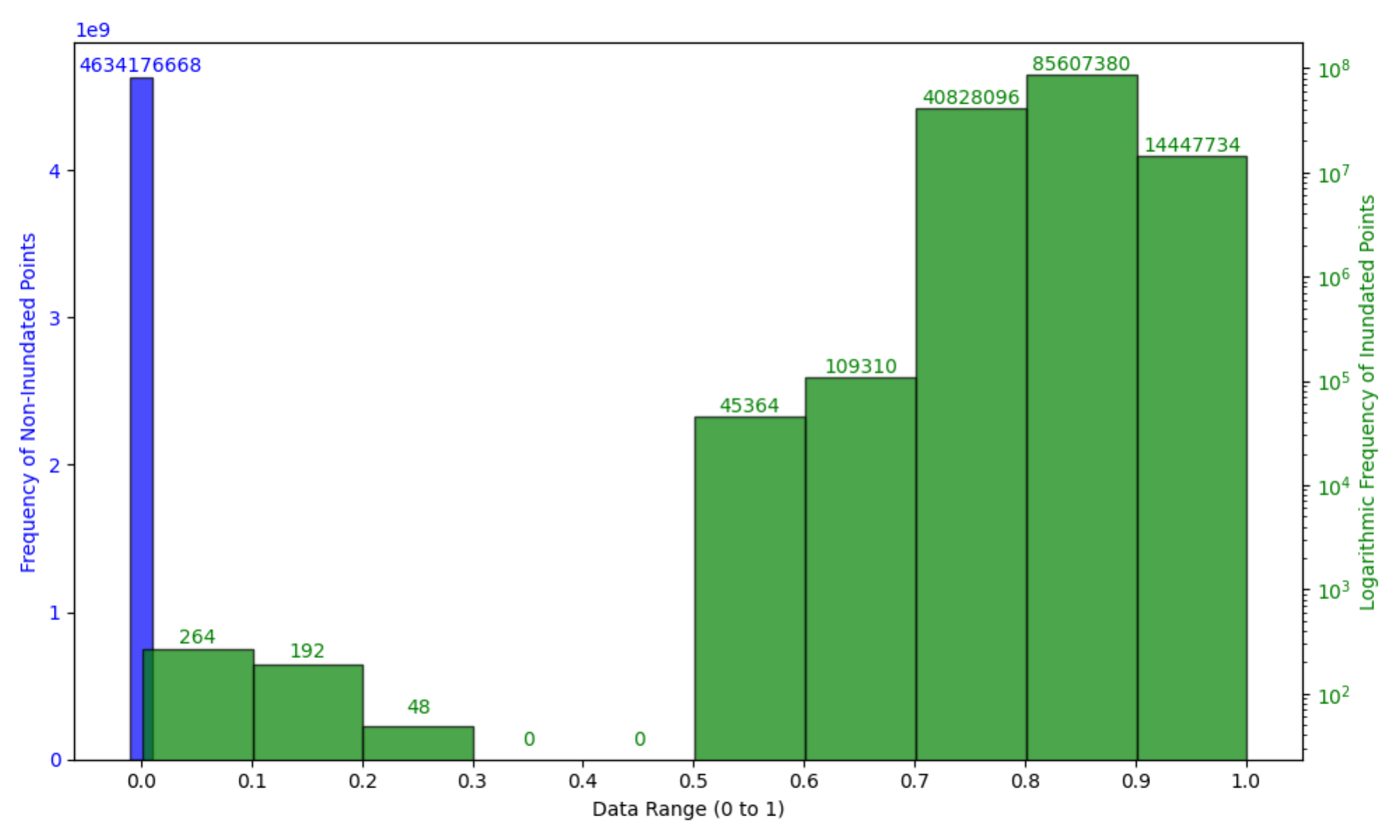}
\caption{Normalized inundation distribution in the complete dataset, depicting a significant class imbalance between non-inundated points (0s) and inundated points (non-0s).}
\label{fig:DataDist}
\end{figure}
%%%%%%%%%%%%%%%%%%%%%%%%%%%%%%%%

\section{Dataset Characteristics}
\label{sec:supp-data}

As mentioned in the main manuscript, the dataset used for training and evaluating the CASPIAN-v2 model exhibits a significant class imbalance. The vast majority of data points correspond to non-inundated areas (a value of 0), while inundated points are comparatively rare. Figure~\ref{fig:DataDist} provides a histogram that visualizes this distribution, illustrating the strong skew that is characteristic of coastal flooding data.

\section{Uncertainty Quantification with Deep Ensembles}
\label{sec:supp_uncertainty}

To address the critical need for uncertainty quantification in risk assessment, we implemented a deep ensemble method, following the well-established approach of Lakshminarayanan et al. (2017). This involved training five CASPIAN-v2 models independently, each initialized with a different random seed. The mean of their predictions was used as the final, robust output, while the pixel-wise standard deviation across the five predictions serves as a direct proxy for the predictive uncertainty of the model.

Table~\ref{tab:supp_ensemble_performance} presents the full quantitative results for each of the five models in the ensemble, as well as for the final ensemble mean prediction. The strong performance and low variance across all models, particularly for the ensemble mean, highlight the stability and robustness of the CASPIAN-v2 architecture. These quantitative results provide the foundation for the uncertainty maps discussed in the main manuscript.

%%%%%%%%%%%%%%%%%%%%%%%%%%%%
% --- FULL SUPPLEMENTARY TABLE --- %
\begin{table}[h!]
\centering
\caption{Full performance metrics for the five individual models in the deep ensemble and for their mean prediction. All metrics are calculated on the combined test set. The robust performance of the ensemble mean highlights the stability of the model.}
\label{tab:supp_ensemble_performance}
\begin{tabular}{l c c c c c | c c}
\toprule
\textbf{Metrics} & \textbf{Model 1} & \textbf{Model 2} & \textbf{Model 3} & \textbf{Model 4} & \textbf{Model 5} & \textbf{Mean} & \textbf{Std} \\
\midrule
AMAE $\downarrow$ & 0.0443 & 0.0453 & 0.0453 & 0.0455 & 0.0452 & \textbf{0.0451} & \textit{0.0004} \\
ARMSE $\downarrow$ & 0.1235 & 0.1326 & 0.1324 & 0.1332 & 0.1319 & \textbf{0.1307} & \textit{0.0036} \\
R² $\uparrow$ & 0.9406 & 0.9383 & 0.9385 & 0.9384 & 0.9387 & \textbf{0.9389} & \textit{0.0009} \\
RTAE $\downarrow$ & 6.5534 & 6.5770 & 6.5459 & 6.5367 & 6.5404 & \textbf{6.5507} & \textit{0.0143} \\
$\delta > 0.5$ $\downarrow$ & 0.8206 & 0.8674 & 0.8886 & 0.8938 & 0.8912 & \textbf{0.8723} & \textit{0.0275} \\
$\delta > 0.1$ $\downarrow$ & 4.2496 & 3.7202 & 3.5476 & 3.5601 & 3.5957 & \textbf{3.7346} & \textit{0.2646} \\
Acc[0] $\uparrow$ & 99.3731 & 99.3119 & 99.3972 & 99.3925 & 99.3950 & \textbf{99.3739} & \textit{0.0322} \\
DSC $\uparrow$ & 0.8206 & 0.8674 & 0.8886 & 0.8938 & 0.8912 & \textbf{0.8723} & \textit{0.0275} \\
\bottomrule
\end{tabular}
\end{table}
%%%%%%%%%%%%%%%%%%%%%%%%%%%%

\section{Notation Table}
\label{sec:supp-notation}

Table \ref{Tablenotation} provides a comprehensive summary of the mathematical notation used throughout the paper to ensure clarity and consistency.

\begin{table}[htb]
\centering
\caption{Standardized mathematical notation used in this study.}
\label{Tablenotation}
\begin{tabular}{l l l}
\toprule
\textbf{Symbol} & \textbf{Type} & \textbf{Description} \\
\midrule
\multicolumn{3}{l}{\textit{\textbf{General Symbols \& Indices}}} \\
$H, W$ & Scalars & Height and width of the spatial grids. \\
$F$ & Scalar & Number of feature channels in a tensor. \\
$N$ & Scalar & Total number of data samples in a set. \\
$d_y$ & Scalar & Total number of pixels in a single sample grid. \\
$K$ & Scalar & Total number of levels/depth in the encoder/decoder. \\
$i$ & Index & Denotes a specific data sample, $i \in \{1, ..., N\}$. \\
$j$ & Index & Denotes a specific pixel within a sample, $j \in \{1, ..., d_y\}$. \\
$k$ & Index & Denotes a specific depth level in the network, $k \in \{1, ..., K\}$. \\
$g$ & Index & Denotes a specific group in a grouped convolution. \\
\midrule
\multicolumn{3}{l}{\textit{\textbf{Input, Output, and Ground Truth}}} \\
$\mathbf{I}$ & Tensor & Input matrix representing shoreline protection scenarios. \\
$\mathbf{O}$ & Tensor & Final predicted flood inundation map. \\
$y_{\text{t},i,j}$ & Scalar & Ground truth PWL value for pixel $j$ of sample $i$. \\
$y_{\text{p},i,j}$ & Scalar & Predicted PWL value for pixel $j$ of sample $i$. \\
$\mathcal{S}$ & Scalar & Sea Level Rise (SLR) value. \\
\midrule
\multicolumn{3}{l}{\textit{\textbf{Model Architecture Tensors}}} \\
$\mathbf{X}_{k}$ & Tensor & Feature map at encoder/decoder depth level $k$. \\
$\mathbf{X}_{\text{enc}}$ & Tensor & Output feature map from the encoder stage. \\
$\mathbf{X}_{\text{cbam}}$ & Tensor & Output feature map from a CBAM module. \\
$\mathbf{M}_{\text{c}}, \mathbf{M}_{\text{s}}$ & Tensors & Channel and spatial attention maps in CBAM. \\
$\mathbf{w}_{\text{see}}$ & Vector & Weighting coefficients learned by the SEE block. \\
$\mathbf{W}, \mathbf{b}$ & Tensors & Learnable weights and biases of a neural network layer. \\
\midrule
\multicolumn{3}{l}{\textit{\textbf{Loss \& Evaluation Symbols}}} \\
$L_{h}, L_{q}, L_{\text{cosh}}$ & Scalars & Huber, Quantile, and Log-Cosh loss values. \\
$\alpha_h, \alpha_q, \alpha_c$ & Scalars & Weighting coefficients for the hybrid loss function. \\
$\delta, \tau, \Delta$ & Scalars & Thresholds for Huber loss, Quantile loss, and exceedance metric. \\
\bottomrule
\end{tabular}
\end{table}

\vspace{20em}

\end{document}